\documentclass[10pt,journal,cspaper,compsoc]{IEEEtran}
\usepackage{amsmath}
\usepackage[dvips]{graphicx}
\DeclareGraphicsExtensions{.eps}
\usepackage{subfigure}
\usepackage{algorithmic}
\usepackage{color}
\usepackage{algorithm}
\usepackage[usenames,dvipsnames]{pstricks}
\usepackage{epsfig}
\usepackage{pst-grad} 
\usepackage{pst-plot} 

\usepackage{array}

\usepackage{stfloats}

\usepackage{url}

\newcommand{\Rmnum}[1]{\uppercase\expandafter {\romannumeral #1}}
\begin{document}
\title{Tag-Weighted Topic Model\\ For Large-scale Semi-Structured Documents}

\author{Shuangyin Li, Jiefei Li, Guan Huang, Ruiyang Tan, and Rong Pan
\IEEEcompsocitemizethanks{\IEEEcompsocthanksitem Shuangyin Li, Jiefei Li, Guan Huang, Ruiyang Tan, and Rong Pan's E-mails: shuangyinli@cse.ust.hk, \{lijiefei@mail2., huangg6@mail2., tanry@mail2., panr@\}sysu.edu.cn. Submitted and reviewed by IEEE Transactions on Knowledge and Data Engineering (TKED).
}
}
\IEEEcompsoctitleabstractindextext{%
\begin{abstract}
To date, there have been massive Semi-Structured Documents (SSDs) during the evolution of the Internet. These SSDs contain both unstructured features (e.g., plain text) and metadata (e.g., tags). Most previous works focused on modeling the unstructured text, and recently, some other methods have been proposed to model the unstructured text with specific tags. To build a general model for SSDs remains an important problem in terms of both model fitness and efficiency. We propose a novel method to model the SSDs by a so-called Tag-Weighted Topic Model (TWTM). TWTM is a framework that leverages both the tags and words information, not only to learn the document-topic and topic-word distributions, but also to infer the tag-topic distributions for text mining tasks. We present an efficient variational inference method with an EM algorithm for estimating the model parameters. Meanwhile, we propose three large-scale solutions for our model under the MapReduce distributed computing platform for modeling large-scale SSDs. The experimental results show the effectiveness, efficiency and the robustness by comparing our model with the state-of-the-art methods in document modeling, tags prediction and text classification. We also show the performance of the three distributed solutions in terms of time and accuracy on document modeling.
\end{abstract}

\begin{keywords}
semi-structured documents, topic model, tag-weighted, variational inference, large-scale, parallelized solutions
\end{keywords}
}

\maketitle
\IEEEdisplaynotcompsoctitleabstractindextext
\IEEEpeerreviewmaketitle

\section{Introduction}\label{sec:Introduction}
\IEEEPARstart{I}{n} the evolution of the Internet, there have been a huge amount of documents in many web applications. Such kinds of documents with both plain text data and document metadata (tags, which can be viewed as features of the corresponding document) are called the Semi-Structured Documents (SSDs). How to characterize the semi-structured document data becomes an important issue addressed in many areas, such as information retrieval, artificial intelligence and data mining etc. The tags can be more important than the text data in document mining. For example, in IMDB \footnote{http://www.imdb.com}, the world's most popular and authoritative source for movie, TV and celebrity content, each movie has lots of tags, like director, writers, stars, country, language and so on, and a storyline as text data. Given a movie with a tag ``Dick Martin", we may have an idea that it has a higher chance to be a comedy, without read the full text of its storyline or watch it. Another example is that in a collection of scientific articles, each document has a list tags(authors and keywords). Before read the main text of paper, we would know what it talks about after we see the authors or the keywords that the paper provides.

Many solutions have been proposed to deal with the semi-structured documents (e.g., SVD, LSI), and shown to be useful in document mining \cite{Bratko2006679,DBLP:journals/corr/abs-0901-0358,DBLP:conf/kdd/YiS00,DBLP:conf/vldb/TreschPL95}, e.g., text classification and  structural information exploiting. For document modeling, topic models have been used to be a powerful method of analyzing and modeling of document corpora, using Bayesian statistics and machine learning to discover the thematic contents of untagged documents. Topic models can discover the latent structures in documents and establish links between them, such as latent Dirichlet allocation (LDA)  \cite{DBLP:journals/jmlr/BleiNJ03}. However, as an unsupervised method, only the words in the documents are modeled in LDA. Thus, LDA could only treat the tags as word features rather than a new kind of information for document modeling.

To model semi-structured documents needs to consider the characteristics of different kinds of objects, including word, topic, document, and tag, and the relationship among them. In this problem, topic is a kind of hidden objects, and the other three are the observations. Relative to tag, word and document are objective; tag can be either objective (e.g., author and venue information of publications) and subjective (e.g., tags in social bookmark marked by people). Similar to the topic models, we should consider binary relationships between the pairs of the objects, including topic-word and document-topic. In addition, we may consider the binary relationships, like tag-word, tag-topic, tag-document, and tag-tag. The tag-document relationship implies that we should consider the weights of the tags in each document. The tag-topic and tag-tag relationships can be more complicated, thus are difficult to model. Some earlier works consider certain tags. For example, the author-topic model in~\cite{DBLP:conf/uai/Rosen-ZviGSS04} considers the authorship information of the documents to be modeled.  In this work, we don't limit the types and number of the tags in each document. In an extreme case, where there is no tag in any document, the new model may degenerates into LDA. On the other hand, since the tags can be created by some people, they should be relevant to topics of the documents; however, some of them may be correlated, redundant, and even noisy. Therefore, the tag-topic relationships should be general enough and we should also model the weights of the tags in each document.

In the past few years, researchers have proposed approaches to model documents with tags or labels \cite{DBLP:conf/uai/MimnoM08,DBLP:conf/emnlp/RamageHNM09,Ramage:2011:PLT:2020408.2020481}. For example, Labeled~LDA~\cite{DBLP:conf/emnlp/RamageHNM09} assumes there is no latent topics and each topic is restricted to be associated with the given labels. PLDA assumes that each topic is associated with only one label~\cite{Ramage:2011:PLT:2020408.2020481}. However, both Labeled LDA and PLDA have implicit assumptions that the given labels should be strongly associated with the topics to be modeled or the labels are independent to each other.

Another problem is that we would get into trouble when we need to deal with large-scale semi-structured documents. A variety of algorithms have been used to estimate the parameters of these proposed topic models for mining documents, such as Monte Carlo Markov chain (MCMC) sampling techniques  \cite{DBLP:journals/ml/AndrieuFDJ03,DBLP:finding}, variational methods  \cite{DBLP:conf/nips/Attias99} and others methods  \cite{DBLP:conf/uai/AsuncionWST09,DBLP:conf/icml/SatoN12}. For sampling methods, actually, we may have to appeal to a tailored solution of MCMC  \cite{DBLP:conf/nips/BleiL05} for a particular model, which would impede the requirement of convergence properties and speed, especially when the corpus comprise millions of words. Variational methods as approximation solutions to some extent improve the learning speed. However, it would also be ineffective on learning speed and model accuracy when it comes to a large-scale corpus.

In this paper, we propose a framework of Tag-Weighted Topic Model (TWTM) to represent the text data and the various tags with weights to evaluate the importance of the tags. Besides learning the topic distributions of documents and generating the topic distributions over words, the framework also infers the topic distributions of tags. The weights of observed tags in each document, which we infer from the dataset, give us an opportunity to provide a method to rank the tags. 

In many web applications, not all the documents in the corpora have tags. There are lots of documents only consist of words without any tags which maybe removed after data preprocessing for denoising. Only consider the weights among tags would not hold this case. To address this problem, we also propose a more flexible model called Tag-Weighted Dirichlet Allocation (TWDA) as an extended model. It is based on TWTM, and learns the weights among a Dirichlet prior and the given tags, not just among the tags. Therefore, TWDA handles not only the semi-structured documents, but also the unstructured documents. For the unstructured documents, TWDA degenerates into latent Dirichlet allocation (LDA). For hybrid corpora which consist of both the semi-structured documents and unstructured documents, TWDA can handle this complex type of corpora more effectively and easily.

For the challenge of modeling large-scale corpora, we propose three distributed schemes for the framework of TWTM model in MapReduce programming framework \cite{dean2008mapreduce}. The proposed model has four principal contributions.
\begin{enumerate}
	\item It is a novel topic modeling method to model the semi-structured documents, not only generating the topic distributions over words, but also inferring the topic distributions of tags.
	\item The TWTM leverages the weights among the observed tags in a document to evaluate the importance of the tags using a function of tag-weighted topic assignment process. The weights are associated with the observed tags in a document providing a way to rank the tags. In addition, this could be used to predict latent tags in the document.
	\item The framework of tag-weighted process is easy to extend for many different real world applications. For example, with the extended model TWDA, we can handle both the multi-tag documents and non-tag documents simultaneously, which is very useful to process some complicated web applications.
	\item Three distributed solutions for TWTM have been proposed that focus on challenges of working at a large-scale semi-structured documents in MapReduce programming framework.
\end{enumerate}

The rest of the paper is organized as follows. In Section~\ref{sec:Relatedworks}, we first analyze and discuss related works. In Section~\ref{sec:TWTM}, after introducing the notations, we present the novel topic modeling framework of TWTM, and give the methods of learning and inference. In Section~\ref{sec:TWDA}, we show the extended model TWDA, and give the process of learning and inference. In Section~\ref{sec:TheoreticalAnalysis}, we will give the theoretical analysis to discuss the differences between TWTM and TWDA, comparing the other topic models. In Section~\ref{sec:LargescaleSolutions}, we propose three distributed solutions of TWTM for a large-scale semi-structured documents. In Section~\ref{sec:ExperimentalAnalysis}, we present the experimental results on three domains to show the performance of the proposed method in document modeling, text classification and the effectiveness and efficiency of the three large-scale solutions on a large scale semi-structured documents modeling. We end the paper in Section~\ref{sec:Conclusion}.

\section{Related Works}\label{sec:Relatedworks}
Topic models provide an amalgam of ideas drawn from mathematics, computer science, and cognitive science to help users understand unstructured data. There are many topic models proposed and shown to be powerful on document analyzing, such as in \cite{DBLP:conf/nips/PettersonSCBN10,DBLP:conf/sigir/Hofmann99,DBLP:journals/jmlr/BleiNJ03,DBLP:conf/nips/BleiM07,DBLP:journals/corr/abs-1002-4665,DBLP:journals/jmlr/ChangB09}, which have been applied to many areas, including document clustering and classification  \cite{DBLP:conf/cikm/CaiMHZ08}, and information retrieval  \cite{Wei:2006:LDM:1148170.1148204}. They are extended to many other topic models for different situation of applications in analyzing text data  \cite{DBLP:conf/nips/IwataYU09,DBLP:conf/nips/Lacoste-JulienSJ08,DBLP:conf/icml/ZhuAX09}. However, most of these models only consider the textual information and can only treat the tag information as plain text as well.

TMBP \cite{DBLP:conf/kdd/DengHZYL11} and cFTM  \cite{DBLP:conf/kdd/ChenZC12} propose the methods to make use of the contextual information of documents for topic modeling. TMBP is a topic model with biased propagation to leveraging contextual information, the authors and venue. TMBP needs to predefine the weights of the author and venue information on word assignment, which limits the usefulness in real applications. The method of cFTM has a very strong assumption that each word is associated with only one tag, either author or venue. In many applications, this assumption may not hold. 

Several models have been proposed to take advantage of tags or labels, such as Labeled LDA  \cite{DBLP:conf/emnlp/RamageHNM09}, DMR  \cite{DBLP:conf/uai/MimnoM08} and PLDA  \cite{Ramage:2011:PLT:2020408.2020481}, or modeling relationships among several variables, such as Author-Topic Model  \cite{DBLP:conf/uai/Rosen-ZviGSS04}. Labeled LDA \cite{DBLP:conf/emnlp/RamageHNM09} get the topic distribution for a document through picking out the several hyperparameter components that correspond to its labels, and draw the topic components by the new hyperparameter without inferring the topic distribution of labels. Labeled LDA does not assume the existence of any latent topics \cite{Ramage:2011:PLT:2020408.2020481}. PLDA  \cite{Ramage:2011:PLT:2020408.2020481} provides another way of modeling the tagged text data, which assumes the generation topics assignment is limited by only one of the given tags for one word, and in the training process, PLDA assumes that each topic takes part in exactly one label, and may optionally share global label present on every document. In Author Topic Model, it obtains the topic distributions of authors, without giving the importance weights among the given authors in each document. DMR  \cite{DBLP:conf/uai/MimnoM08} is a Dirichlet-multinomial regression topic model that includes a log-linear prior on the document-topic distributions, which is an exponential function of the given features of the document. However, DMR doesn't output the tag weights either \cite{DBLP:conf/uai/Rosen-ZviGSS04}, which is useful for tag ranking. 

So in this work, we propose a tag-weighted topic modeling framework which leverages the tag information given in a document by a list of weight values to model the topic distribution of the document. Meanwhile, for a mixture collection of semi-structured documents and unstructured documents, we present an extended model called tag-weighted Dirichlet Allocation which considers both a Dirichlet prior and the tags by the weight values among them. Based on the framework of Tag-Weighted Topic Model, we also show three large-scale solutions under the MapReduce distributed computing platform for large-scale semi-structured documents.

\section{TWTM Model and Algorithms}\label{sec:TWTM}
In this section, we will mathematically define the tag-weighted topic model (TWTM), and discuss the learning and inference methods.
\subsection{Notation}
Similar to LDA \cite{DBLP:journals/jmlr/BleiNJ03}, we formally define the following terms. Consider a semi-structured corpus, a collection of $M$ documents. We define the corpus $D = \{(\mathbf{w}^{1}, \mathbf{t}^{1}), \ldots, (\mathbf{w}^{M}, \mathbf{t}^{M}) \}$, where each 2-tuple $(\mathbf{w}^{d}, \mathbf{t}^{d})$ denotes a document, the bag-of-word representation $\mathbf{w}^{d} = (w^d_1, \ldots, w^d_N)$, $\mathbf{t}^{d} = (t_1^{d}, \ldots,  t_{L}^{d})$ is the document tag vector, each element of which being a binary tag indicator, and $L$ is the size of the tag set in the corpus $D$. For the convenience of the inference in this paper, $\mathbf{t}^{d}$ is expanded to a $l^{d} \times L $ matrix $T^{d}$, where $l^d$ is the number of tags in the document $d$. For each row number $i \in \{1,\ldots,l^{d}\}$ in $T^{d}$, $T^{d}_{i\cdot}$ is a binary vector, where $T^{d}_{ij} = 1 $ if and only if the $i$-th tag of the document $d$\footnote{Note that we can sort the tags of the document $d$ by the index of the tag set of the corpus $D$.} is the $j$-th tag of the tag set in the corpus $D$. In this paper, we wish to find a probabilistic model for the corpus $D$ that assigns high likelihood to the documents in the corpus and other documents alike utilizing the given tag information. 
\subsection{Tag-Weighted Topic Model}
TWTM is a probabilistic graphical model that describes a process for generating a semi-structured document collection. In the previous topic models, a document $d$ is typically characterized by a multinomial distribution over topics, $\theta^{d}$, and each topic $k$ is represented by $\psi_k$, over words in a vocabulary.
Take LDA  \cite{DBLP:journals/jmlr/BleiNJ03} as an example, the generative process of topic distribution of document $d$ is assumed as follows.
 \begin{equation*}
 	\begin{aligned}
		\text{Choose }\theta^d \sim&\text{Dirichlet}(\alpha), \\
		\text{and choose }z_{ni} \sim&\text{Multinomial}(\theta^d), 
 	\end{aligned}
 \end{equation*}
where $\alpha$ is the hyperparameter of $\theta^d$. 
In LDA, the topic distribution $\theta^{d}$ is drawn from a hyperparameter $\alpha$, without considering the given tags. However, the tag information should be more useful for the generation of $\theta^{d}$ than a Dirichlet prior.

\begin{figure}[t]
\begin{center}
\scalebox{0.8} 
{
\begin{pspicture}(0,-2.44)(8.26,2.44)
\definecolor{color409b}{rgb}{0.6,0.6,0.6}
\pscircle[linewidth=0.02,dimen=outer](2.2209375,-0.08){0.3}
\usefont{T1}{ptm}{m}{n}
\rput(2.3123438,0.47){$\varepsilon^{d}$}
\psline[linewidth=0.02cm,arrowsize=0.05291667cm 2.0,arrowlength=1.4,arrowinset=0.4]{->}(2.5009375,-0.08)(3.5009375,-0.08)
\pscircle[linewidth=0.02,dimen=outer](3.8009374,-0.08){0.3}
\usefont{T1}{ptm}{m}{n}
\rput(3.8065624,-0.6821875){\scriptsize $\vartheta^{d}$}
\pscircle[linewidth=0.02,dimen=outer](5.3209376,-0.08){0.3}
\usefont{T1}{ptm}{m}{n}
\rput(5.322344,-0.69){$z$}
\pscircle[linewidth=0.02,dimen=outer,fillstyle=solid,fillcolor=color409b](6.9009376,-0.08){0.3}
\usefont{T1}{ptm}{m}{n}
\rput(6.952344,-0.69){$w$}
\psline[linewidth=0.02cm,arrowsize=0.05291667cm 2.0,arrowlength=1.4,arrowinset=0.4]{->}(4.1009374,-0.08)(5.0209374,-0.08)
\psline[linewidth=0.02cm,arrowsize=0.05291667cm 2.0,arrowlength=1.4,arrowinset=0.4]{->}(5.6209373,-0.08)(6.6209373,-0.08)
\psframe[linewidth=0.02,dimen=outer](7.6609373,0.41)(4.7309375,-1.08)
\usefont{T1}{ptm}{m}{n}
\rput(7.4446874,-0.935){\tiny $N$}
\psframe[linewidth=0.02,dimen=outer](8.020938,0.73)(1.2909375,-2.44)
\usefont{T1}{ptm}{m}{n}
\rput(7.8046875,-2.295){\tiny $D$}
\pscircle[linewidth=0.02,dimen=outer,fillstyle=solid,fillcolor=color409b](2.2209375,-1.4){0.3}
\psframe[linewidth=0.02,dimen=outer](2.8609376,-0.91)(1.6909375,-2.28)
\usefont{T1}{ptm}{m}{n}
\rput(2.6746874,-2.095){\tiny $L$}
\usefont{T1}{ptm}{m}{n}
\rput(2.2265625,-1.9375){\scriptsize $t$}
\pscircle[linewidth=0.02,dimen=outer](0.4009375,-1.4){0.3}
\usefont{T1}{ptm}{m}{n}
\rput(0.40234375,-1.97){$\eta$}
\psline[linewidth=0.02cm,arrowsize=0.05291667cm 2.0,arrowlength=1.4,arrowinset=0.4]{->}(0.7009375,-1.4)(1.9409375,-1.4)
\pscircle[linewidth=0.02,dimen=outer](3.8009374,1.6){0.3}
\usefont{T1}{ptm}{m}{n}
\rput(3.8223438,2.11){$\theta$}
\psframe[linewidth=0.02,dimen=outer](4.9409375,2.44)(3.2909374,1.04)
\psline[linewidth=0.02cm,arrowsize=0.05291667cm 2.0,arrowlength=1.4,arrowinset=0.4]{->}(3.8009374,1.32)(3.8009374,0.2)
\usefont{T1}{ptm}{m}{n}
\rput(4.5346875,1.185){\tiny $L \times K$}
\pscircle[linewidth=0.02,dimen=outer](6.9009376,1.6){0.3}
\psframe[linewidth=0.02,dimen=outer](8.020938,2.44)(6.3709373,1.04)
\usefont{T1}{ptm}{m}{n}
\rput(6.952344,2.11){$\psi$}
\psline[linewidth=0.02cm,arrowsize=0.05291667cm 2.0,arrowlength=1.4,arrowinset=0.4]{->}(6.9009376,1.32)(6.9009376,0.2)
\usefont{T1}{ptm}{m}{n}
\rput(7.6146874,1.185){\tiny $K \times V$}
\pscircle[linewidth=0.02,dimen=outer](0.4009375,-0.08){0.3}
\usefont{T1}{ptm}{m}{n}
\rput(0.44234374,-0.640625){$\pi$}
\psline[linewidth=0.02cm,arrowsize=0.05291667cm 2.0,arrowlength=1.4,arrowinset=0.4]{->}(0.7009375,-0.08)(1.9409375,-0.08)
\psline[linewidth=0.02cm,arrowsize=0.05291667cm 2.0,arrowlength=1.4,arrowinset=0.4]{->}(2.2209375,-1.12)(2.2209375,-0.36)
\end{pspicture} 
}
\end{center}
\caption{Graphical model representation for TWTM, where $\theta$ is distribution matrix of the whole tags, $\psi$ is distribution matrix of words, $\epsilon^d$ represents the weight vector of the tags, and $\vartheta^d$ indicates the topic components for each document. $\pi$ is a Dirichlet prior and $\eta$ is a Bernoulli prior. }
\label{fig:twtm}
\end{figure}
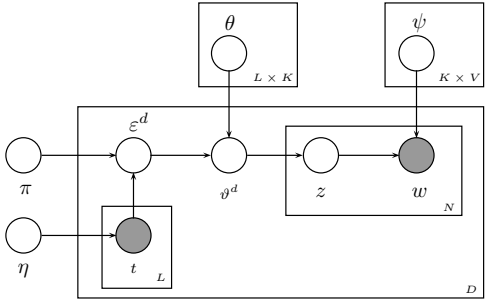

In this paper, we use $\vartheta^{d}$, instead of $\theta^{d}$, to denote the topic distribution of document $d$ as shown in Figure~\ref{fig:twtm}. Let $\theta$ represent a $L \times K$ topic distribution matrix over the tag set, where $K$ is the number of topics. 
Let $\psi$ represent a $K \times V$ distribution matrix over words in the dictionary, where $V$ is the number of words in the dictionary of $D$. Similar to LDA, TWTM models the document $d$ as a mixture of underlying topics and generates each word from one topic.
The topic proportions $\vartheta^{d}$ of the document $d$ is a mixture of tag-topic distributions, not only controlled by a hyperparameter described as in LDA.

The generative process for TWTM is given in the following procedure:
{\small
\begin{enumerate}\setlength{\itemsep}{8pt}
	\item For each topic $k \in \{1,\ldots,K\}$, draw $\psi_k \sim$ Dir($\beta$) , where $\beta$ is a $V$ dimensional prior vector of $\psi$.
	\item For each tag $t \in \{1,\ldots,L\}$, draw $\theta_t \sim$ Dir($\alpha$), where $\alpha$ is a $K$ dimensional prior vector of $\theta$.
	\item  \label{enu:3} For each document $d$: 
	\begin{enumerate}\setlength{\itemsep}{6pt}
			\item For each $l \in \{1,\ldots, L\}$, draw ${t}^{d}_l \sim$~Bernoulli($\eta_l$).
			\item Generate  $T^{d}$ by $\mathbf{t}^{d}$.
			\item Draw $\varepsilon^{d} \sim Dir( T^{d} \times \pi)$.
			\item Generate $\vartheta^{d} = (\varepsilon^{d})^\mathrm{T} \times ( T^{d} \times \theta) $.
			\item For each word $w_{di}$:	
				\begin{enumerate}\setlength{\itemsep}{4pt}
					\item Draw $z_{di} \sim $Mult($\vartheta^{d}$).
					\item Draw $w_{di} \sim $Mult($\psi_{z_{di}}$).
				\end{enumerate}
	\end{enumerate}  
\end{enumerate}
}
In this process, Dir($\cdot$) designates a Dirichlet distribution, Mult($\cdot$) is a multinomial distribution, and $\pi$ is a $L \times 1$ column vector, a Dirichlet prior. Note that $\varepsilon^{d}$ indicates the weight vector of the observed tags in constituting the topic proportions of the document $d$, and $(\varepsilon^{d})^\mathrm{T}$ is the transpose of $\varepsilon^{d}$. Furthermore, $\varepsilon^{d}$ is drawn from a Dirichlet prior which obtained by the matrix multiplication of $T^{d} \times \pi$. Clearly, the result of $T^{d} \times \pi$ will be a ($l^{d} \times 1$) vector whose dimension is depended on the number of the observed tags in the document $d$.

In Step~\ref{enu:3}, for one document $d$, we first generate the document's tags ${t}^{d}_l$ using a Bernoulli coin toss with a prior probability $\eta_{l}$, as shown in step (a). After draw out the $\varepsilon^{d}$, we generate the $\vartheta^{d}$ through $\varepsilon^{d}$, $T^{d}$ and $\theta$. The remaining part of the generative process is just familiar with LDA \cite{DBLP:journals/jmlr/BleiNJ03}. As shown above, in TWTM, we introduce a novel way to model the topic proportions of semi-structured document by document-special tags and text data. The key discussed in this paper is the tag's weight topic assignment by which $\vartheta^{d}$ is generated through $\varepsilon^{d}$, $T^{d}$, and $\theta$, which provides an effective and direct method to infer the weights of the tags.
\subsection{Tag-Weighted Topic Assignment}
As we assume that all the observed tags in the document $d$ make contributions to infer the topic distribution $\vartheta^{d}$ of the document, it is expected that different tags works corresponding to their own weights. For example, in some blog application, a blog has tags of an author, a blog's date, a blog category and a blog's url. Clearly, compared to other tags, the tag of the author plays the most important role in constituting the topic components of the blog.

The function of how to leverage the tag information or contextual to infer topic distribution of a document is defined as follows:
{\footnotesize
\begin{equation*}
\begin{aligned}
\vartheta \longleftarrow f(t_1, \cdots, t_l),
\end{aligned}
\end{equation*}
}
where $f(\cdot)$ is the way of making use of the tag information.
Topic models using tag information or contextual take advantage of the different $f(\cdot)$ in the past.
In TWTM, we assume that  $\vartheta^{d}$ is made up by all the observed tags with their own weights. 
Figure~\ref{fig:twtm} shows that how TWTM works in a probabilistic graphical model. 
As shown in Figure~\ref{fig:twtm}, $\vartheta^{d}$ is controlled by two sides, the topic distributions over tags $\theta$, and the weights of the given tags of the document $d$. It is important to distinguish TWTM from the Author-Topic Model  \cite{DBLP:conf/uai/Rosen-ZviGSS04}. In the author-topic model, the words $w$ is chose only by one of the given tags' distribution, while in TWTM, for word $w$, all the observed tags in the document would make the contributions. 

The $f(\cdot)$ in the proposed model is assumed as this, for the document $d$,
{\footnotesize
\begin{equation*}
\begin{aligned}
&f(\vartheta^{d}) = (\varepsilon^{d})^\mathrm{T} \times T^{d} \times \theta,
\end{aligned}
\end{equation*}
}
where the linear multiplication of $(\varepsilon^{d})^\mathrm{T}$, $T^{d}$ and $\theta$ maintains the condition of $\sum_{k=1}^K \vartheta_k^{d} = 1$ without normalization of $\vartheta^{d}$, since $\varepsilon^{d}$ and $\theta$ satisfy
{\footnotesize
\begin{equation*}
\begin{aligned}
&\sum_{i=1}^{l^{d}} \varepsilon_i^{d} = 1, \sum_{k=1}^K \theta_{lk} =1.
\end{aligned}
\end{equation*}
}
Firstly, we pick out the topic distributions of the given tags in the document $d$ from $\theta$ by $T^{d} \times \theta$, where $T^{d}$ is a $l^{d} \times L$ matrix and $\theta$ is a $L \times K$ matrix. Here we define
{\footnotesize
\begin{equation*}
\begin{aligned}
\Theta^{d} = T^{d} \times \theta,
\end{aligned}
\end{equation*}
}
where the $\Theta^{d}$ is a $l^{d} \times K$ topic distribution matrix of the given tags in $d$ as sub-components of $\theta$. Secondly, $\varepsilon^{d}$ is the weight vector of the observed tags in $d$, and each dimension of $\varepsilon^{d}$ represents the weight or importance associated to the corresponding tag. Thus, $\vartheta^{d}$ is mixed by $\Theta^{d}$ with corresponding weight values. 
{\footnotesize
\begin{equation*}
\begin{aligned}
&\vartheta^{d} =(\sum^{l^{d}}_{i=1}{\varepsilon^{d}_i \Theta^{d}_{i1}}, \ldots, \sum^{l^{d}}_{i=1}{\varepsilon^{d}_i \Theta^{d}_{ij}}, \ldots, \sum^{l^{d}}_{i=1}{\varepsilon^{d}_i \Theta^{d}_{iK}}).
\end{aligned}
\end{equation*}
}
With $\vartheta^{d}$, TWTM generates all the words in the document $d$ with the assumption of bag-of-words.

Based on the above framework, we can define a special topic assignment function $f(\cdot)$ in an extended model for a real world application. 
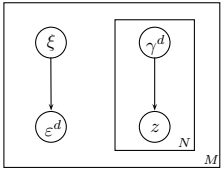
\begin{figure}[t]
\begin{center}
\psscalebox{0.7 0.7} 
{
\begin{pspicture}(0,-1.56)(5.0410156,1.56)
\pscircle[linecolor=black, linewidth=0.02, dimen=outer](0.89960927,-0.7866667){0.3}
\rput[bl](0.7810157,-0.955){\normalsize{$\varepsilon^d$}}
\pscircle[linecolor=black, linewidth=0.02, dimen=outer](0.89960927,0.8){0.3}
\psline[linecolor=black, linewidth=0.02, arrowsize=0.05291666666666668cm 2.0,arrowlength=1.4,arrowinset=0.4]{->}(0.8970011,0.5199932)(0.90221745,-0.4799932)
\rput[bl](0.82,0.64){$\xi$}
\pscircle[linecolor=black, linewidth=0.02, dimen=outer](2.8596094,-0.7866667){0.3}
\pscircle[linecolor=black, linewidth=0.02, dimen=outer](2.8596094,0.8){0.3}
\psline[linecolor=black, linewidth=0.02, arrowsize=0.05291666666666668cm 2.0,arrowlength=1.4,arrowinset=0.4]{->}(2.8596094,0.52)(2.8596094,-0.48)
\rput[bl](2.7,0.58){$\gamma^d$}
\rput[bl](2.78,-0.88){$z$}
\psframe[linecolor=black, linewidth=0.02, dimen=outer](3.62,1.24)(2.1,-1.24)
\rput[bl](3.3,-1.16){$_N$}
\psframe[linecolor=black, linewidth=0.02, dimen=outer](4.1,1.56)(0.0,-1.56)
\rput[bl](3.78,-1.48){$_M$}
\end{pspicture}
}
\end{center}
\caption{Graphical model representation of the variational distribution used to approximate the posterior in TWTM.}
\label{fig:twtm_vi}
\end{figure}

\subsection{Inference for TWTM}
In the topic models, the key inferential problem that we need to solve is to compute the posterior distribution of the hidden variables given a document $d$.
Given the document $d$, we can easily get the posterior distribution of the latent variables in the proposed model, as:
{\footnotesize
\begin{equation}
\begin{aligned}
p(\varepsilon^{d}, \mathbf{z} | \mathbf{w}^{d}, T^{d}, \theta, \eta,\psi,\pi)
= \frac{p(\varepsilon^{d}, \mathbf{z}, \mathbf{w}^{d}, T^{d} | \theta, \eta,\psi,\pi)}{p(\mathbf{w}^{d}, T^{d} | \theta, \eta,\psi,\pi)}.\label{eq:twtm_pd}
\end{aligned}
\end{equation}
}
In Eq.~(\ref{eq:twtm_pd}), integrating over $\varepsilon$ and summing out $z$, we easily obtain the marginal distribution of $d$:
{\footnotesize
\begin{equation*}
\begin{aligned}
p(\mathbf{w}^{d}, T^{d} &| \eta,\theta,\psi,\pi) = p(\mathbf{t}^{d} | \eta) \int p\left (\varepsilon^{d} | (T^{d} \times \pi) \right ) \\
& \cdot \prod_{i=1}^N \sum_{z^{d}_i=1}^K p(z^{d}_i | (\varepsilon^{d})^\mathrm{T} \times T^{d} \times \theta)
  p(w^{d}_i | z^{d}_i,\psi_{1:K}) ~d\varepsilon^{d}. 
\end{aligned}
\end{equation*}
}

In this work, we make use of mean-field variational EM algorithm  \cite{DBLP:journals/jei/BishopN07} to efficiently obtain an approximation of this posterior distribution of the latent variables. In the mean-field variational inference, we minimize the KL divergence between the variational posterior probability and the true posterior probability through by maximizing the evidence lower bound (ELBO) $\mathcal{L}(\cdot)$ \cite{DBLP:conf/nips/BleiM07}. For a single document $d$, we obtain the $\mathcal{L}(\cdot)$ using Jensen's inequality:
{\footnotesize
\begin{equation*}
\begin{aligned}
\mathcal{L}(\xi_{1:l^{d}}, \gamma_{1:K}; \eta_{1:L},& \pi_{1:L}, \theta_{1:L}, \psi_{1:K}) \\
 & = E[\log p(T_{1:l^{d}} | \eta_{1:L})] + E[\log p(\varepsilon^{d}| T^{d} \times \pi)]  \\
 & +\sum_{i=1}^N E[\log p(z_i | (\varepsilon^{d})^\mathrm{T} \times T^{d} \times \theta)] \\
 &+\sum_{i=1}^N E[\log p(w_i | z_i, \psi_{1:K})] + H(q),
\end{aligned}
\end{equation*}
}
where $\xi$ is a $l^{d}$-dimensional Dirichlet parameter vector and $\gamma$ is $1 \times K $ vector, both of which are variational parameters of variational distribution shown in Figure~\ref{fig:twtm_vi}, and $H(q)$ indicates the entropy of the variational distribution:
{\footnotesize
\begin{equation*}
\begin{aligned}
H(q) = - E[\log q(\varepsilon^{d})] - E[\log q(z)].
\end{aligned}
\end{equation*}
}
Here the exception is taken with respect to a variational distribution $q(\varepsilon^{d}, z_{1:N})$, and we choose the following fully factorized distribution: 
{\footnotesize
\begin{equation*}
\begin{aligned}
&q(\varepsilon^{d}, z_{1:N} | \xi_{1:L}, \gamma_{1:K} ) = q( \varepsilon^{d} | \xi ) \prod_{i=1}^N q(z_i | \gamma_i).
\end{aligned}
\end{equation*}
}
The dimension of parameter $\xi$ is changed with different documents. It could be difficult to compute the expected log probability of a topic assignment by the way of tag-weighted topic assignment used in TWTM.

Then, we maximize the lower bound $\mathcal{L}(\cdot)$ with respect to the variational parameters $\xi$ and $\gamma$, using a variational expectation-maximization(EM) procedure as follows.
\subsubsection{Variational E-step}
We first maximize $\mathcal{L}(\cdot)$ with respect to $\xi_i$ for the document $d$. Maximize the terms which contain $\xi$:
{\footnotesize
\begin{equation}
\begin{aligned}
\mathcal{L}_{[\xi]} &= 
\sum_{i=1}^{l^{d}}( \sum_{l^{'}=1}^L \pi_{l^{'}} T_{i{l^{'}}}^{d} - 1)(\Psi(\xi_i) - \Psi(\sum_{j^{'}=1}^{l^{d}} \xi_{j^{'}})) \\
& +\sum_{i=1}^N \sum_{k=1}^K \gamma_{ik} \cdot \sum_{j=1}^{l^{d}} \log \theta_k^{(j)} {\xi_j}/{ \sum_{j^{'}=1}^{l^{d}} \xi_{j^{'}} } \\
&- \log \Gamma(\sum_{i=1}^{l^{d}}{\xi_i}) + \sum_{i=1}^{l^{d}} \log \Gamma(\xi_i) \\
&- \sum_{i=1}^{l^{d}} (\xi_i - 1) ( \Psi(\xi_i) - \Psi(\sum_{j^{'}=1}^{l^{d}} \xi_{j^{'}})), \label{eq:twtm_xi}
\end{aligned}
\end{equation}
}
where $\Psi(\cdot)$ denotes the digamma function, the first derivative of the log of the Gamma function. Here we use gradient descent method to find the $\xi$ to make the maximization of $\mathcal{L}_{[\xi]}$.

Next, we maximize $\mathcal{L}(\cdot)$ with respect to $\gamma_{ik}$. Adding the Lagrange multipliers to the terms which contain $\gamma_{ik}$, taking the derivative with respect to $\gamma_{ik}$, and setting the derivative to zero yields, we obtain the update equation of $\gamma_{ik}$:
{\footnotesize
\begin{equation}
\begin{aligned}
&\gamma_{ik} \propto \psi_{k,v^{w_i}} \exp\{ \sum_{j=1}^{l^{d}} \log \theta_k^{(j)} \frac{\xi_j}{ \sum_{j'=1}^{l^{d}} \xi_{j'} } \}, \label{eq:twtm_gamma}
\end{aligned}
\end{equation}
}
where $v^{w_i}$ denotes the index of $w_i$ in the dictionary.

In E-step, we update the $\xi$ and $\gamma$ for each document with the initialized model parameters. For the reason of different document with different number of tags, we have to keep all the $\xi$ updated by each document for the M-step estimation.
\subsubsection{M-step estimation}
The M-step needs to update four parameters: $\eta$, the tagging prior probability, $\pi$, the Dirichlet prior of the tags' weights, $\theta$, the topic distribution over all tags in the corpus, and $\psi$, the probability of a word under a topic. Because each document's tag-set is observed, the Bernoulli prior $\eta$ is unused included for model completeness. For a given corpus, the $\eta_i$ is estimated by adding up the number of $i$-th tag which appears in the corpus.

For the document $d$, the terms that involve the Dirichlet prior $\pi$:
{\footnotesize
\begin{equation}
\begin{aligned}
\mathcal{L}_{[\pi]} &=
\log \Gamma \left(\sum_{i=1}^{l^{d}}{(T^{d} \times \pi)}_i \right) - \sum_{i=1}^{l^{d}}\log \Gamma \left({(T^{d} \times \pi)}_i \right) \\
 &+ \sum_{i=1}^{l^{d}} \left({(T^{d} \times \pi)}_i - 1 \right) \left(\Psi(\xi_i) - \Psi(\sum_{j=1}^{l^{d}} \xi_j) \right),\label{eq:twtm_pi}
\end{aligned}
\end{equation}
}
where ${(T^{d} \times \pi)}_i$ = $\sum_{i=1}^{l^{d}} \sum_{l=1}^L \pi_l T_{il}^{d}$.
We use gradient descent method by taking derivative of Eq.~(\ref{eq:twtm_pi}) with respect to $\pi_l$ on the corpus to find the estimation of $\pi$.

To maximize with respect to $\theta$ and $\psi$, we obtain the following update equations:
{\footnotesize
\begin{equation}
\begin{aligned}
\theta_{lk} & \propto \sum_{d=1}^D \sum_{i=1}^N \gamma_{ik}^{d} \frac{\xi_l^{d} {t}^{d}_l}{\sum_{l=1}^L ( \xi_l^{d} {t}^{d}_l )},\label{eq:twtm_theta}
\end{aligned}
\end{equation}
} 
and 
{\footnotesize
\begin{equation}
\begin{aligned}
\hspace{-10mm}\psi_{kj} & \propto \sum_{d=1}^D \sum_{i=1}^N \gamma_{ik}^{d} ({w^{d}})_i^{j}.\label{eq:twtm_psi}
\end{aligned}
\end{equation}
}

We provide a detailed derivation of the variational EM algorithm for TWTM in Appendix A. And we show the variational expectation maximization (EM) procedure of TWTM in Algorithm~\ref{table:twtm_em}.
\begin{algorithm}
\caption{The variational expectation maximization (EM) algorithm of TWTM}
\label{table:twtm_em}
\begin{algorithmic}[1]
\STATE{\textbf{Input:} a semi-structured corpora including totally V unique words, L unique tags, and the expected number K of topics.}
\STATE{\textbf{Output:} Topic-word distributions $\psi$, Tag-topic distributions $\theta$, $\pi$, topic distribution $\vartheta^d$ and weight vector $\varepsilon^{d}$ of each training document.}
\STATE{initialize $\pi$, and initialize $\theta$ and $\psi$ with the constraint of $\sum_{k=1}^K \theta_{lk}$ equals 1 and $\sum_{i=1}^V \psi_{ki}$ equals 1.}
\REPEAT 
\FOR{ each document $d$}
\STATE{update $\xi^d$ with Eq.~(\ref{eq:twtm_xi}) using gradient descent method.}
\STATE{update $\gamma_{ik}$ with Eq.~(\ref{eq:twtm_gamma}).}
\ENDFOR
\STATE{update $\pi$ with Eq.~(\ref{eq:twtm_pi}) using gradient descent method.}
\STATE{update $\theta$ by Eq.~(\ref{eq:twtm_theta}).}
\STATE{update $\psi$ by Eq.~(\ref{eq:twtm_psi}).}

\UNTIL{convergence}
\end{algorithmic}
\end{algorithm}

\section{Tag-Weighted Dirichlet Allocation}\label{sec:TWDA}
In a real world application, a corpus is very likely to contain both semi-structured documents and unstructured documents. Many documents in the corpus have no tags, just with unstructured text data. In this case, TWTM does not work, which generates the topic distribution of a document by leveraging the weights among the observed tags. Our proposed solution to the problem is to add a Dirichlet prior to the topic distribution $\vartheta^{d}$, which means that we learn the weights among the Dirichlet prior and the given tags, not just among the tags. We call this solution Tag-Weighted Dirichlet Allocation (TWDA). When handling the unstructured documents in a hybrid corpus, TWDA degenerates into LDA \cite{DBLP:journals/jmlr/BleiNJ03} which just draws the topic proportions for a document from a Dirichlet distribution.

As an extended model of TWTM, TWDA uses the same parameter notations. Unlike TWTM, for the convenience of the inference in TWDA, $\mathbf{t}^{d}$ is expanded to a $l^{d} \times (L+1) $ matrix $T^{d}$, where $l^d$ is one more than the number of the given tags in the document $d$ (For example, if the document $d$ has five tags, $l^d$ is six). For each row number $i \in \{1,\ldots,l^{d}\}$ in $T^{d}$, $T^{d}_{i\cdot}$ is a binary vector, where $T^{d}_{ij} = 1 $ if and only if the $i$-th tag of the document $d$ is the $j$-th tag of the tag set in the corpus $D$. Note that, we set the last dimension of the last row in $T^{d}$ to $1$, and the other dimensions of the last row equal to 0 for all documents. The detail of the above setting will be shown later.
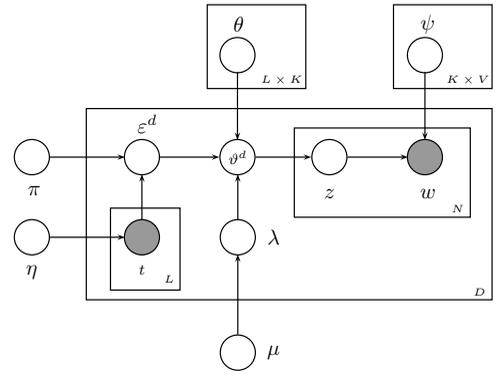
\begin{figure}[t]
\begin{center}
\scalebox{0.8} 
{
\begin{pspicture}(0,-3.44)(8.26,2.44)
\definecolor{color409b}{rgb}{0.6,0.6,0.6}
\pscircle[linewidth=0.02,dimen=outer](2.2209375,-0.08){0.3}
\usefont{T1}{ptm}{m}{n}
\rput(2.3123438,0.47){$\varepsilon^{d}$}
\psline[linewidth=0.02cm,arrowsize=0.05291667cm 2.0,arrowlength=1.4,arrowinset=0.4]{->}(2.5009375,-0.08)(3.5009375,-0.08)
\pscircle[linewidth=0.02,dimen=outer](3.8009374,-0.08){0.3}
\usefont{T1}{ptm}{m}{n}
\rput(3.8065624,-0.08){\scriptsize $\vartheta^{d}$}
\pscircle[linewidth=0.02,dimen=outer](5.3209376,-0.08){0.3}
\usefont{T1}{ptm}{m}{n}
\rput(4.4,-1.4){$\lambda$}
\usefont{T1}{ptm}{m}{n}
\rput(4.4,-3.3){$\mu$}
\pscircle[linewidth=0.02,dimen=outer](3.8065624,-1.4){0.3}
\psline[linewidth=0.02cm,arrowsize=0.05291667cm 2.0,arrowlength=1.4,arrowinset=0.4]{->}(3.8065624,-1.12)(3.8065624,-0.36)
\psline[linewidth=0.02cm,arrowsize=0.05291667cm 2.0,arrowlength=1.4,arrowinset=0.4]{->}(3.8065624,-3)(3.8065624,-1.6801)
\pscircle[linewidth=0.02,dimen=outer](3.8065624,-3.3){0.3}
\usefont{T1}{ptm}{m}{n}
\rput(5.322344,-0.69){$z$}
\pscircle[linewidth=0.02,dimen=outer,fillstyle=solid,fillcolor=color409b](6.9009376,-0.08){0.3}
\usefont{T1}{ptm}{m}{n}
\rput(6.952344,-0.69){$w$}
\psline[linewidth=0.02cm,arrowsize=0.05291667cm 2.0,arrowlength=1.4,arrowinset=0.4]{->}(4.1009374,-0.08)(5.0209374,-0.08)
\psline[linewidth=0.02cm,arrowsize=0.05291667cm 2.0,arrowlength=1.4,arrowinset=0.4]{->}(5.6209373,-0.08)(6.6209373,-0.08)
\psframe[linewidth=0.02,dimen=outer](7.6609373,0.41)(4.7309375,-1.08)
\usefont{T1}{ptm}{m}{n}
\rput(7.4446874,-0.935){\tiny $N$}
\psframe[linewidth=0.02,dimen=outer](8.020938,0.73)(1.2909375,-2.44)
\usefont{T1}{ptm}{m}{n}
\rput(7.8046875,-2.295){\tiny $D$}
\pscircle[linewidth=0.02,dimen=outer,fillstyle=solid,fillcolor=color409b](2.2209375,-1.4){0.3}
\psframe[linewidth=0.02,dimen=outer](2.8609376,-0.91)(1.6909375,-2.28)
\usefont{T1}{ptm}{m}{n}
\rput(2.6746874,-2.095){\tiny $L$}
\usefont{T1}{ptm}{m}{n}
\rput(2.2265625,-1.9375){\scriptsize $t$}
\pscircle[linewidth=0.02,dimen=outer](0.4009375,-1.4){0.3}
\usefont{T1}{ptm}{m}{n}
\rput(0.40234375,-1.97){$\eta$}
\psline[linewidth=0.02cm,arrowsize=0.05291667cm 2.0,arrowlength=1.4,arrowinset=0.4]{->}(0.7009375,-1.4)(1.9409375,-1.4)
\pscircle[linewidth=0.02,dimen=outer](3.8009374,1.6){0.3}
\usefont{T1}{ptm}{m}{n}
\rput(3.8223438,2.11){$\theta$}
\psframe[linewidth=0.02,dimen=outer](4.9409375,2.44)(3.2909374,1.04)
\psline[linewidth=0.02cm,arrowsize=0.05291667cm 2.0,arrowlength=1.4,arrowinset=0.4]{->}(3.8009374,1.32)(3.8009374,0.2)
\usefont{T1}{ptm}{m}{n}
\rput(4.5346875,1.185){\tiny $L \times K$}
\pscircle[linewidth=0.02,dimen=outer](6.9009376,1.6){0.3}
\psframe[linewidth=0.02,dimen=outer](8.020938,2.44)(6.3709373,1.04)
\usefont{T1}{ptm}{m}{n}
\rput(6.952344,2.11){$\psi$}
\psline[linewidth=0.02cm,arrowsize=0.05291667cm 2.0,arrowlength=1.4,arrowinset=0.4]{->}(6.9009376,1.32)(6.9009376,0.2)
\usefont{T1}{ptm}{m}{n}
\rput(7.6146874,1.185){\tiny $K \times V$}
\pscircle[linewidth=0.02,dimen=outer](0.4009375,-0.08){0.3}
\usefont{T1}{ptm}{m}{n}
\rput(0.44234374,-0.640625){$\pi$}
\psline[linewidth=0.02cm,arrowsize=0.05291667cm 2.0,arrowlength=1.4,arrowinset=0.4]{->}(0.7009375,-0.08)(1.9409375,-0.08)
\psline[linewidth=0.02cm,arrowsize=0.05291667cm 2.0,arrowlength=1.4,arrowinset=0.4]{->}(2.2209375,-1.12)(2.2209375,-0.36)
\end{pspicture} 
}
\end{center}
\caption{Graphical model representation for TWDA, where $\mu$ is a Dirichlet prior of $\lambda$.}
\label{fig:twda}
\end{figure}

TWDA defines a Dirichlet prior $\mu$ over a latent topic distribution of a document, and mixes the latent topic proportion with these topic distributions of the given tags by importance or weight (tag-weighted) to form the final topic distribution of the document. Figure~\ref{fig:twda} shows the graphical model representation of TWDA, and the generative process for TWDA is given in the following procedure:
{\small
\begin{enumerate}\setlength{\itemsep}{8pt}
	\item For each topic $k \in \{1,\ldots,K\}$, draw $\psi_k \sim$ Dir($\beta$) , where $\beta$ is a $V$ dimensional prior vector of $\psi$.
	\item For each tag $t \in \{1,\ldots,L\}$, draw $\theta_t \sim$ Dir($\alpha$), where $\alpha$ is a $K$ dimensional prior vector of $\theta$.
	\item For each document $d$: 
	\begin{enumerate}\setlength{\itemsep}{6pt}
		  \item Draw $\lambda \sim Dir(\mu)$.
			\item Generate  $T^{d}$ by $\mathbf{t}^{d}$.
			\item Draw $\varepsilon^{d} \sim Dir( T^{d} \times \pi)$.
			\item Generate $\vartheta^{d} = (\varepsilon^{d})^\mathrm{T} \times T^{d} \times (\frac{ \theta}{\lambda}) $ .
			\item For each word $w_{di}$:	
				\begin{enumerate}\setlength{\itemsep}{4pt}
					\item Draw $z_{di} \sim $Mult($\vartheta^{d}$) .
					\item Draw $w_{di} \sim $Mult($\psi_{z_{di}}$) .
				\end{enumerate}
	\end{enumerate}
\end{enumerate}
}
Note that, $L$ is the number of tags appeared in the corpora and $K$ is the number of topics. Different from TWTM, here $\pi$ is a $(L+1) \times 1$ column vector and $\mu$ is a $K \times 1$ column vector. Both of them are Dirichlet prior.
 $\lambda$ is a $1 \times K$ row vector which is drawn from $\mu$. $(\varepsilon^{d})^\mathrm{T}$ is the transpose of $\varepsilon^{d}$, and $\varepsilon^{d}$ is drawn from a Dirichlet prior which obtained by the matrix multiplication of $T^{d} \times \pi$. Clearly, the result of $T^{d} \times \pi$ will be a ($l^{d} \times 1$) vector whose dimension is depended on the number of the observed tags in the document $d$. Note that, $l^d$ is one more than the number of tags given in $d$ as we described above. 

In other words, we treat the $\lambda$ as a topic distribution of one latent tag, the Dirichlet prior $\mu$. Each document is controlled by a latent tag, that is the same idea both TWDA and Latent Dirichlet Allocation (LDA). The form of $(\frac{ \theta}{\lambda})$ is the augmented matrix of $\theta$ and $\lambda$, which represents that we add the vector $\lambda$ to the matrix $\theta$ as the last row, so $(\frac{ \theta}{\lambda})$ becomes a $(L+1) \times K$ matrix. As we show above, $T^{d}$ is the matrix form of the given tags in the document $d$,  and the last row of $T^{d}$ is a binary vector, of which only the last dimension equals to 1 and the others equal 0. 
Here we define 
{\footnotesize
\begin{equation*}
\begin{aligned}
\Theta^{d} = T^{d} \times (\frac{ \theta}{\lambda}).
\end{aligned}
\end{equation*}
}

Clearly, $\Theta^{d}$ is a $l^d \times K$ matrix, whose last row is $\lambda$.
Actually, the purpose of $\Theta^{d}$  is to pick out the rows corresponded to the tags appeared in $d$ from tag-topic distribution matrix $\theta$.

The key idea of tag-weighted Dirichlet allocation is to model the topic proportions of semi-structured documents by document-special tags and text data. Different from LDA, the topic proportion of one document assumed in this model is controlled not only by a Dirichlet prior $\mu$, but also by all the observed tags. The way to generate the normalized topic distribution of the document $d$ is that we mix both Dirichlet allocation and tags information through a weight vector $\varepsilon^{d}$. Thus, we use the function $f(\cdot)$ of topic assignment to obtain the topic distribution of $d$ by
{\footnotesize
\begin{equation*}
\begin{aligned}
f(\vartheta^{d}) = (\varepsilon^{d})^\mathrm{T} \times T^{d} \times (\frac{ \theta}{\lambda}).
\end{aligned}
\end{equation*}
}

It is worth to note that the $\varepsilon^{d}$ is draw by a Dirichlet prior $\pi$, each row of $\theta$ is draw by a Dirichlet prior $\alpha$, and $\lambda$ is draw by a Dirichlet prior $\mu$, so $\varepsilon^{d}$ and $\theta$ satisfy
{\footnotesize
\begin{equation*}
\begin{aligned}
&\sum_{i=1}^{l^{d}} \varepsilon_i^{d} = 1 , \sum_{k=1}^K \theta_{lk} =1, \text{and }\sum_{k=1}^K \lambda_k =1.
\end{aligned}
\end{equation*}
}

Therefore, the linear multiplication of $(\varepsilon^{d})^\mathrm{T}$, $T^{d}$, $\theta$ and $\lambda$ maintains the condition of $\sum_{k=1}^K \vartheta_k^{d} = 1$ without normalization of $\vartheta^{d}$. With $\vartheta^{d}$, the topic proportions of the document $d$, the remaining part of the generative process is just familiar with LDA.
\begin{figure}[t]
\begin{center}
\psscalebox{0.7 0.7} 
{
\begin{pspicture}(0,-1.56)(6.14,1.56)
\pscircle[linecolor=black, linewidth=0.02, dimen=outer](0.59960926,-0.7866667){0.3}
\rput[bl](0.48101568,-0.955){\normalsize{$\varepsilon^d$}}
\pscircle[linecolor=black, linewidth=0.02, dimen=outer](0.59960926,0.8){0.3}
\psline[linecolor=black, linewidth=0.02, arrowsize=0.05291666666666668cm 2.0,arrowlength=1.4,arrowinset=0.4]{->}(0.59960926,0.52)(0.59960926,-0.48)
\rput[bl](0.52,0.64){$\xi$}
\pscircle[linecolor=black, linewidth=0.02, dimen=outer](2.4796093,-0.7866667){0.3}
\pscircle[linecolor=black, linewidth=0.02, dimen=outer](2.4796093,0.8){0.3}
\psline[linecolor=black, linewidth=0.02, arrowsize=0.05291666666666668cm 2.0,arrowlength=1.4,arrowinset=0.4]{->}(2.48,0.52)(2.4796093,-0.48)
\pscircle[linecolor=black, linewidth=0.02, dimen=outer](4.359609,-0.7866667){0.3}
\pscircle[linecolor=black, linewidth=0.02, dimen=outer](4.359609,0.8){0.3}
\psline[linecolor=black, linewidth=0.02, arrowsize=0.05291666666666668cm 2.0,arrowlength=1.4,arrowinset=0.4]{->}(4.359609,0.52)(4.359609,-0.48)
\rput[bl](2.4,0.68){$\rho$}
\rput[bl](2.4,-0.92){$\lambda$}
\rput[bl](4.2,0.58){$\gamma^d$}
\rput[bl](4.28,-0.88){$z$}
\psframe[linecolor=black, linewidth=0.02, dimen=outer](5.12,1.24)(3.6,-1.24)
\rput[bl](4.8,-1.16){$_N$}
\psframe[linecolor=black, linewidth=0.02, dimen=outer](5.6,1.56)(0.0,-1.56)
\rput[bl](5.28,-1.48){$_M$}
\end{pspicture}
}
\end{center}
\caption{Graphical model representation of the variational distribution used to approximate the posterior in TWDA.}
\label{fig:twda_vi}
\end{figure}
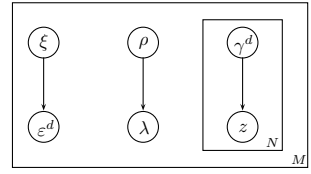

\subsection{Inference for TWDA}
In TWDA, we treat $\pi$, $\mu$, $\eta$, $\theta$ and $\psi$ as unknown constants to be estimated. Similar to TWTM, the marginal distribution of $d$ is not efficiently computable as follows:
{\footnotesize
\begin{equation*}
\begin{aligned}
p(\mathbf{w}^{d}, T^{d} | \eta,\theta,\psi,\pi, \mu) &
= p(\mathbf{t}^{d} | \eta) \int p\left (\varepsilon^{d} | (T^{d} \times \pi) \right ) \\
& \cdot p(\lambda | \mu) \prod_{i=1}^N \sum_{z^{d}_i=1}^K p(z^{d}_i | (\varepsilon^{d})^\mathrm{T} \times T^{d} \times (\frac{ \theta}{\lambda})) \\
& \cdot p(w^{d}_i | z^{d}_i,\psi_{1:K}) ~d\varepsilon^{d}. 
\end{aligned}
\end{equation*}
}
In this case, We also use a variational expectation-maximization (EM) procedure to carry out approximate maximum likelihood estimation of TWDA.
\subsubsection{Variational inference}
In TWDA, we use the following fully factorized distribution as shown in Figure~\ref{fig:twda_vi}: 
{\footnotesize
\begin{equation*}
\begin{aligned}
&q(\varepsilon^{d}, \lambda^{d}, z_{1:N} | \xi_{1:L}, \rho_{1:K}, \gamma_{1:K} ) = q( \varepsilon^{d} | \xi ) q(\lambda^{d} | \rho ) \prod_{i=1}^N q(z_i | \gamma_i),
\end{aligned}
\end{equation*}
}
and the entropy of the variational distribution will be {
\footnotesize
\begin{equation*}
\begin{aligned}
H(q) = - E[\log q(\varepsilon^{d})] - E[\log q(\lambda)] - E[\log q(z)].
\end{aligned}
\end{equation*}
}

For the variational parameter $\xi$, we take the terms which contain $\xi$ out of the evidence lower bound (ELBO) $\mathcal{L}(\cdot)$ of TWDA to form $\mathcal{L}_{[\xi]}$, and we use gradient descent method to find the $\xi$ to make the maximization of $\mathcal{L}_{[\xi]}$:
{\footnotesize
\begin{equation}
\begin{aligned}
\mathcal{L}_{[\xi]} &= 
\sum_{i=1}^{l^{d}}( \sum_{l^{'}=1}^{L+1} \pi_{l^{'}} T_{i{l^{'}}}^{d} - 1)(\Psi(\xi_i) - \Psi(\sum_{j^{'}=1}^{l^{d}} \xi_{j^{'}})) \\
& +\sum_{i=1}^N \sum_{k=1}^K \gamma_{ik} \cdot \sum_{j=1}^{l^{d}} C_k^{(j)} \frac{\xi_j}{ \sum_{j^{'}=1}^{l^{d}} \xi_{j^{'}} } \\
&- \log \Gamma(\sum_{i=1}^{l^{d}}{\xi_i}) + \sum_{i=1}^{l^{d}} \log \Gamma(\xi_i) \\
&- \sum_{i=1}^{l^{d}} (\xi_i - 1) ( \Psi(\xi_i) - \Psi(\sum_{j^{'}=1}^{l^{d}} \xi_{j^{'}})), \label{eq:twda_xi}
\end{aligned}
\end{equation}
}
where 
{\footnotesize
\begin{equation*}
\begin{aligned}
C_k^{(j)}=
\begin{cases}
\log \theta_k^{(j)} & j \in \{ 1, \cdots, l^{d}-1 \}\\
\Psi(\rho_k) - \Psi(\sum_{j^{'}=1}^{K} \rho_{j^{'}}) & j=l^d
\end{cases}, 
\end{aligned}
\end{equation*}
}
and $\Psi(\cdot)$ denotes the digamma function, the first derivative of the log of the Gamma function.

In particular, by computing the derivatives of the $\mathcal{L}(\cdot)$ and setting them equal to zero, we obtain the following pair of update equations for the variational parameters $\rho^d$ and $\gamma_{ik}$:
{\footnotesize
\begin{equation}
\begin{aligned}
\rho_i \propto \mu_i + \sum_{n=1}^N \gamma_{ni} \cdot \frac{\xi_{l^d}}{\sum_{j=1}^{l^d} \xi_j},\label{eq:twda_rho}
\end{aligned}
\end{equation}
}
{\footnotesize
\begin{equation}
\begin{aligned}
&\gamma_{ik} \propto \psi_{k,v^{w_i}} \exp\{ \sum_{j=1}^{l^{d}} C_k^{(j)} \frac{\xi_j}{ \sum_{j'=1}^{l^{d}} \xi_{j'} } \},\label{eq:twda_gamma}
\end{aligned}
\end{equation}
}
where $v^{w_i}$ denotes the index of $w_i$ in the dictionary.

In the E-step, we update the variational parameters $\xi$, $\rho$ and $\gamma$ for each document with the initialized model parameters. We show the detailed derivation of the variational parameters for TWDA in Appendix B.
\subsubsection{Model Parameter Estimation}
There are four model parameters that need to estimate in M-step, $\pi$, the Dirichlet prior of the tags' weights, $\theta$, the topic distribution over all tags in the corpus, $\psi$, the probability of a word under a topic, and $\mu$, a Dirichlet prior of model. In TWDA, we can estimate $\pi$, $\theta$ and $\psi$ as same as in TWTM.

Different from TWTM, TWDA has an extra Dirichlet prior $\mu$. The involved terms of $\mu$ are:
{\footnotesize
\begin{equation}
\begin{aligned}
\mathcal{L}_{[\mu]} &= \sum_{d=1}^D (\log \Gamma(\sum_{j=1}^K \mu_j) - \sum_{i=1}^K \log \Gamma(\mu_i) \\ 
&+ \sum_{i=1}^K(\mu_i -1)(\Psi(\rho_i^d) - \Psi(\sum_{j=1}^K \rho_j^d))).\label{eq:twda_mu}
\end{aligned}
\end{equation}
}
We can invoke the linear-time Newton-Raphson algorithm to estimate $\mu$ as same as the Dirichlet parameter described in LDA \cite{DBLP:journals/jmlr/BleiNJ03}.

In the variational expectation maximization (EM) procedure of TWDA, we update the variational parameters $\xi^d$, $\rho$ and $\gamma_{ik}$ with Eqs.~(\ref{eq:twda_xi}), (\ref{eq:twda_rho}) and (\ref{eq:twda_gamma}) respectively in the E-step. In the M-step, besides the update of $\pi$, $\theta$ and $\psi$, we also update $\mu$ with Eq.~(\ref{eq:twda_mu}) by Newton-Raphson algorithm. The detailed derivation of the model parameter estimation in TWDA is shown in Appendix B.

\section{Analysis of TWDA}\label{sec:TheoreticalAnalysis}
In TWDA, we introduce a better way to directly model the semi-structured documents and unstructured documents by adding a latent tag to each documents, which the topic distribution of a document is controlled by the observed tags and one latent tag. In LDA, the topic distribution of a document is drawn from a hyperparameter, without considering the given tags, and while in TWTM, the topic distribution is controlled by a list of given tags with corresponding weight values. The main difference among the models which handle the unstructured text (e.g., LDA and CTM \cite{DBLP:conf/nips/BleiL05}) or the semi-structured documents (e.g., ATM \cite{DBLP:conf/uai/Rosen-ZviGSS04}, Label-LDA \cite{DBLP:conf/emnlp/RamageHNM09}, DMR \cite{DBLP:conf/uai/MimnoM08} and PLDA \cite{Ramage:2011:PLT:2020408.2020481}) is the function that how to generate the topic distribution of a document, or, in other words, the assumption that what distribution the topic of a document follows.

In TWDA, the topic proportions $\vartheta^{d}$ for a document $d$ is obtained by the following function:
{\normalsize
\begin{equation*}
\begin{aligned}
\vartheta^{d} = (\varepsilon^{d})^\mathrm{T} \times T^{d} \times (\frac{ \theta}{\lambda}) \label{eq:twda_vartheta}
\end{aligned}
\end{equation*}
}

When we ignore the tags in a document, the $T^{d}$ in Eq.~(\ref{eq:twda_vartheta}) becomes a binary row vector and the last dimension equals to 1 and the others are 0. In this case, $(\frac{ \theta}{\lambda})$ is simplified to $\lambda$:
{\normalsize
\begin{equation*}
\begin{aligned}
\vartheta^{d} &= (\varepsilon^{d})^\mathrm{T} \times T^{d} \times (\frac{ \theta}{\lambda}) \\
&=\lambda.
\end{aligned}
\end{equation*}
}
The topic distribution of $d$ is simplified to $\lambda$, and as we shown above, $\lambda$ is draw by a Dirichlet prior $\mu$. It means that the topic proportions for the document $d$ as a draw from a Dirichlet distribution which is the basic assumption of LDA \cite{DBLP:journals/jmlr/BleiNJ03}. In others words, when handling the unstructured documents, TWDA degenerates into LDA.

In other words, the topic distribution of a document in TWTM is the weighted average of the topic distributions of the given tags, and to some extent, it is a linear relation between the topic distribution of a document and the tags. While, in TWDA, with the addition of the Dirichlet prior $\mu$, which is equal to generate a latent tag for each document with a special topic distribution, it is a non-linear topic generation procedure in each document.

\section{Large Scale Solutions}\label{sec:LargescaleSolutions}
Currently, many web applications appear with large scale tagged documents, and highlight the issues of large scale semi-structured documents in many areas. In this paper, we propose and compare three different distributed methods based on the framework of TWTM, which focus on the challenge of working at a large scale, in the MapReduce programming framework.
\subsection*{Solution \Rmnum{1}}
The first solution is a tailored parallel algorithm for TWTM. The learning and inference of the proposed model are based on variational method with an EM algorithm. Thus, we design a parallel algorithm for TWTM using MapReduce programming framework.

As shown above, we need to update the global parameters $\pi$, $\theta$, and $\psi$ for a corpus. Every document has associated with the corresponding variational parameters $\xi$ and $\gamma$. The mapper computes these variational parameters for each document and uses them to generate the sufficient statistics to update $\pi$, $\theta$, and$\psi$. And the reducer updates the global parameters $\pi$, $\theta$, and $\psi$.
\begin{enumerate}
	\item \textit{Mapper}: For each document $d$, we compute $\gamma^d$ using the update equation Eq.~(\ref{eq:twtm_gamma}) and obtain $\xi^d$ by Eq.~(\ref{eq:twtm_xi}). The sufficient statistics are kept for each document.
	\item \textit{Reducer}: The Reduce function adds the value to the global parameters $\theta$ and $\psi$ using the sufficient statistics as in Eqs.~(\ref{eq:twtm_theta}), and (\ref{eq:twtm_psi}).
	\item \textit{Driver}: The driver program marshals the entire inference process. At the beginning, the driver initializes all the model parameters $K$, $L$, $\theta$, $\psi$, and $\pi$. The topic number K is user specified; the number of tags $L$ is determined by the data; the initial value of $\pi$ is given by the user, $\theta$ and $\psi$ is randomly initialized. After each MapReduce iteration, the driver normalizes the global $\theta$ and $\psi$.
\end{enumerate}
Note that, because $\pi$ is a global parameter over the corpus, we have to update $\pi$ at the end of each iteration in driver. However, this will lead to a large scale data migration to compute the $\pi$ by Eq.~(\ref{eq:twtm_pi}), since $\pi$ is associated with each document and different documents have different tags which affect the different dimensions in $\pi$. The whole corpus data would migrate to the single driver node. This could generate a bottleneck in the driver.
\subsection*{Solution \Rmnum{2}}
On account of the bottleneck in Solution \Rmnum{1}, we optimize the calculation of $\pi$ through an approximate method as the Solution \Rmnum{2}. The MapReduce procedure of Solution \Rmnum{2} is as follows.
\begin{enumerate}
	\item \textit{Mapper}: For each document $d$, we compute $\gamma^d$ and $\xi^d$ by Eqs.~(\ref{eq:twtm_xi}) and (\ref{eq:twtm_gamma}) and the sufficient statistics for updating $\theta$ and $\psi$. Different with Solution \Rmnum{1}, we obtain a $\pi^s$ for each map data split $s$ by Eq.~(\ref{eq:twtm_pi}).
	\item \textit{Reducer}: The Reduce function adds the value to the global parameters $\theta$ and $\psi$ using the sufficient statistics as in Eqs.~(\ref{eq:twtm_theta}), and (\ref{eq:twtm_psi}).
	\item \textit{Driver}: In the driver function, we only need to compute an average of $\pi^s, s \in (1, \cdots, S)$ where $S$ is the total number of mapper in the cluster.
The driver also normalizes the global $\theta$ and $\psi$ for next iteration.
\end{enumerate}
Solution \Rmnum{2} is an approximate solution of TWTM, which computes the $\pi_s$ for each mapper and takes their average as the solution of $\pi$ to avoid the large scale data migration. 
\subsection*{Solution \Rmnum{3}}
As shown in Eq.~(\ref{eq:twtm_pi}), $\pi_l, l \in (1, \cdots, L)$ is only associated with the documents who contain the $l^{th}$ tag. Thus, before running TWTM, we can cluster the documents into several clusters with the condition that the documents which contain one or a plurality of the same tag should be in the same cluster. It means that the documents are divided into the mutually independent space by the tags. We show the detailed process of the clustering in Appendix C. The MapReduce procedure of Solution \Rmnum{3} is the following procedure.
\begin{enumerate}
	\item \textit{Mapper}: The input of mapper is clusters. For each cluster, we obtain a $\pi^c$ for the cluster $c$, $c \in (1, \cdots, C)$, where $C$ is the number of document clusters, which is the sufficient statistics for updating $\theta$ and $\psi$. 
	\item \textit{Reducer}: The Reduce function adds the value to the global parameters $\theta$ and $\psi$ using the sufficient statistics as in Eqs.~(\ref{eq:twtm_theta}), and (\ref{eq:twtm_psi}).
	\item \textit{Driver}: In the driver, we update $\theta$ and $\psi$. Note that there is no need to recompute $\pi$, and we combine all the $\pi^c$ to obtain the final $\pi$ for current iteration.
\end{enumerate}
Solution \Rmnum{3} is an exact solution for TWTM, and it is equivalent to Solution \Rmnum{1} when the documents are all belong to one cluster. However, Solution \Rmnum{3} provides a more efficient method than Solution \Rmnum{1}, and this depends on the result of document clustering, which would be anther bottleneck in some real applications. Although Solution \Rmnum{2} is an approximate method for modeling the semi-structured documents, it effectively avoids the bottleneck brought by Solution \Rmnum{1} and Solution \Rmnum{3}. The experiment results in Section~\ref{sec:ExperimentalAnalysis} show that Solution \Rmnum{2} works better than Solution \Rmnum{1} and Solution \Rmnum{3}.

It is worth note that all the solutions need to iterate the MapReduce procedure in driver function until convergence or maximum number of iterations is reached. In Section~\ref{sec:ExperimentalAnalysis}, we show the experimental results about the comparisons of the three solutions on document modeling and efficiency.

\section{Experimental Analysis}\label{sec:ExperimentalAnalysis}
\begin{figure*}[t]
	\centering
	\subfigure[TWTM, TWDA, LDA and CTM]{
		\includegraphics[width=0.24\textwidth]{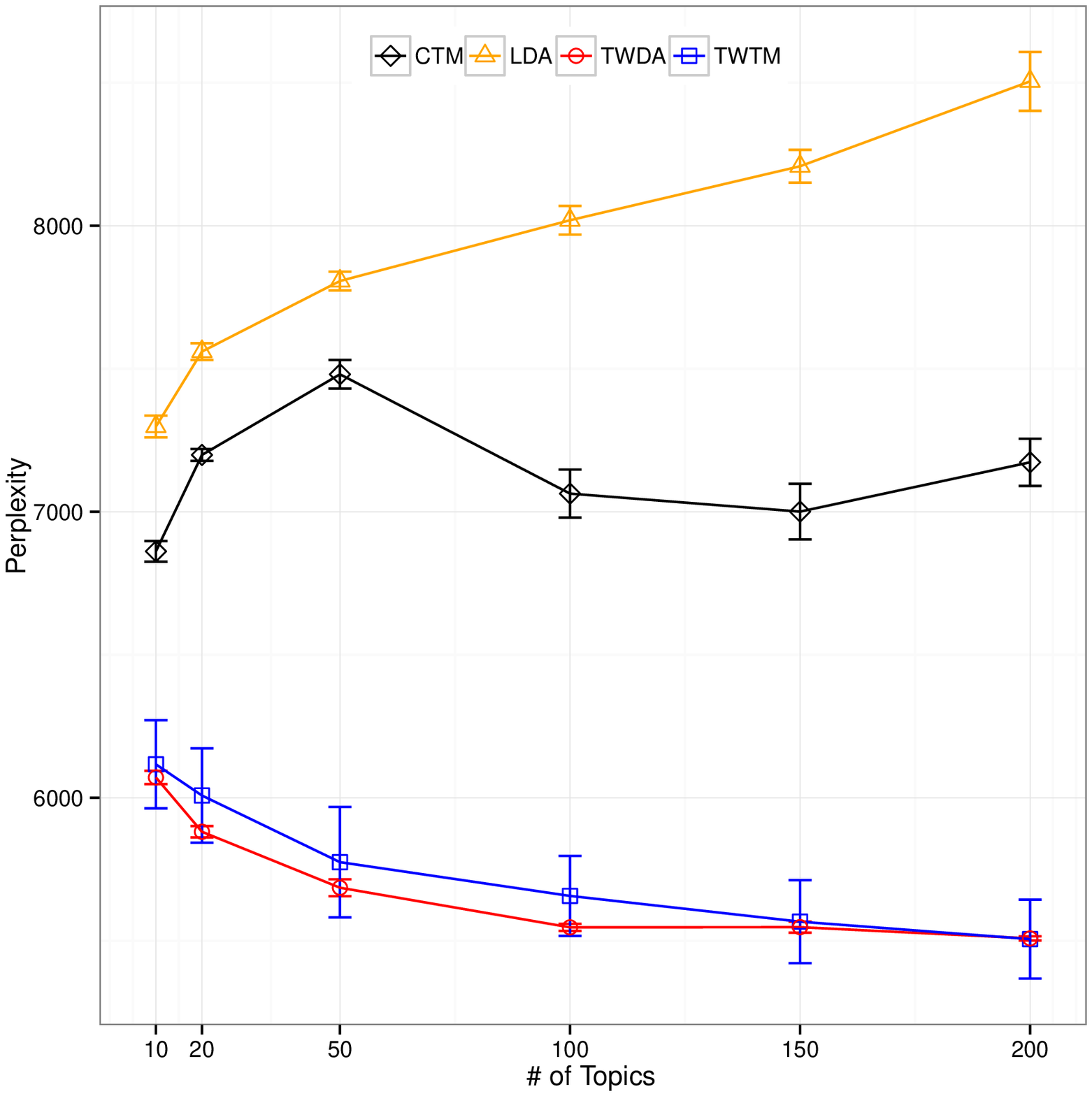}
		\label{fig:per-lda-ctm-twtm-twda-IMDB}}
	\subfigure[TWTM, TWDA, DMR, ATM, CorrLDA, LDA and CTM]{
		\includegraphics[width=0.24\textwidth]{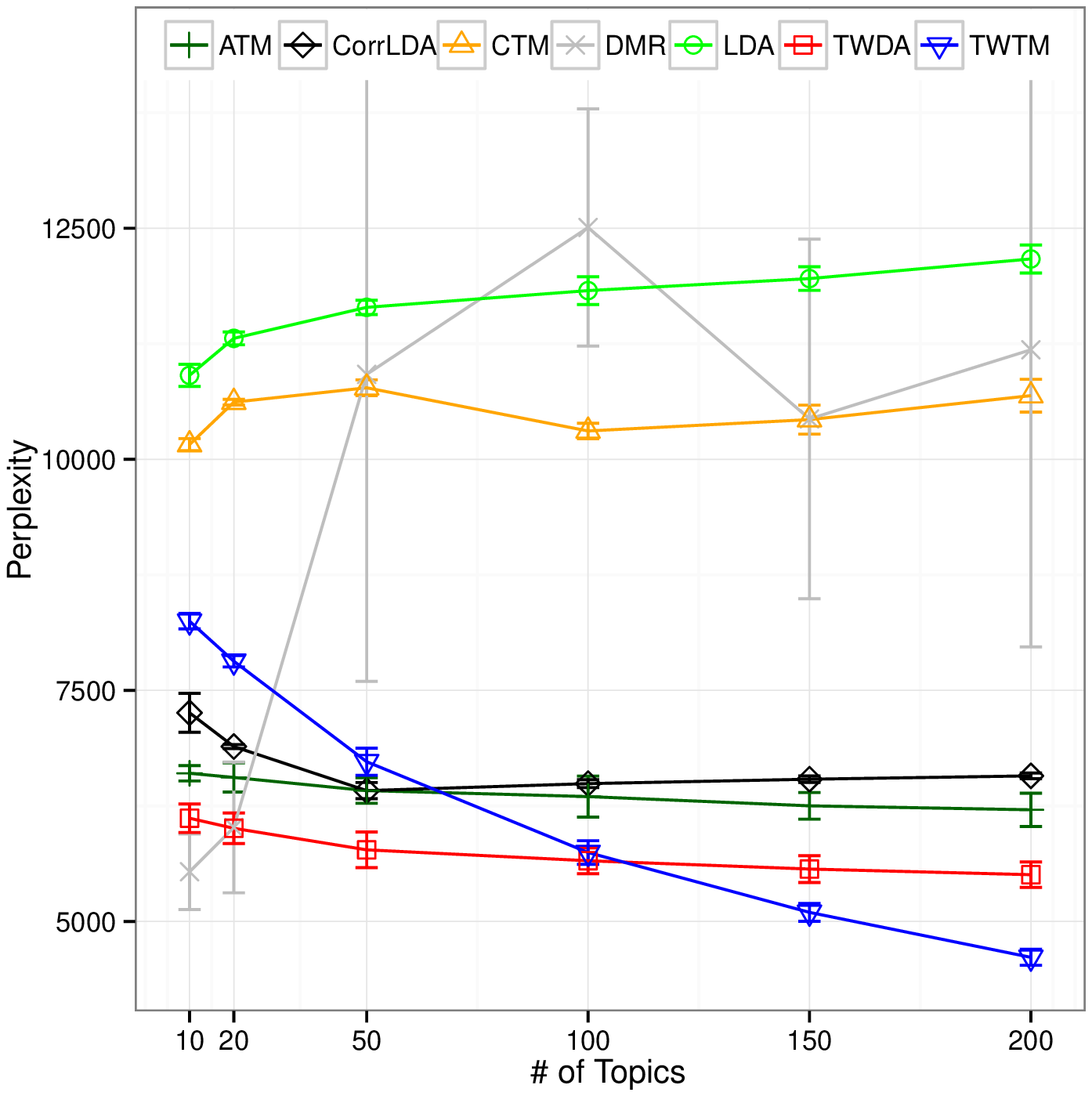}
		\label{fig:per-corrlda-atm-dmr-lda-ctm-twtm-twda-IMDB}}
	\subfigure[TWTM, TWDA and PLDA]{
		\includegraphics[width=0.24\textwidth]{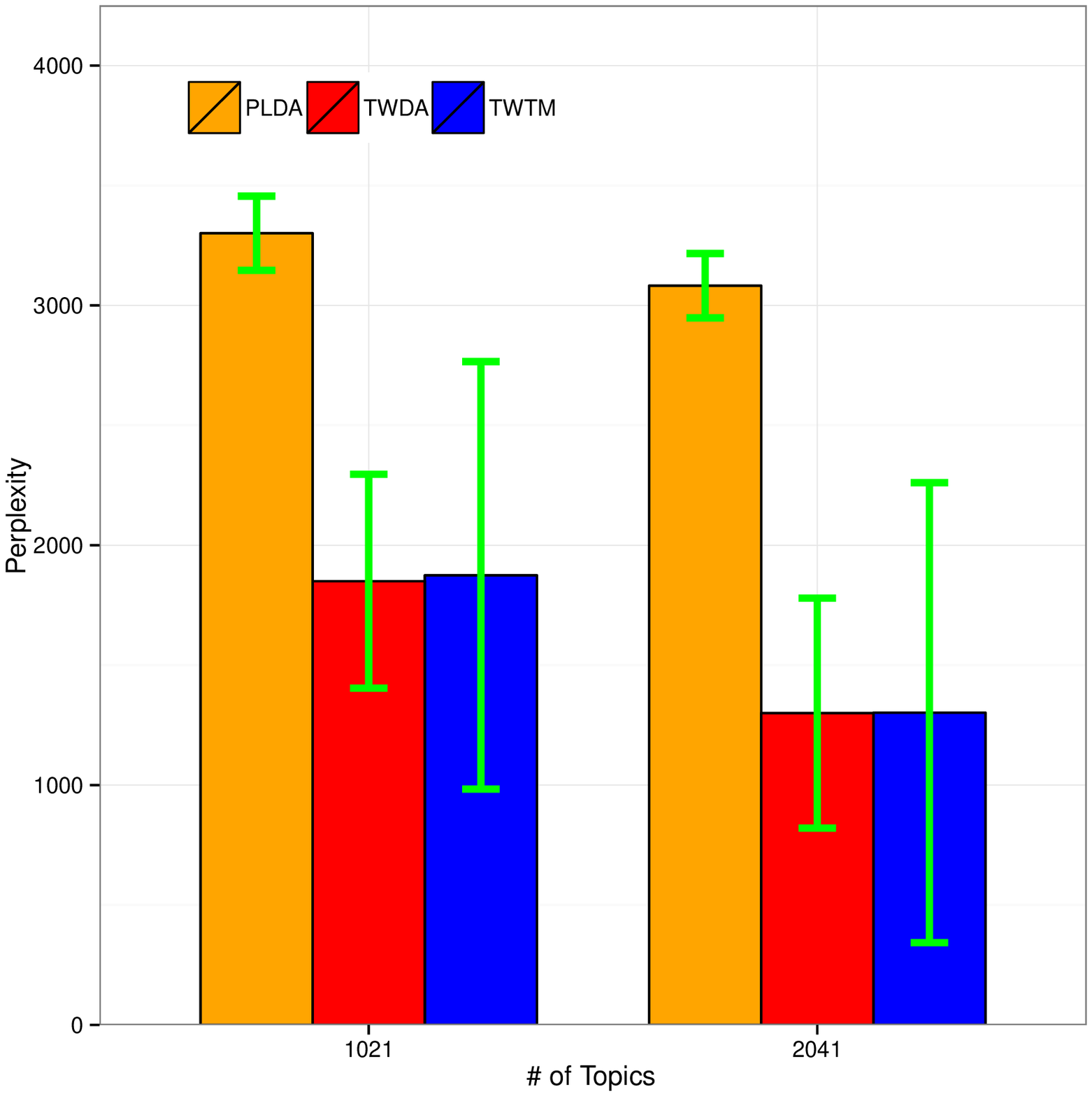}
		\label{fig:per-plda-twtm-twda-IMDB}}
	\caption{Perplexity results of different models on IMDB corpora. LDA and CTM only use the words when training in (a), and add the tags as the word features during the training process in (b).}
	\label{fig:perplexity}
\end{figure*}
\subsection{Experiment Settings}\label{sec:ExperimentalSettings}
In the experiments of this work, we used three semi-structured corpora. The first document collection is the data from Internet Movie Database (IMDB). The data set includes 12,091 movie storylines, 52,274 words after removing stop words, and 3,654 tags. These movies belong to 29 genres. And the tags we used contain directors, stars, time, and movie keywords. The second one consists of technical papers of the Digital Bibliography and Library Project (DBLP) data set\footnote{http://www.informatik.uni-trier.de/$\scriptstyle\mathtt{\sim}$ley/db/}, which is a collection of bibliographic information on major computer science journals and proceedings. In this paper, we use a subset of DBLP that contains abstracts of $D$=27,435 papers, with $W$=70,062 words in the vocabulary and $L$=6,256 unique tags. The tags we used in DBLP include authors and keywords. The last corpus we used contains about 967,012 Wordpress blog posts\footnote{http://wordpress.com} from Kaggle\footnote{http://www.kaggle.com/c/predict-wordpress-likes/data}. In the corpus, there are 163,504 tags and 2,592,562 words. We used this corpus to test the effectiveness and performance of TWTM over a large scale dataset. We implemented the three distributed methods of TWTM using Hadoop 1.1.1 and ran all experiments on a cluster containing 7 physical nodes; each node has 4 cores and 8 threads, and could be configured to run a maximum of 7 mappers and 7 reducers of tasks. With the configuration, we build different scales distributed environments by setting the maximum of mappers used in each node.

We have released the codes on GitHub\footnote{https://github.com/Shuangyinli} including TWTM, TWDA and the three distributed solutions using the Hadoop platform.
\subsection{Results on Documents Modeling}
In order to evaluate the generalization capability of the model, we use the perplexity score that described in \cite{DBLP:journals/jmlr/BleiNJ03}. For a test set of D documents, the perplexity is:
\begin{equation*}
\begin{aligned}
perplexity = \exp \left\{ -\frac{\sum_d^D \log p(\mathbf{w}_d)}{\sum_d^D N_d} \right\},
\end{aligned}
\end{equation*}
where a lower perplexity score represents better document modeling performance.  

There are two parts of the experiments. First, We trained four latent variable models including LDA  \cite{DBLP:journals/jmlr/BleiNJ03}, CTM  \cite{DBLP:conf/nips/BleiL05}, TWTM and TWDA, on the corpora of a set of movie documents in IMDB, to compare the generalization performance of the four models. In this part, LDA and CTM trains text data without taking advantage of tag information. We removed the stop words and conducted experiments using 5-fold cross-validation.
Figure~\ref{fig:per-lda-ctm-twtm-twda-IMDB} demonstrates the perplexity results on the IMDB data set. Clearly, TWTM and TWDA excel both CTM and LDA significantly and consistently.

Second, in order to compare the performance of TWTM and TWDA with other topic models which take advantage of the tag information, we trained TWTM, TWDA, DMR\footnote{We used the Mallet code (http://mallet.cs.umass.edu/).}, PLDA\footnote{We used the code in Stanford Topic Modeling Toolbox (http://www-nlp.stanford.edu/software/tmt/tmt-0.4/).}, Author Topic Model (ATM) \cite{DBLP:conf/uai/Rosen-ZviGSS04}, CorrLDA\cite{Blei03modelingannotated}, CTM, and LDA on the set of movie documents in IMDB and computed the perplexity on test data set. Since CTM and LDA could not handle corpus with tags easily, in this experiment, we treated the given tags as word features for them. In CorrLDA, we used the tags in each document to represent the image segments, so that the CorrLDA can handle the SSDs. Figure~\ref{fig:per-corrlda-atm-dmr-lda-ctm-twtm-twda-IMDB} demonstrates the perplexity results of the six models on the IMDB data. The experiment results shows that TWTM and TWDA are better than the other models, and when $T$ increases, CorrLDA, DMR, CTM and LDA are running into over-fitting, while the trend of TWTM and TWDA keeps going down and the perplexity is significantly lower than those of the baselines.

As PLDA \cite{Ramage:2011:PLT:2020408.2020481} assumes that one of tags may optionally denote as a tag ``latent'' present on every document $d$, thus, we trained PLDA, TWTM and TWDA over 1021 and 2041 topics on IMDB data set with 1020 tags, since in PLDA, each latent topic takes part in exactly one tag in a collection. As shown in \cite{Ramage:2011:PLT:2020408.2020481}, PLDA builds on Labeled LDA  \cite{DBLP:conf/emnlp/RamageHNM09}, and when it set one latent topic and one topic for each tag, it is approximately equivalent to Labeled LDA. For this case, we trained PLDA over 1021 topics.  Figure~\ref{fig:per-plda-twtm-twda-IMDB} shows the perplexity results of TWTM, TWDA and PLDA. Note that TWDA has less mean squared error (MSE) than TWTM.
As the results of Figure~\ref{fig:perplexity} shown, TWTM and TWDA both work well compared with the other topic models which make use of tag information.
\subsection{Results on Tags prediction}
In this section we use TWDA to demonstrated the performance of our works on the tags prediction by process the paper collection in DBLP. In addition to predicting the tags given a document, we evaluate the ability of the proposed model, compared with ATM, DMR and CorrLDA, to predict the tags of the document conditioned on words in the document. In this part, we treat the authors of each paper as the tags, and the abstract as the word features, and we predict the authors of one paper by modeling the paper abstract document data using ATM, DMR, CorrLDA, and TWDA.
For each model, we evaluate the likelihood of the authors given the word features in a document, and rank each possible author by the likelihood function of the author. First, for each model, we can get the topic distribution over a test document $d_{test}$ given one author $a$. Then, we evaluate the $p(d_{test} | a)$ for $d_{test}$ over each author $a$ in the tags(authors) set by
\begin{eqnarray*}
p(d_{test} | a) = \prod_i^N (\sum_z p(z|a) p(w_i  | z)).
\end{eqnarray*}
\begin{figure}[t]
	\centering
	\subfigure[K=100]{
		\includegraphics[width=0.22\textwidth]{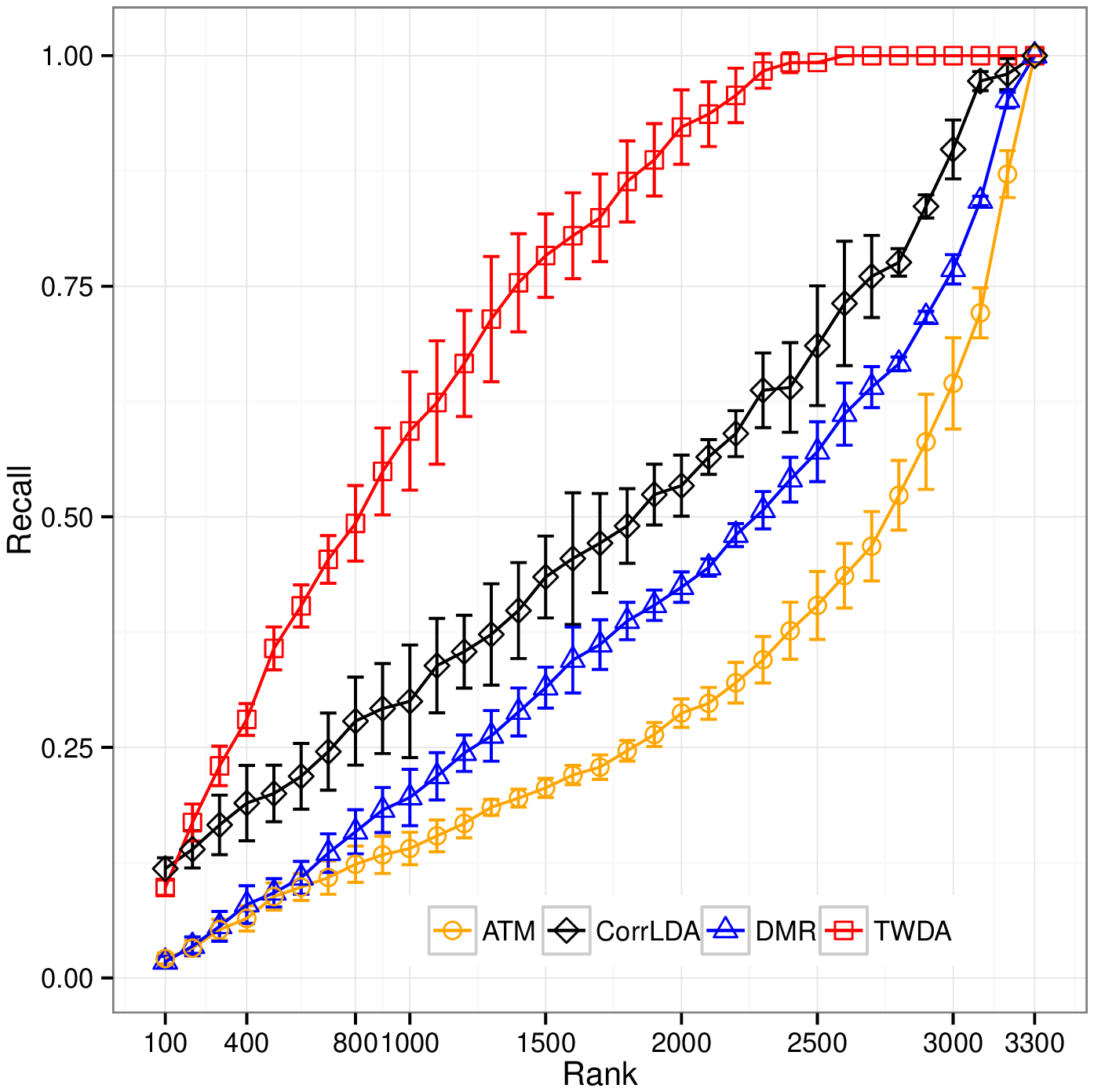}
		\label{fig:predict-atm-dmr-twda-DBLP-100}}
	\subfigure[K=200]{
		\includegraphics[width=0.22\textwidth]{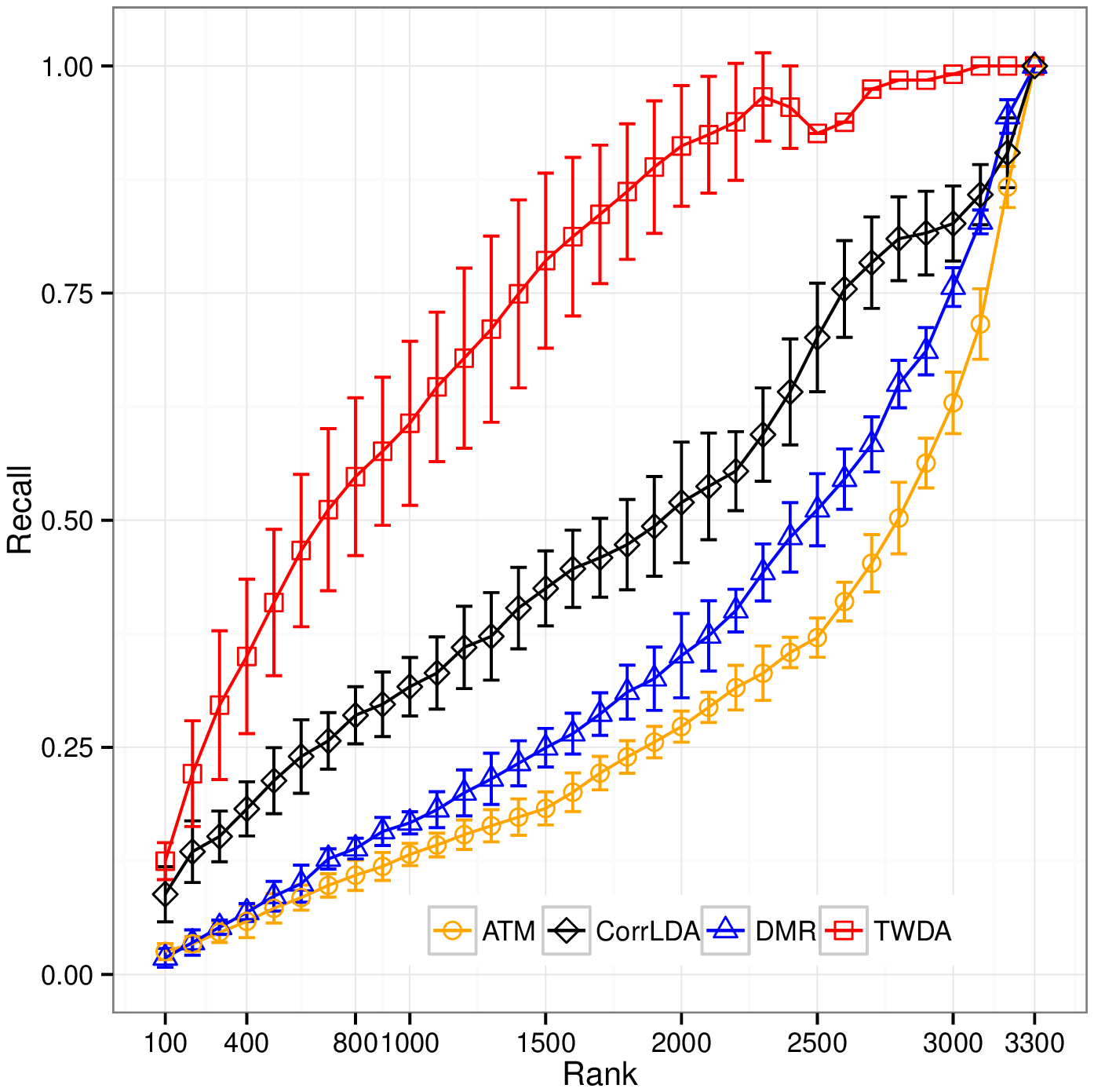}
		\label{fig:predict-atm-dmr-twda-DBLP-200}}
	\caption{Prediction results of TWDA, DMR and ATM for authors on DBLP corpora. We set the number of topic in the corpora to be 100 in (a) and 200 in (b).}
	\label{fig:predict-atm-dmr-twda-DBLP}
\end{figure}
For CorrLDA, we let authors represent image regions, and used $p(d_{test} | region)$ shown in \cite{Blei03modelingannotated} to evaluate the likelihood of a author given a document. For DMR and ATM, the method which define $p(d_{test} | a)$ is shown as \cite{DBLP:conf/uai/MimnoM08}. Note that the likelihoods for a given author over a document are not necessarily comparable among the topic models, however, what we are interested in is the ranking as same as \cite{DBLP:conf/uai/MimnoM08}. 

We trained the three models on DBLP data set using 5-fold cross-validation and shows the recall when the topic in the corpora is set to be 100 and 200. Results are shown in Figure~\ref{fig:predict-atm-dmr-twda-DBLP-100} and Figure~\ref{fig:predict-atm-dmr-twda-DBLP-200}. TWDA ranks authors consistently higher than the other models.
\subsection{Results on Feature Construction for Classification}
\begin{figure}[b]
	\centering
		\includegraphics[width=0.27\textwidth]{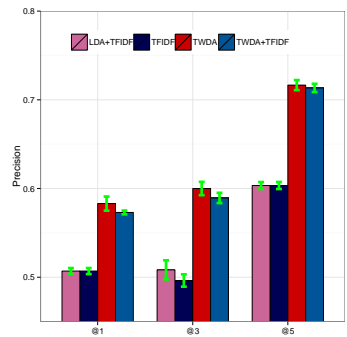}
	\caption{Classification results of different features on @1, @3 and @5 with 5-fold cross-validation.}
	\label{fig:classification-tfidf-lda-twda-IMDB}
\end{figure}
The next experiment is to test the classification performance utilizing feature sets generated by TWDA and other baselines. For the base classifier, we use LIBSVM \cite{DBLP:journals/tist/ChangL11} with Gaussian kernel and the default parameters. For the purpose of comparison, we trained four SVMs on tf-idf word features, features induced by a 30-topic LDA model and tf-idf word features, features generated by a TWDA model with the same number of topics, and features induced by a 30-topic TWDA model and tf-idf word features respectively.

In these experiments, we conducted multi-class classification experiments using the IMDB data set, which contains $29$ genres. We calculated the evaluation metrics @1, @3 and @5 with the provided class tags of movies' genres, using 5-fold cross-validation. We report the movie classification performance of the different methods in Figure~\ref{fig:classification-tfidf-lda-twda-IMDB}, where we see that there is significant improvement in classification performance when using LDA and TWDA comparing with only using tf-idf features, and TWDA outperforms both LDA and tf-idf in terms of @1, @3 and @5.

In order to show the classification performance better, we also calculated the evaluation metrics F-Measure (F1-score). The results of F-Measure is reported in Table~\ref{table:classification-F-Measure}. TWDA provides substantially better performance on F-Measure.
\begin{table}[t]
\renewcommand{\arraystretch}{1.2}
\caption{Classification results of different features on F1-score}
\label{table:classification-F-Measure}
\centering
\begin{tabular}{|c|c|c|c|}
\hline
 F1-score & @1 & @3 & @5 \\
\hline
TFIDF & 0.5 & 0.41 & 0.39 \\
\hline
LDA+TFIDF & 0.5 & 0.42 & 0.39 \\
\hline
TWDA & 0.57 & 0.5 & 0.47 \\
\hline
TWDA+TFIDF & 0.58 & 0.5 & 0.47 \\
\hline
\end{tabular}
\end{table}
\subsection{Results on Model Robustness}
\begin{figure}[b]
	\centering
	\subfigure[K=100]{
		\includegraphics[width=0.22\textwidth]{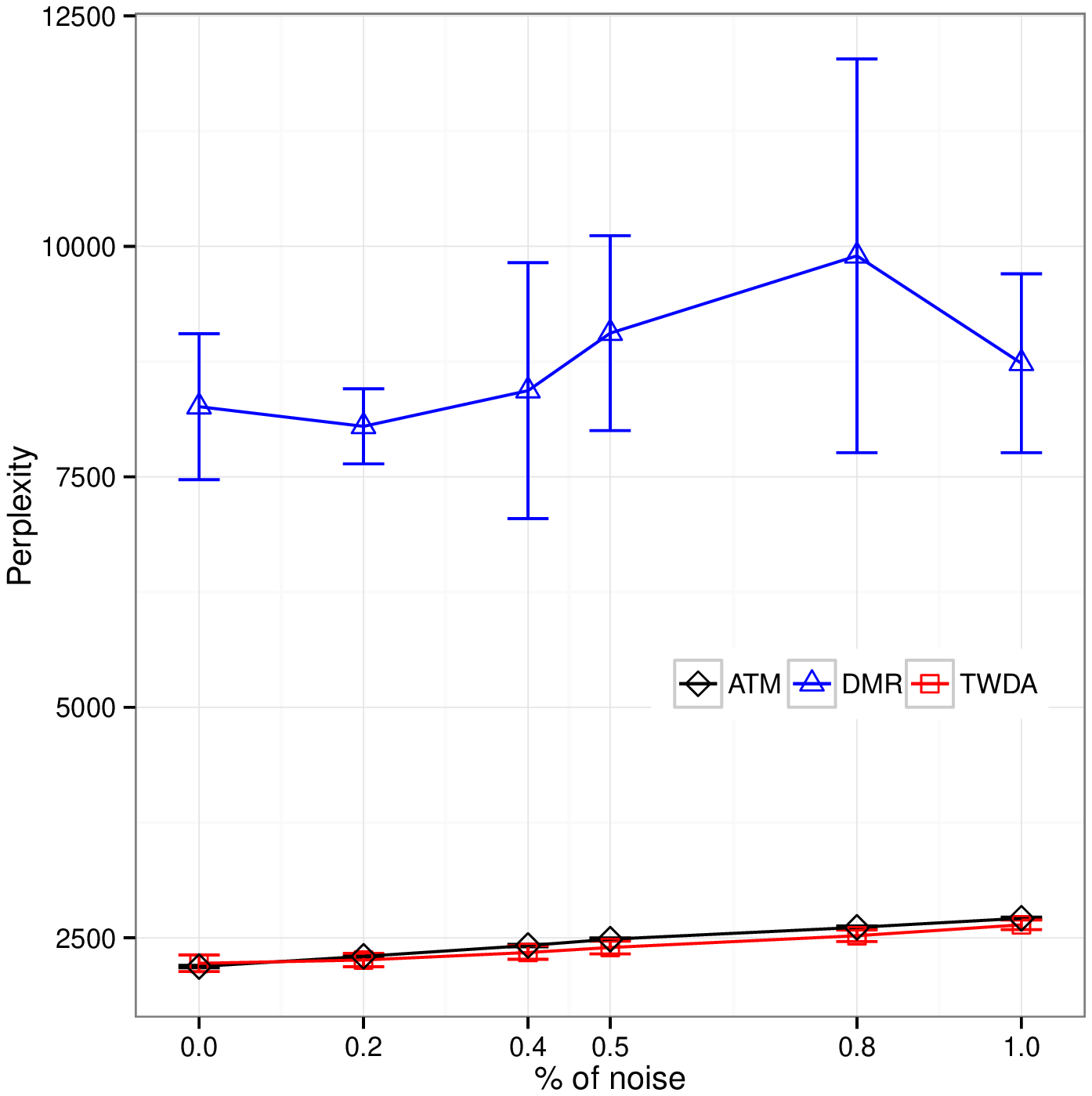}
		\label{fig:robustness-twda-atm-dmr-DBLP-100}}
	\subfigure[K=200]{
		\includegraphics[width=0.22\textwidth]{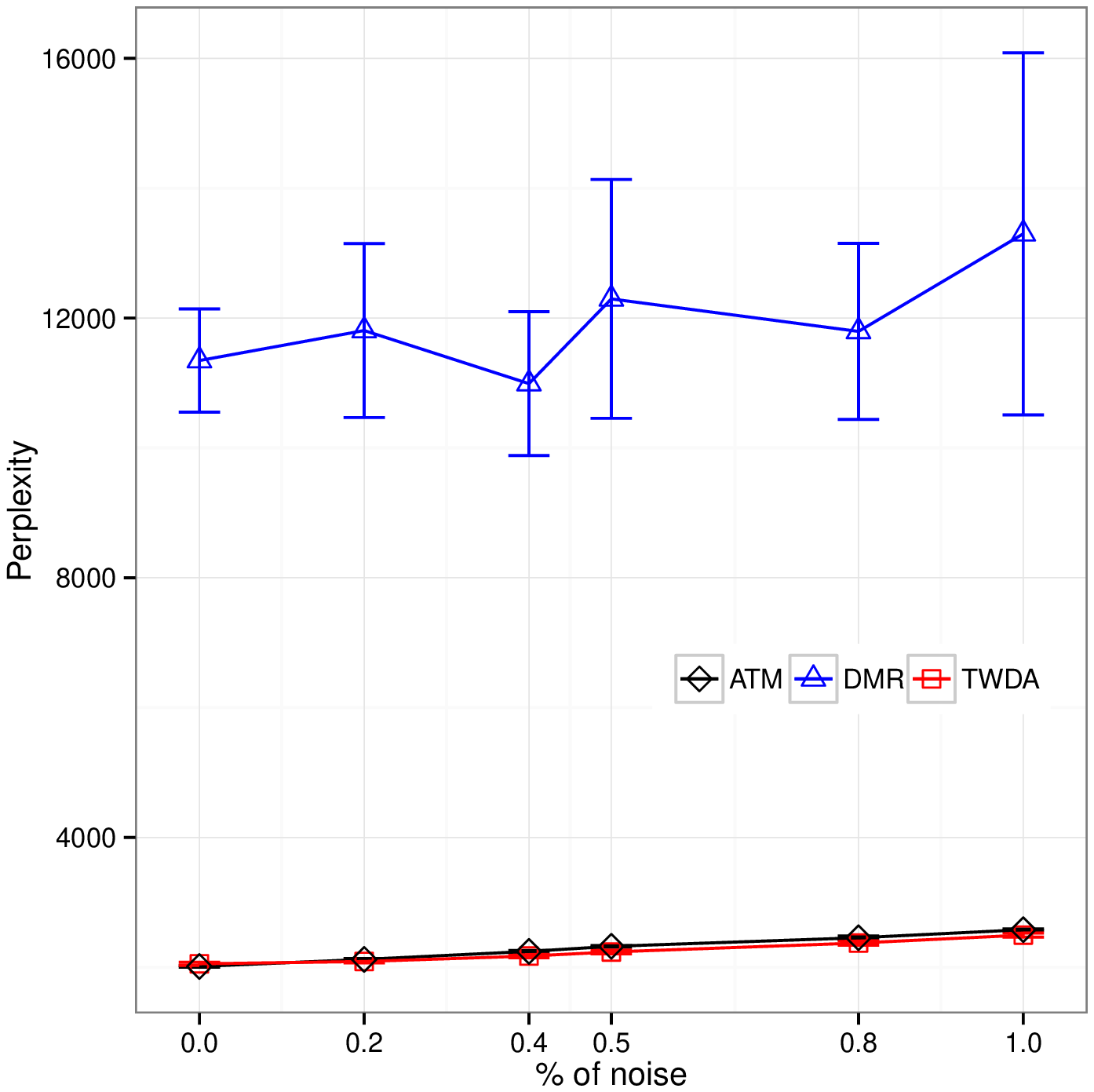}
		\label{fig:robustness-twda-atm-dmr-DBLP-200}}
	\caption{The Results of adding noise to different models(ATM, DMR and TWDA). (a) set K=100, and (b) set K=200.  Steady trending means a good performance on model robustness.}
	\label{fig:robustness-twda-atm-dmr-DBLP}
\end{figure}
We demonstrated the performance of our work on model robustness in this part of experimental analysis. In this part, we measured and compared the perplexity when we added noise tags information to the test documents using DBLP data set. Respectively, we randomly added $20\%$, $40\%$, $50\%$, $80\%$ and $100\%$ noise tags into a test document and then calculated the perplexity. For example, if a paper document in DBLP has five authors, adding $20\%$ noise is that we randomly selected one author from the author set of the DBLP corpora and added into the paper as a noise author. 

In some real-world applications, the noise tags appeared in a document may have some relevance to the real tags. So in this experiment, we selected the noise tags from the author-tag set to meet the real applications to some extent. In this experiment, the DBLP corpora contains more than 6,000 tags, the noise tags we added into a test document would be very sparse for the whole tag set in the corpora. So, we added the different percentages noise tags into the test document to show the trend of perplexity as the noise content increases.
Figure~\ref{fig:robustness-twda-atm-dmr-DBLP} shows that both TWDA and ATM have a more steady trend as the noise level increases, compared with DMR.
\begin{table}[t]
\renewcommand{\arraystretch}{1.3}
\caption{Some examples of the normalized weights among the original tags and noise tags. The noise tags are in red, and the numbers are the corresponding weight values.}
\label{table:robustness-twda-atm-dmr-DBLP}
\begin{tabular}{|l|}
\hline
 ``Bug isolation via remote program sampling \cite{Liblit:2003:BIV:780822.781148}"\\

 Ben Liblit: 0.185  \quad \quad \quad \quad  Alex Aiken: 0.2257\\
aAlice X. Zheng: 0.228  \quad \quad \quad \quad   Michael I. Jordan: 0.349\\
\textsl{\color{red} K. G. Shin: 0.01}\\

\hline
 ``Web question answering: is more always better? \cite{Dumais:2002:WQA:564376.564428}"\\

Susan Dumais: 0.986  \quad \quad \quad \quad  Michele Banko: 0.0032\\
Eric Brill: 0.0038  \quad \quad \quad \quad  Jimmy Lin: 0.0038\\
Andrew Ng: 0.0024\\
\textsl{\color{red} R. Katz: 0.00018}\\

\hline
 ``Contextual search and name disambiguation in email \\ using graphs \cite{Minkov:2006:CSN:1148170.1148179}"\\

 Einat Minkov: 0.425  \quad \quad \quad \quad  William W. Cohen: 0.342\\
Andrew Y. Ng: 0.128   \\
\textsl{\color{red} J. Ma: 0.033 \quad \quad \quad \quad   D. Ferguson: 0.07}\\

\hline
 ``A Sparse Sampling Algorithm for Near-Optimal Planning \\
in Large Markov Decision Processes \cite{Kearns:2002:SSA:599616.599698}"\\

Michael Kearns: 0.296  \quad \quad \quad \quad   Yishay Mansour:0.166\\
Andrew Y. Ng: 0.31 \\
\textsl{\color{red} J. Blythe: 0.089 \quad \quad \quad \quad  B. Adida: 0.027}\\
\textsl{\color{red} P. J. Modi: 0.1}\\

\hline
 ``The nested Chinese restaurant process and bayesian \\ nonparametric inference of topic \cite{Blei:2010:NCR:1667053.1667056}"\\

 David M. Blei:0.46 \quad \quad \quad \quad  Thomas L. Griffiths:0.186\\
Michael I. Jordan:0.225 \\
\textsl{\color{red}B. Clifford:0.031 \quad \quad \quad \quad R. Szeliski:0.048}\\
\textsl{\color{red} X. Wang:0.05}\\
\hline
\end{tabular}
\end{table}
Table~\ref{table:robustness-twda-atm-dmr-DBLP} shows some examples about the weights between the original tags and noise tags. 
The red tags are the noise added into the test data, and the values behind are the weights among the tags we inference from the TWDA model. Note that, we showed the weight values after normalized.
As the results shown, TWDA has a good performance on model robustness, for the weight values of the noise tags are much smaller than the other original tags. In some applications, we can use the proposed model to rank the tags given in a document, which would be a good approach to tag recommendation and annotation.
\subsection{Results on Large-scale Datasets}
We demonstrated the performance of the three proposed parallelized solutions of TWTM for a large-scale dataset from training time and accuracy on document modeling, which are suitable for TWDA as well.

\begin{figure}[b]
	\centering
	  \subfigure[]{     
    \includegraphics[width=0.22\textwidth]{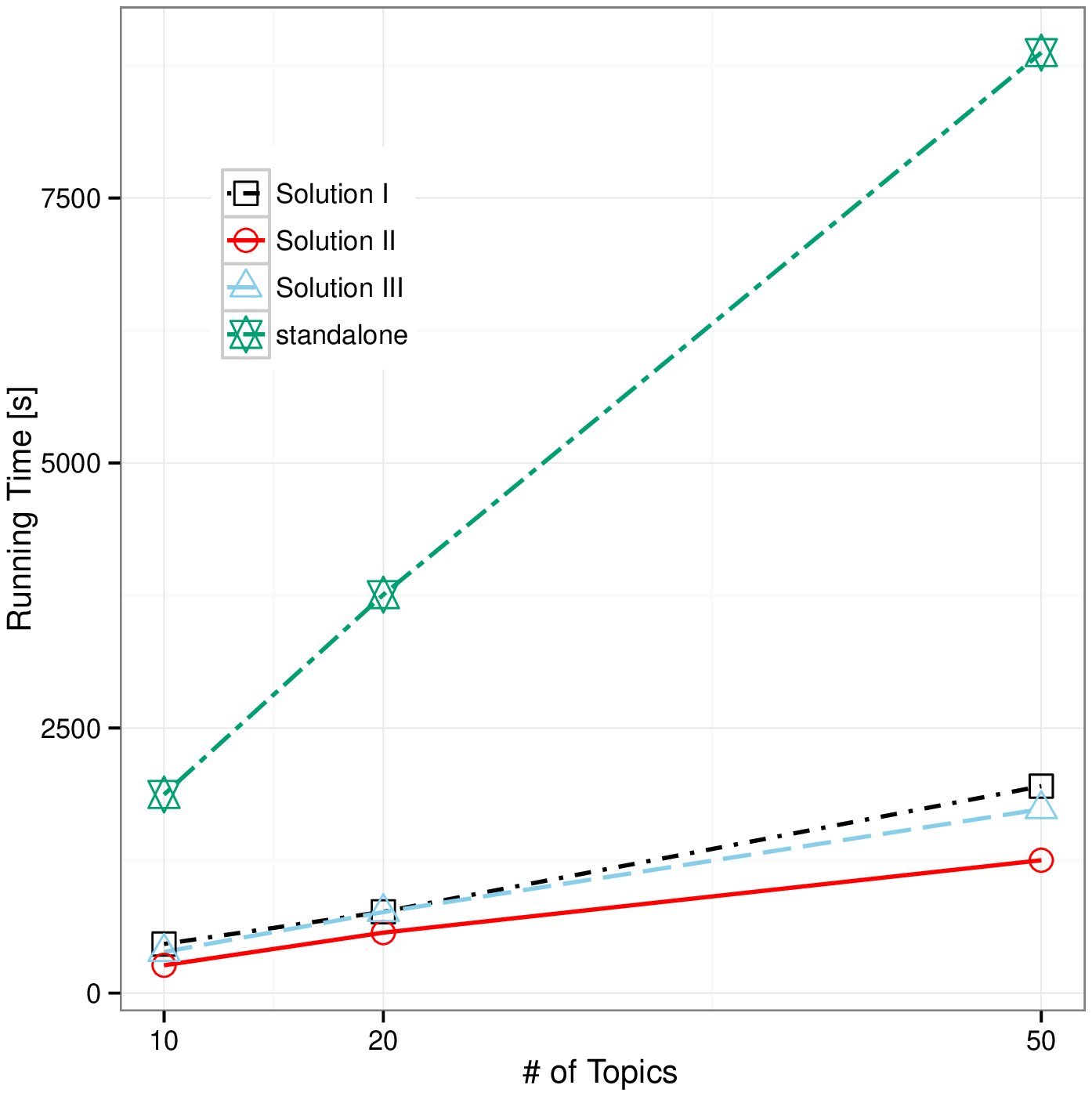}
		\label{fig:largescale-Runtime-Numtopics}
		}
		\subfigure[]{     
    \includegraphics[width=0.22\textwidth]{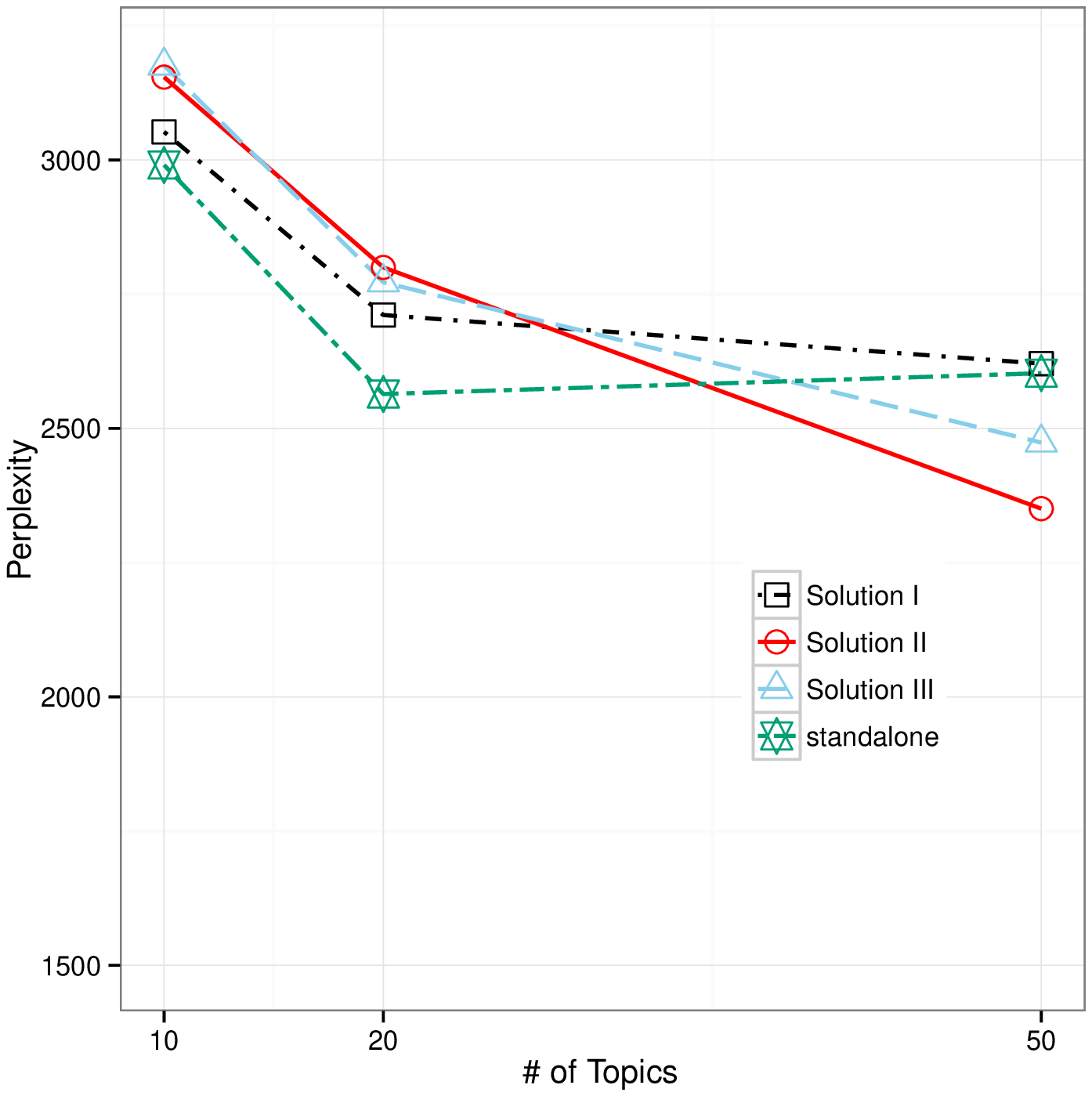}
		\label{fig:largescale-perplexity-threeSolutions}
		}
	\caption{(a) The average training time per iteration for Solution \Rmnum{1}, \Rmnum{2}, \Rmnum{3} with different number of topics compared with the standard TWTM. (b) The perplexity results for Solution \Rmnum{1}, \Rmnum{2}, \Rmnum{3}, and the standard TWTM.}
	\label{fig:largescale-Runtime-Perplexity}
\end{figure}

Firstly, we measured and compared the training time of Solutions \Rmnum{1}, \Rmnum{2} and \Rmnum{3} using the Wordpress blog data set with the same system setting and model parameters. We used a doc-indexed sparse storage mode for the matrix of $\xi$-document, for the matrix would be very huge over a large scale data set. Figure~\ref{fig:largescale-Runtime-Numtopics} shows the performance on the average training time per iteration of the three solutions compared with the standard TWTM as the baseline, when we set the number of topic $K$ = 10, 20 and 50 respectively.

\begin{figure}[t]
	\centering
		\subfigure[K = 20]{     
    \includegraphics[width=0.22\textwidth]{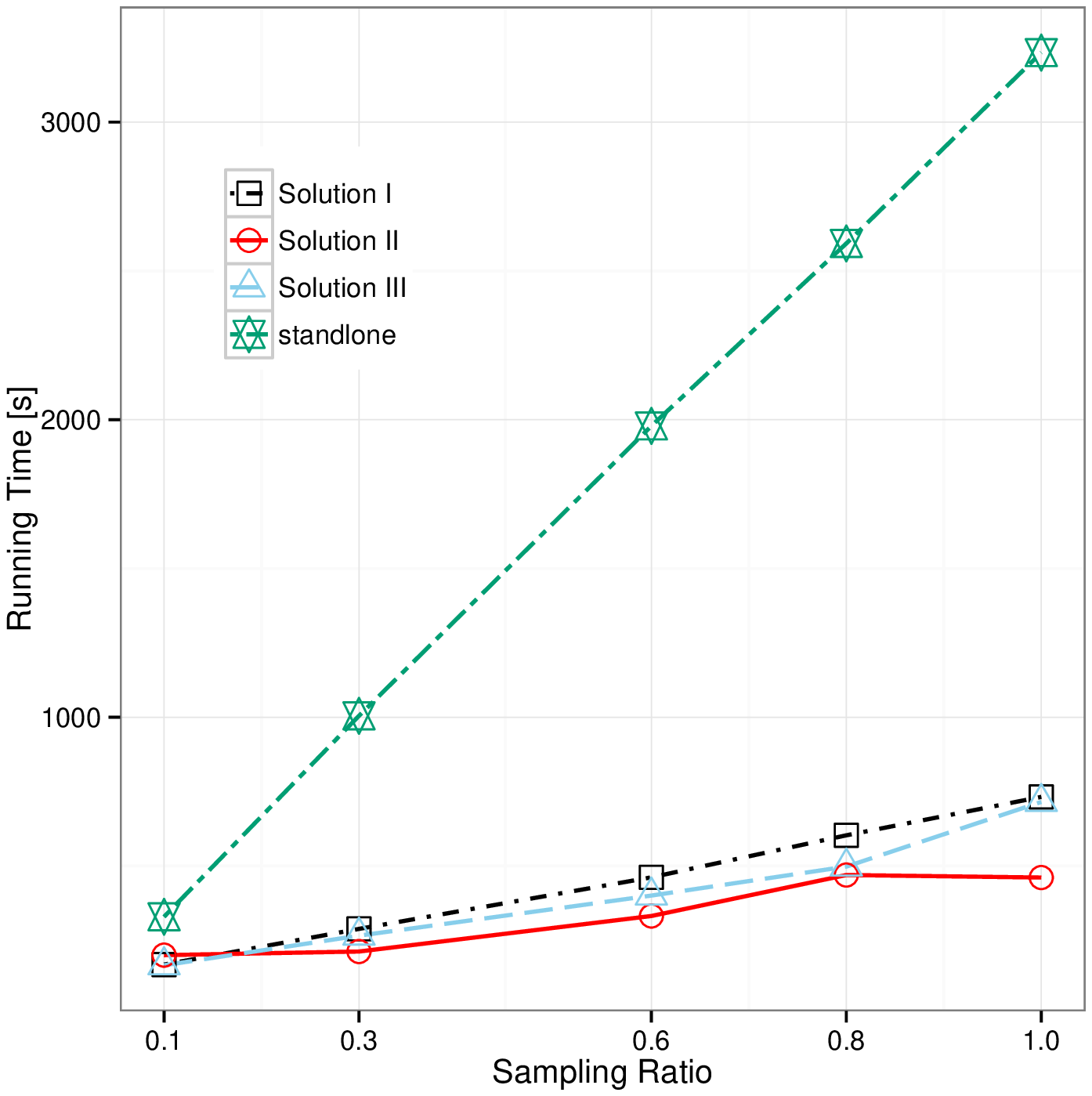}
		\label{fig:largescale-Runtime-sampling-T20}
		}
		\subfigure[K = 50]{     
    \includegraphics[width=0.22\textwidth]{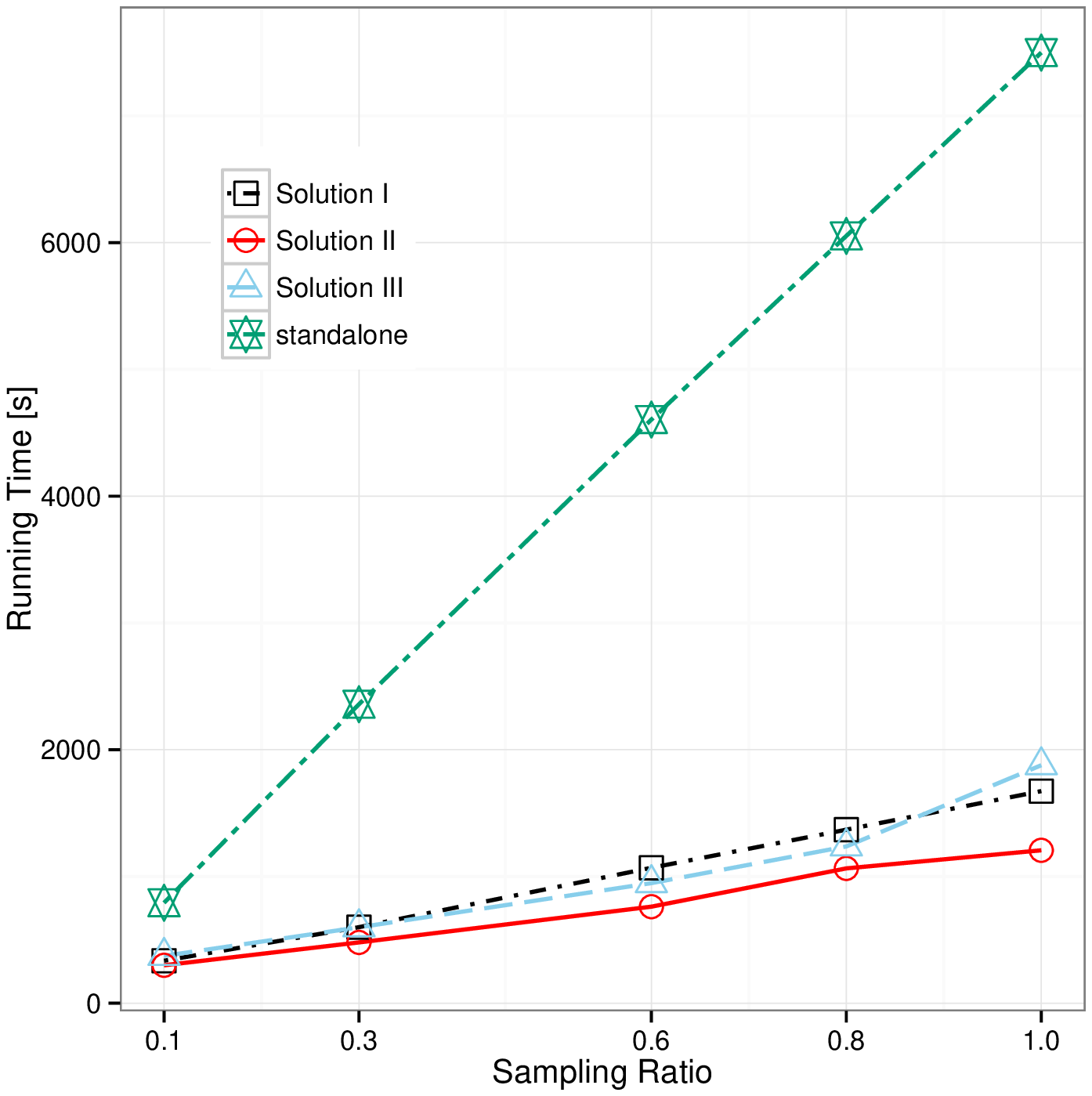}
		\label{fig:largescale-Runtime-sampling-T50}
		}
	\caption{The average training time per iteration on the Wordpress corpus with different number of sampling radios for Solution \Rmnum{1}, \Rmnum{2}, \Rmnum{3}.}
	\label{fig:largescale-Runtime-sampling}
\end{figure}

\begin{figure}[t]
	\centering
		\subfigure[K = 20]{     
    \includegraphics[width=0.22\textwidth]{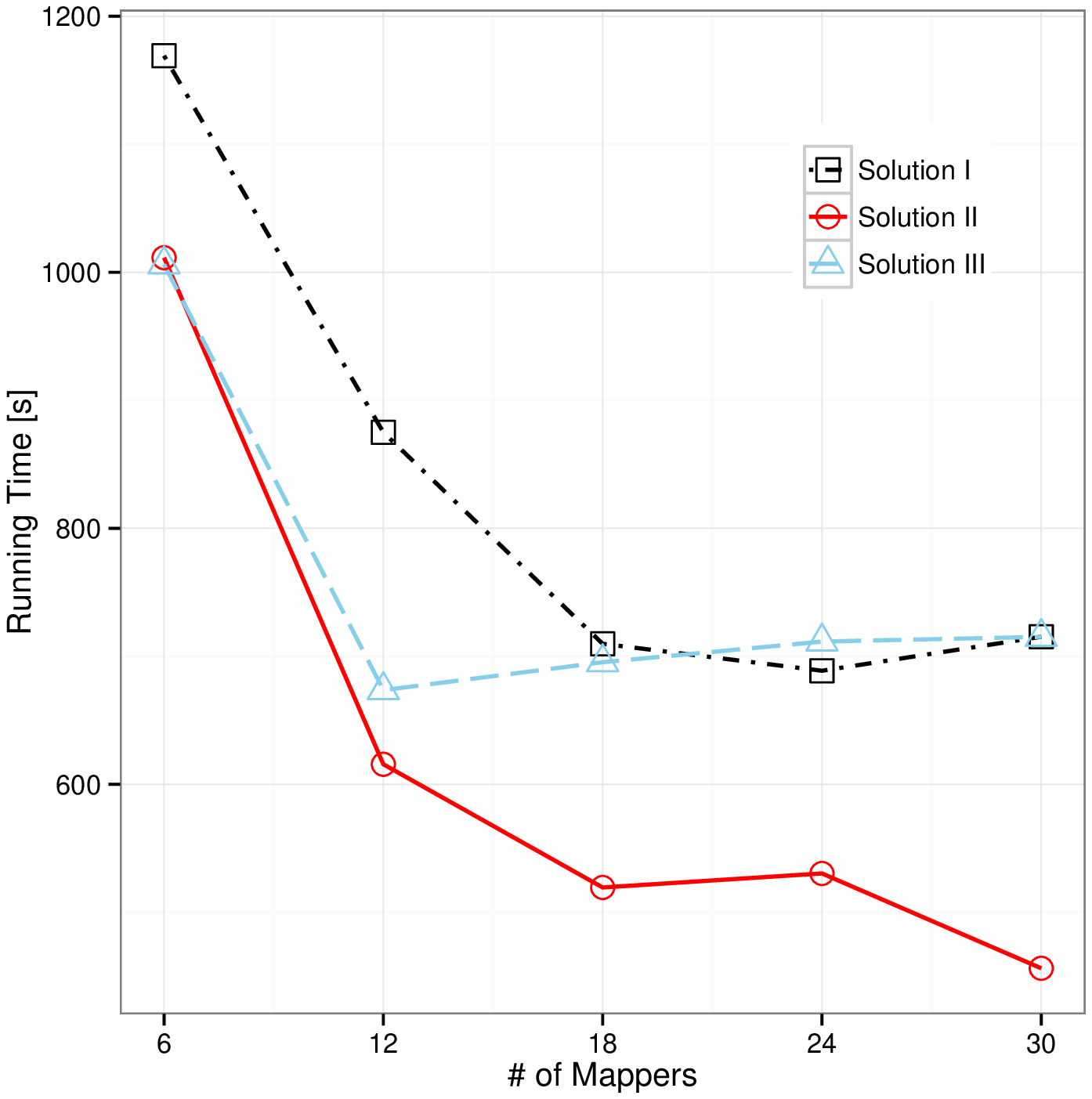}
		\label{fig:largescale-Runtime-NumMapper-T20}
		}
		\subfigure[K = 50]{     
    \includegraphics[width=0.22\textwidth]{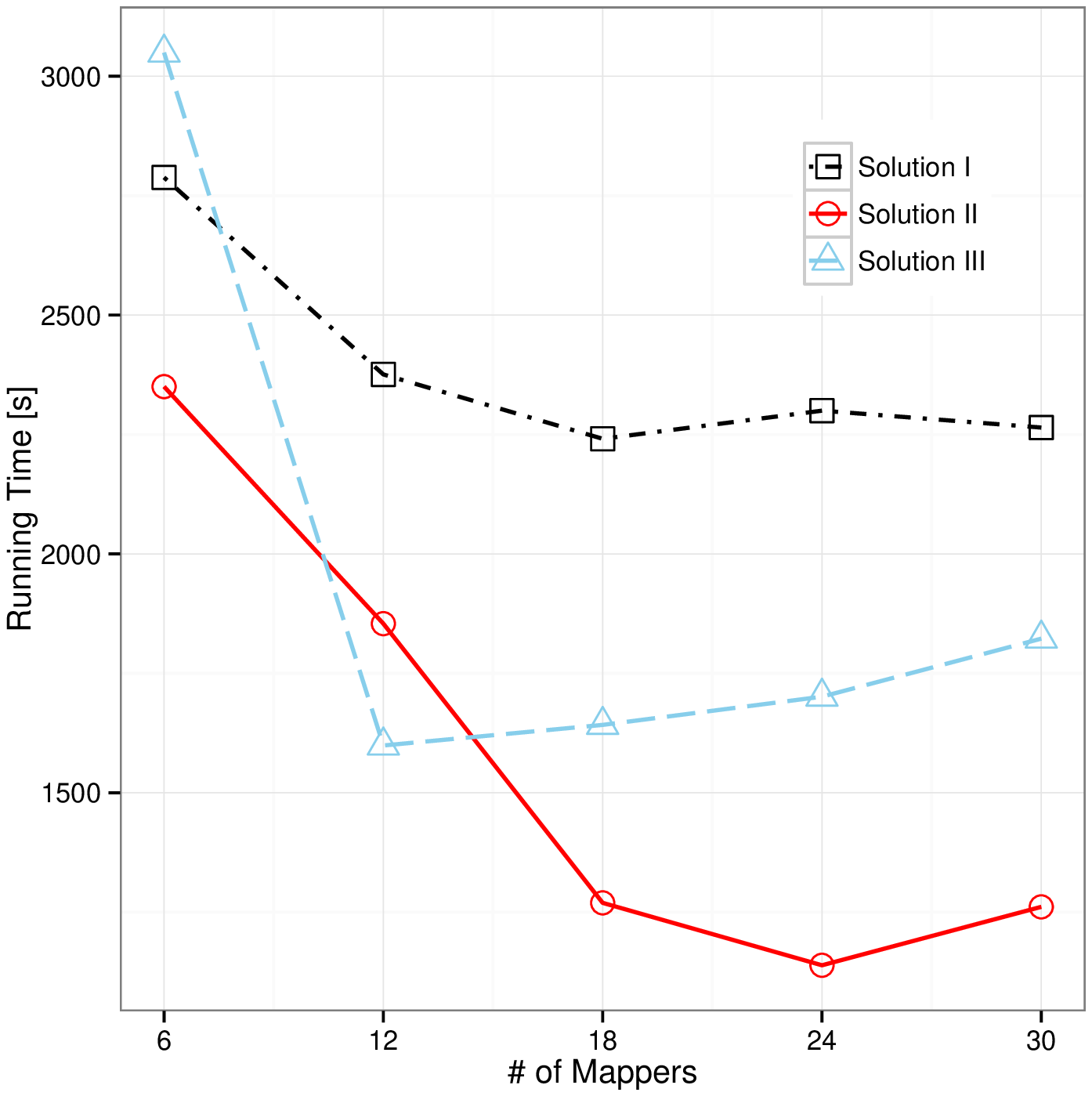}
		\label{fig:largescale-Runtime-NumMapper-T50}
		}
	\caption{The average training time per iteration on the Wordpress corpus with different number of Mappers for Solution \Rmnum{1}, \Rmnum{2}, \Rmnum{3}. Note that the horizontal axis repesents the maximum number of Mappers used in a training task.}
	\label{fig:largescale-Runtime-NumMapper}
\end{figure}

Secondly, We sampled the training dataset from the Wordpress corpus with different sample ratios, 0.1, 0.3, 0.6, 0.8 and 1.0, to show the performance of running time by different size of training dataset. In addition, we limited the maximum number of Mappers in the configuration when we trained the model as described in Section~\ref{sec:ExperimentalSettings}, to demonstrate the comparison performance of the three solutions under the restricted resources. Figure~\ref{fig:largescale-Runtime-sampling} and Figure~\ref{fig:largescale-Runtime-NumMapper} show the results about the average training time per iteration of the three solutions using different sample ratios and Mappers of dataset when training, by setting the number of topic $K$ = 10, 20 and 50 respectively. From this part of experiments, we find that Solution \Rmnum{2} has a better performance of efficiency than Solution \Rmnum{1} and \Rmnum{3}.


\begin{table}[b]
\renewcommand{\arraystretch}{1.2}
\caption{The average training time (second) per iteration for Solution \Rmnum{2} and PLDA }
\label{table:runtime-twtm-plda}
\centering
\begin{tabular}{|c|c|c|c|c|c|}
\hline
Sampling radio   & 0.1 & 0.3 & 0.6 & 0.8 & 1.0\\
\hline
PLDA & 66.6 & 114.8 & 193.4 & 250.6 & 276.4\\
\hline
Solution \Rmnum{2} &  77.6 & 88.6  & 104.8 & 116.2 & 120.8\\
\hline
\end{tabular}
\end{table}
Meanwhile, in order to compare with other model, such as PLDA, we used the Wordpress dataset with 1,000 tags to train a PLDA model with $K_l =1$ (we used the code from Stanford Topic Modeling Toolbox). We trained TWTM by Solution \Rmnum{2} with $K = 5$. Table~\ref{table:runtime-twtm-plda} shows the comparison of PLDA and TWTM by Solution \Rmnum{2}.

As described in Section~\ref{sec:LargescaleSolutions}, in Solution \Rmnum{1}, it would spend a great deal of time on data migration to update $\pi$ in Driver process, and in Solution \Rmnum{3}, a lot of resources are taken on the clustering process in each Mapper, especially when the corpus is non-homogeneous which leads to uneven loading of each Mapper. While, Solution \Rmnum{2} avoids these problems by a approximation method.

Lastly, we measured the generalization capability of the three solutions using the perplexity and conducted experiments. We held out $20\%$ of the data for test and trained the three solutions on the remaining $80\%$. We observe that there is relatively little difference among the solutions compared with the standard TWTM in terms of perplexity as shown in Figure~\ref{fig:largescale-perplexity-threeSolutions} when the number of topic increases. That is, all the three solutions are good approximations in terms of model fitness. It is worthy to note that Solution \Rmnum{2} has almost the same performance as Solution \Rmnum{1} and Solution \Rmnum{3}.

\section{Conclusion}\label{sec:Conclusion}
With the tag-weighted topic model proposed in the paper, we provide and analyze a probabilistic approach for mining semi-structured documents. Meanwhile, three distributed solutions for TWTM are presented to handle the large scale problems. Besides, TWTM is able to obtain the topics distribution of tags in the corpus, which is very useful for text classification, clustering and other data mining applications. At the same time, we propose a novel framework of processing the tagged text with a high extensibility, and uses a novel function of tag-weighted topic assignment of documents. As an extended model, TWDA shows the capability on handling the mixture corpora of semi-structured documents and unstructured documents. The second benefit of the tag-weighted topic model is that it allows one to incorporate different types of tags in modeling documents, and provides a general framework for multi-tag modeling at not only the level of tags but also the level of documents. It helps provide a different approach in classification, clustering, recommendation, and so on. For large scale semi-structured documents, the proposed solutions are shown to be effective and efficient for some complex web applications. In the future, we plan to apply TWTM to different practical areas (e.g., image classification and annotation, video retrieval).

\bibliographystyle{plain}
\bibliography{tkde}

\begin{thebibliography}{10}

\bibitem{DBLP:journals/ml/AndrieuFDJ03}
Christophe Andrieu, Nando de~Freitas, Arnaud Doucet, and Michael~I. Jordan.
\newblock An introduction to mcmc for machine learning.
\newblock {\em Machine Learning}, 50(1-2):5--43, 2003.

\bibitem{DBLP:conf/uai/AsuncionWST09}
Arthur~U. Asuncion, Max Welling, Padhraic Smyth, and Yee~Whye Teh.
\newblock On smoothing and inference for topic models.
\newblock In {\em UAI}, pages 27--34, 2009.

\bibitem{DBLP:conf/nips/Attias99}
Hagai Attias.
\newblock A variational baysian framework for graphical models.
\newblock In {\em NIPS}, pages 209--215, 1999.

\bibitem{DBLP:journals/jei/BishopN07}
Christopher~M. Bishop and Nasser~M. Nasrabadi.
\newblock {\it Pattern Recognition and Machine Learning}.
\newblock {\em J. Electronic Imaging}, 16(4):049901, 2007.

\bibitem{Blei:2010:NCR:1667053.1667056}
David~M. Blei, Thomas~L. Griffiths, and Michael~I. Jordan.
\newblock The nested chinese restaurant process and bayesian nonparametric
  inference of topic hierarchies.
\newblock {\em J. ACM}, 57(2):7:1--7:30, February 2010.

\bibitem{Blei03modelingannotated}
David~M. Blei, Michael I, David~M. Blei, and Michael I.
\newblock Modeling annotated data.
\newblock In {\em In Proc. of the 26th Intl. ACM SIGIR Conference}, 2003.

\bibitem{DBLP:conf/nips/BleiL05}
David~M. Blei and John~D. Lafferty.
\newblock Correlated topic models.
\newblock In {\em NIPS}, 2005.

\bibitem{DBLP:conf/nips/BleiM07}
David~M. Blei and Jon~D. McAuliffe.
\newblock Supervised topic models.
\newblock In {\em NIPS}, 2007.

\bibitem{DBLP:journals/jmlr/BleiNJ03}
David~M. Blei, Andrew~Y. Ng, and Michael~I. Jordan.
\newblock Latent dirichlet allocation.
\newblock {\em Journal of Machine Learning Research}, 3:993--1022, 2003.

\bibitem{DBLP:journals/corr/abs-1002-4665}
Jordan~L. Boyd-Graber and David~M. Blei.
\newblock Syntactic topic models.
\newblock {\em CoRR}, abs/1002.4665, 2010.

\bibitem{Bratko2006679}
Andrej Bratko and Bogdan Filipic.
\newblock Exploiting structural information for semi-structured document
  categorization.
\newblock {\em Information Processing and Management}, 42(3):679 -- 694, 2006.

\bibitem{DBLP:conf/cikm/CaiMHZ08}
Deng Cai, Qiaozhu Mei, Jiawei Han, and Chengxiang Zhai.
\newblock Modeling hidden topics on document manifold.
\newblock In {\em CIKM}, pages 911--920, 2008.

\bibitem{DBLP:journals/tist/ChangL11}
Chih-Chung Chang and Chih-Jen Lin.
\newblock Libsvm: A library for support vector machines.
\newblock {\em ACM TIST}, 2(3):27, 2011.

\bibitem{DBLP:journals/jmlr/ChangB09}
Jonathan Chang and David~M. Blei.
\newblock Relational topic models for document networks.
\newblock {\em Journal of Machine Learning Research - Proceedings Track},
  5:81--88, 2009.

\bibitem{DBLP:conf/kdd/ChenZC12}
Xu~Chen, Mingyuan Zhou, and Lawrence Carin.
\newblock The contextual focused topic model.
\newblock In {\em KDD}, pages 96--104, 2012.

\bibitem{dean2008mapreduce}
Jeffrey Dean and Sanjay Ghemawat.
\newblock Mapreduce: simplified data processing on large clusters.
\newblock {\em Communications of the ACM}, 51(1):107--113, 2008.

\bibitem{DBLP:conf/kdd/DengHZYL11}
Hongbo Deng, Jiawei Han, Bo~Zhao, Yintao Yu, and Cindy~Xide Lin.
\newblock Probabilistic topic models with biased propagation on heterogeneous
  information networks.
\newblock In {\em KDD}, pages 1271--1279, 2011.

\bibitem{Dumais:2002:WQA:564376.564428}
Susan Dumais, Michele Banko, Eric Brill, Jimmy Lin, and Andrew Ng.
\newblock Web question answering: is more always better?
\newblock In {\em Proceedings of the 25th annual international ACM SIGIR
  conference on Research and development in information retrieval}, SIGIR '02,
  pages 291--298, New York, NY, USA, 2002. ACM.

\bibitem{DBLP:finding}
Thomas~L. Griffiths and Mark Steyvers.
\newblock Finding scientific topics.
\newblock In {\em PNAS}, pages 449--455, 2004.

\bibitem{DBLP:conf/sigir/Hofmann99}
Thomas Hofmann.
\newblock Probabilistic latent semantic indexing.
\newblock In {\em SIGIR}, pages 50--57, 1999.

\bibitem{DBLP:conf/nips/IwataYU09}
Tomoharu Iwata, Takeshi Yamada, and Naonori Ueda.
\newblock Modeling social annotation data with content relevance using a topic
  model.
\newblock In {\em NIPS}, pages 835--843, 2009.

\bibitem{Kearns:2002:SSA:599616.599698}
Michael Kearns, Yishay Mansour, and Andrew~Y. Ng.
\newblock A sparse sampling algorithm for near-optimal planning in large markov
  decision processes.
\newblock {\em Mach. Learn.}, 49(2-3):193--208, November 2002.

\bibitem{DBLP:conf/nips/Lacoste-JulienSJ08}
Simon Lacoste-Julien, Fei Sha, and Michael~I. Jordan.
\newblock Disclda: Discriminative learning for dimensionality reduction and
  classification.
\newblock In {\em NIPS}, pages 897--904, 2008.

\bibitem{Liblit:2003:BIV:780822.781148}
Ben Liblit, Alex Aiken, Alice~X. Zheng, and Michael~I. Jordan.
\newblock Bug isolation via remote program sampling.
\newblock {\em SIGPLAN Not.}, 38(5):141--154, May 2003.

\bibitem{DBLP:journals/corr/abs-0901-0358}
Pierre-Francois Marteau, Gildas M{\'e}nier, and Eugen Popovici.
\newblock Weighted naive bayes model for semi-structured document
  categorization.
\newblock {\em CoRR}, abs/0901.0358, 2009.

\bibitem{DBLP:conf/uai/MimnoM08}
David~M. Mimno and Andrew McCallum.
\newblock Topic models conditioned on arbitrary features with
  dirichlet-multinomial regression.
\newblock In {\em UAI}, pages 411--418, 2008.

\bibitem{Minkov:2006:CSN:1148170.1148179}
Einat Minkov, William~W. Cohen, and Andrew~Y. Ng.
\newblock Contextual search and name disambiguation in email using graphs.
\newblock In {\em Proceedings of the 29th annual international ACM SIGIR
  conference on Research and development in information retrieval}, SIGIR '06,
  pages 27--34, New York, NY, USA, 2006. ACM.

\bibitem{DBLP:conf/nips/PettersonSCBN10}
James Petterson, Alexander~J. Smola, Tib{\'e}rio~S. Caetano, Wray~L. Buntine,
  and Shravan Narayanamurthy.
\newblock Word features for latent dirichlet allocation.
\newblock In {\em NIPS}, pages 1921--1929, 2010.

\bibitem{DBLP:conf/emnlp/RamageHNM09}
Daniel Ramage, David Hall, Ramesh Nallapati, and Christopher~D. Manning.
\newblock Labeled lda: A supervised topic model for credit attribution in
  multi-labeled corpora.
\newblock In {\em EMNLP}, pages 248--256, 2009.

\bibitem{Ramage:2011:PLT:2020408.2020481}
Daniel Ramage, Christopher~D. Manning, and Susan Dumais.
\newblock Partially labeled topic models for interpretable text mining.
\newblock In {\em Proceedings of the 17th ACM SIGKDD international conference
  on Knowledge discovery and data mining}, KDD '11, pages 457--465, New York,
  NY, USA, 2011. ACM.

\bibitem{DBLP:conf/uai/Rosen-ZviGSS04}
Michal Rosen-Zvi, Thomas~L. Griffiths, Mark Steyvers, and Padhraic Smyth.
\newblock The author-topic model for authors and documents.
\newblock In {\em UAI}, pages 487--494, 2004.

\bibitem{DBLP:conf/icml/SatoN12}
Issei Sato and Hiroshi Nakagawa.
\newblock Rethinking collapsed variational bayes inference for lda.
\newblock In {\em ICML}, 2012.

\bibitem{DBLP:conf/vldb/TreschPL95}
Markus Tresch, Neal Palmer, and Allen Luniewski.
\newblock Type classification of semi-structured documents.
\newblock In {\em VLDB}, pages 263--274, 1995.

\bibitem{Wei:2006:LDM:1148170.1148204}
Xing Wei and W.~Bruce Croft.
\newblock Lda-based document models for ad-hoc retrieval.
\newblock In {\em Proceedings of the 29th annual international ACM SIGIR
  conference on Research and development in information retrieval}, SIGIR '06,
  pages 178--185, New York, NY, USA, 2006. ACM.

\bibitem{DBLP:conf/kdd/YiS00}
Jeonghee Yi and Neel Sundaresan.
\newblock A classifier for semi-structured documents.
\newblock In {\em KDD}, pages 340--344, 2000.

\bibitem{DBLP:conf/icml/ZhuAX09}
Jun Zhu, Amr Ahmed, and Eric~P. Xing.
\newblock Medlda: maximum margin supervised topic models for regression and
  classification.
\newblock In {\em ICML}, page 158, 2009.

\end{thebibliography}
\vspace{-15 mm}
\vfill
\begin{IEEEbiography}[{\includegraphics[width=1in,height=1.25in,clip,keepaspectratio]{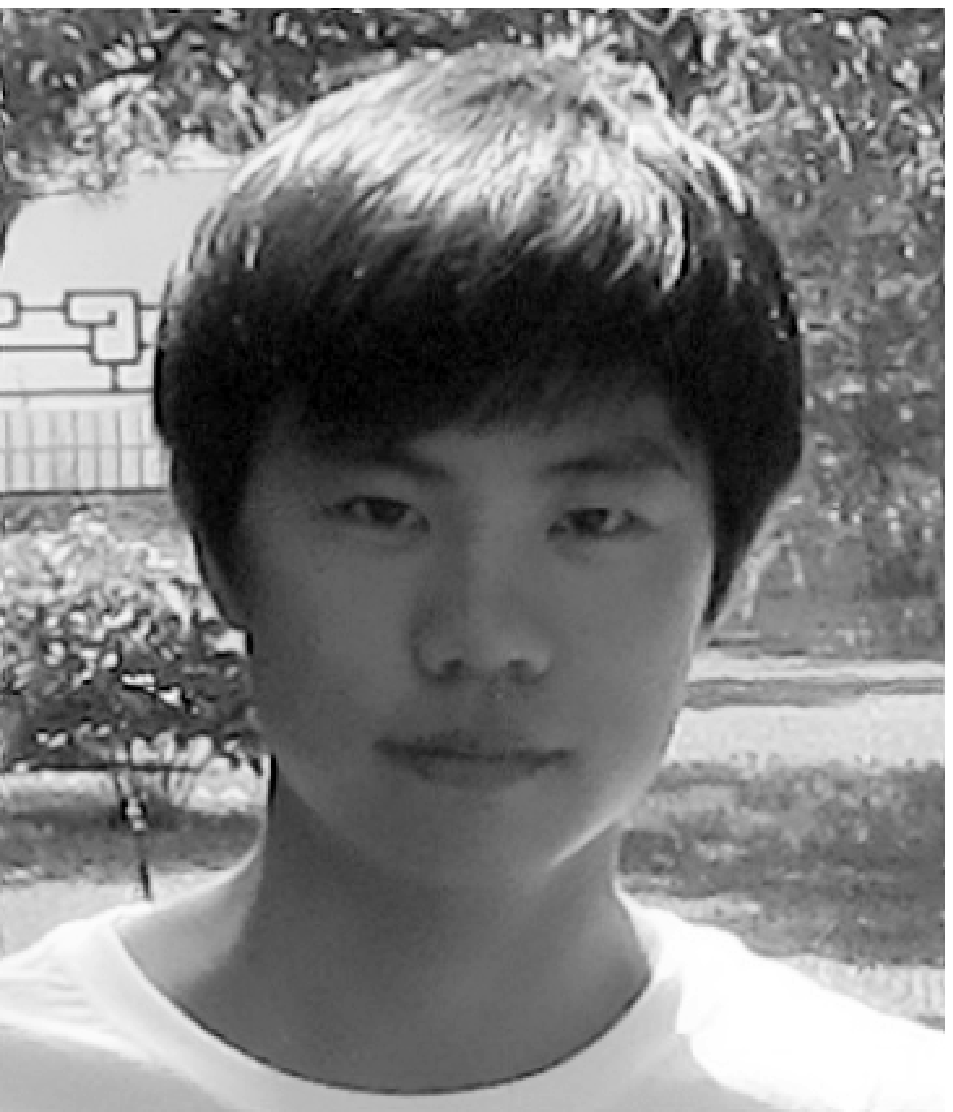}}]{Shuangyin Li} received the Master degree in School of Information Science and Technology, Sun Yat-sen University, China, in 2011. During the Master's program, he focused on the research of large scale image retrieval system on Hadoop platform. Currently, he is active within the field of Text Mining and Artificial Intelligence, and continues his research in a PhD track at Sun Yat-sen University. His PhD research focuses on Topic Model and Deep Neural Networks, and he has published several research mainly focused on the semi-structured documents modeling.
\end{IEEEbiography}
\vspace{-12 mm}
\vfill
\begin{IEEEbiography}[{\includegraphics[width=1in,height=1.25in,clip,keepaspectratio]{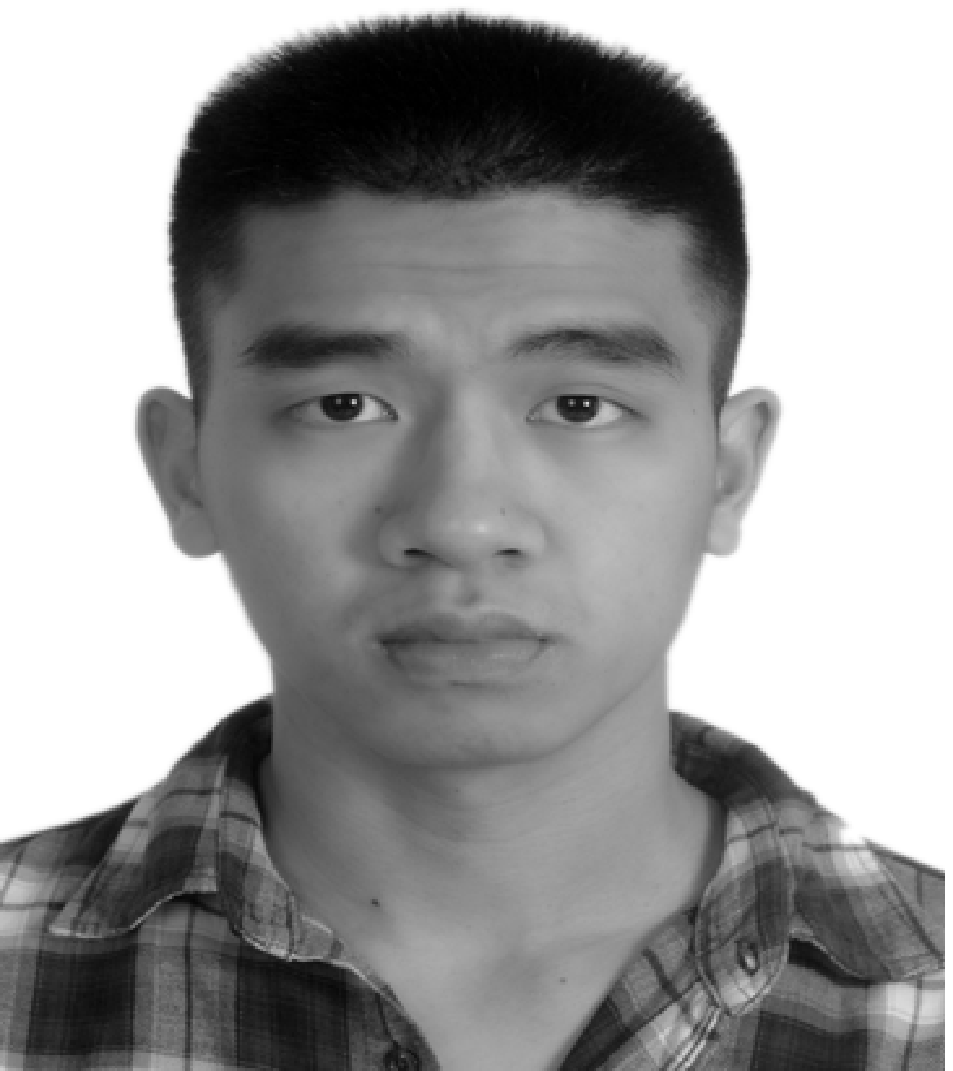}}]{Jiefei Li}  received the Bachelor's degree in Department of Computer Science, Sun Yat-sen University, in 2011. Currently, he is studying for a master's degree in Sun Yat-sen University. His research focuses on Topic Model.
\end{IEEEbiography}
\vspace{-14 mm}
\vfill
\begin{IEEEbiography}[{\includegraphics[width=1in,height=1.25in,clip,keepaspectratio]{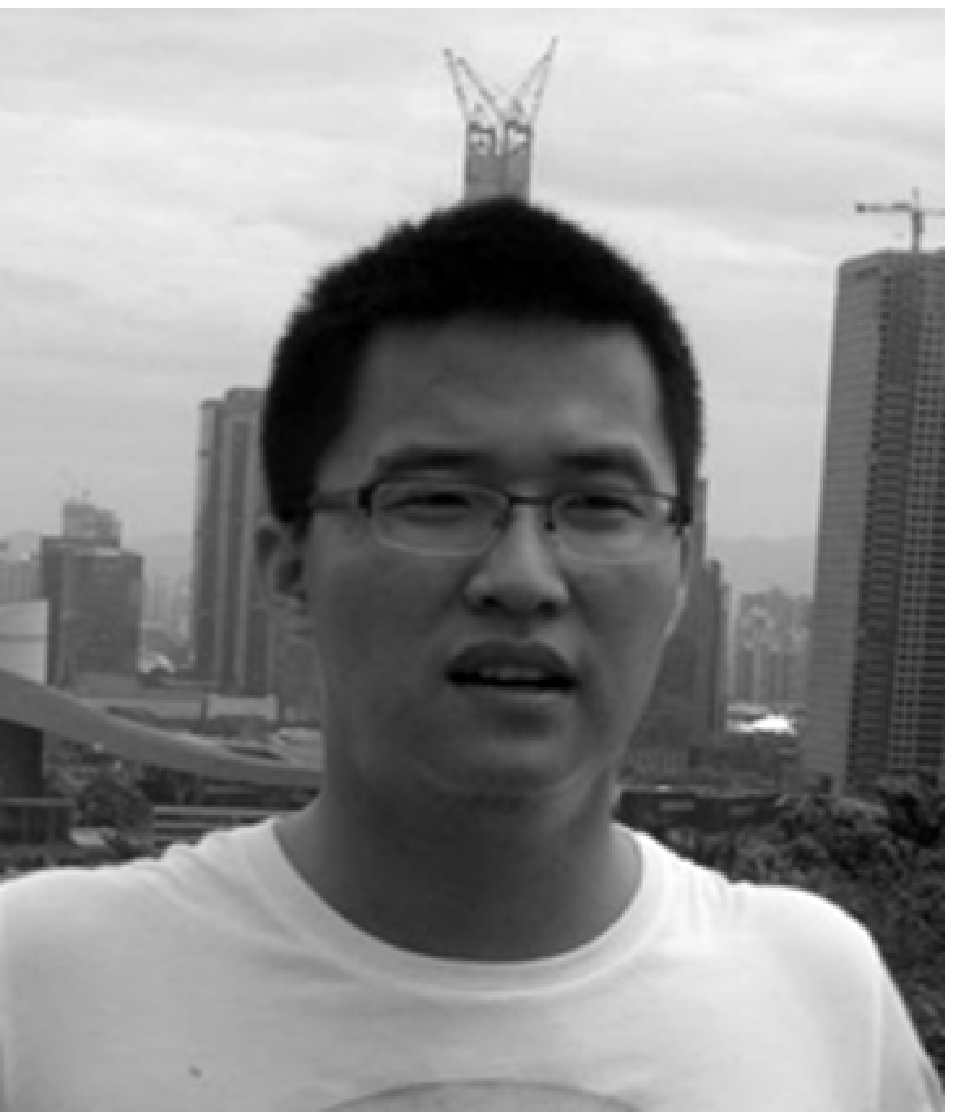}}]{Guan Huang} received the Bachelor's degree in Department of Computer Science, Sun Yat-sen University, in 2009. Currently, he is studying for a master 's degree in Sun Yat-sen University in the filed of word embedding and topic model, learning to rank.
\end{IEEEbiography}
\vspace{-15 mm}
\vfill
\begin{IEEEbiography}[{\includegraphics[width=1in,height=1.25in,clip,keepaspectratio]{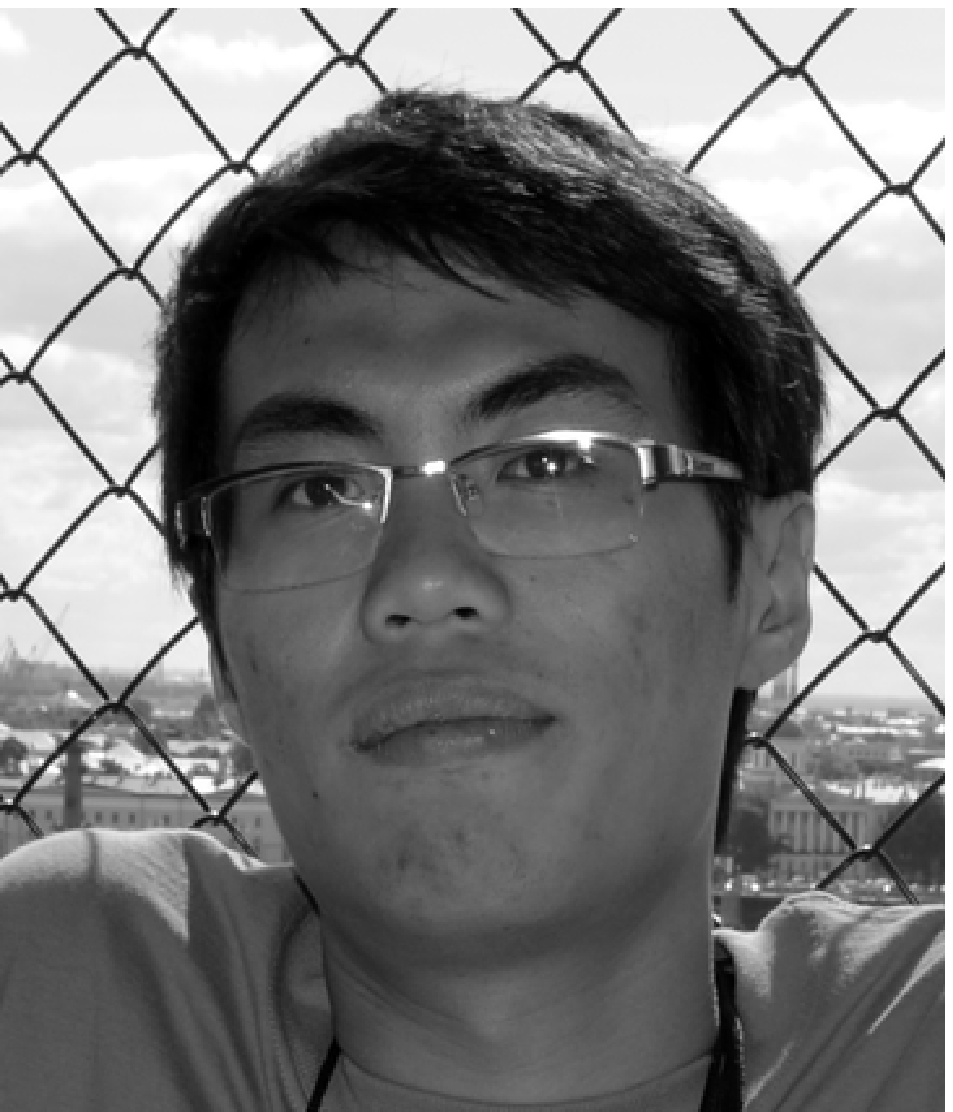}}]{Ruiyang Tan} is studying for a Bachelor's degree in Department of Computer Science, Sun Yat-sen University. He has participated in ACM/ICPC twice times and won two Asia regional champions.
\end{IEEEbiography}
\vspace{-15 mm}
\vfill
\begin{IEEEbiography}[{\includegraphics[width=1in,height=1.25in,clip,keepaspectratio]{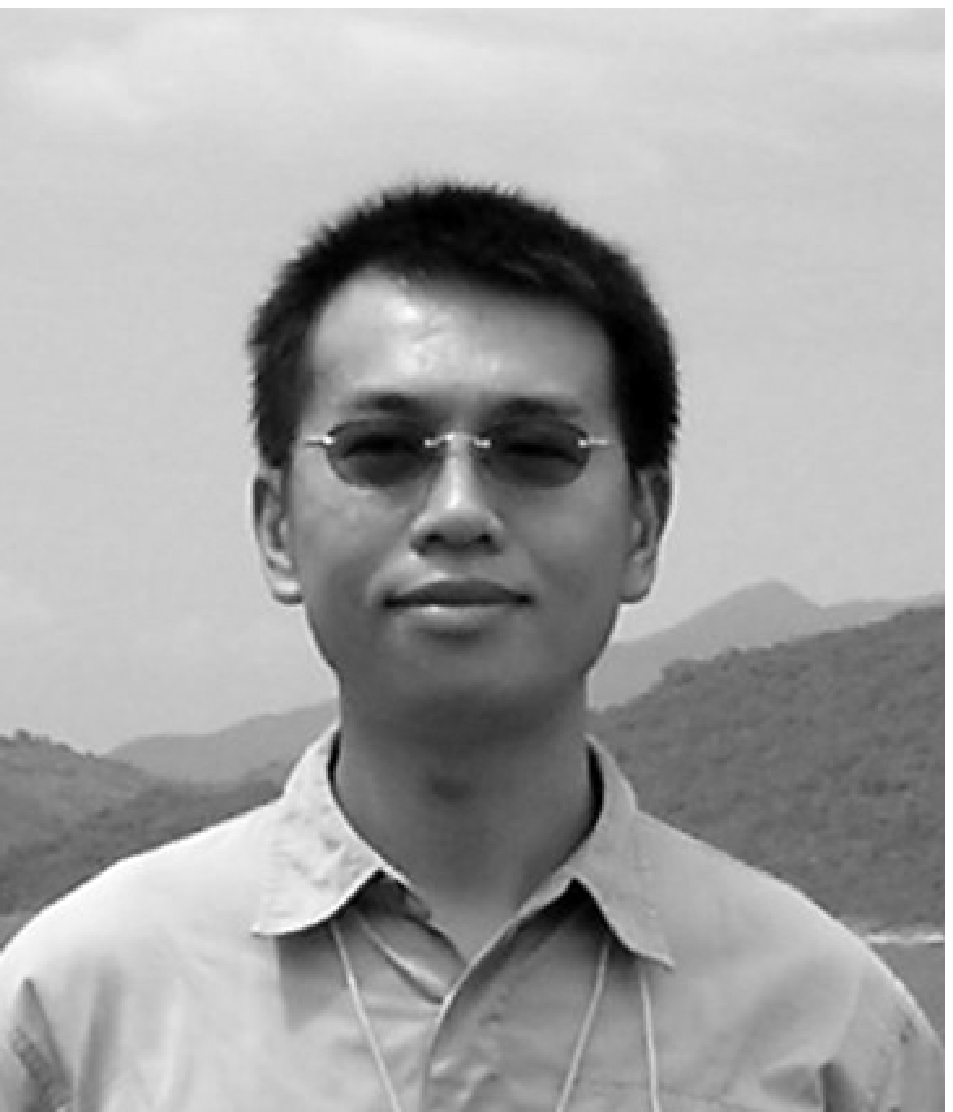}}]{Rong Pan} received the BSc and PhD degrees in applied mathematics from Sun Yat-sen University, China, in 1999 and 2004, respectively. He was a postdoctoral fellow at the Hong Kong University of Science and Technology (2005~2007) and HP Labs (2007~2009). Since then, he has been a faculty member of Department of Computer Science in Sun Yat-sen University. His research interest includes text mining, recommender systems, data mining, and machine learning.
\end{IEEEbiography}

\onecolumn
\appendices
\section{Tag-Weighted Topic Model}
In the topic models, the key inferential problem that we need to solve is to compute the posterior distribution of the hidden variables given a document $d$. Given the document $d$, we can easily get the posterior distribution of the latent variables in the proposed model, as:
{\normalsize
\begin{equation*}
\begin{aligned}
p(\varepsilon^{d}, \mathbf{z} | \mathbf{w}^{d}, T^{d}, \theta, \eta,\psi,\pi)
= \frac{p(\varepsilon^{d}, \mathbf{z}, \mathbf{w}^{d}, T^{d} | \theta, \eta,\psi,\pi)}{p(\mathbf{w}^{d}, T^{d} | \theta, \eta,\psi,\pi)}.
\end{aligned}
\end{equation*}
}
Integrating over $\varepsilon$ and summing out $z$, we easily obtain the marginal distribution of $d$:
{\normalsize
\begin{equation*}
\begin{aligned}
&p(\mathbf{w}^{d}, T^{d} | \eta,\theta,\psi,\pi) 
=  p(\mathbf{t}^{d} | \eta) \int p\left (\varepsilon^{d} | (T^{d} \times \pi) \right ) \cdot \prod_{i=1}^N \sum_{z^{d}_i=1}^K p(z^{d}_i | (\varepsilon^{d})^\mathrm{T} \times T^{d} \times \theta)
  p(w^{d}_i | z^{d}_i,\psi_{1:K}) ~d\varepsilon^{d}. 
\end{aligned}
\end{equation*}
}

In this work, we make use of mean-field variational EM algorithm to efficiently obtain an approximation of this posterior distribution of the latent variables. In the mean-field variational inference, we minimize the KL divergence between the variational posterior probability and the true posterior probability through by maximizing the evidence lower bound (ELBO) $\mathcal{L}(\cdot)$. For a single document $d$, we obtain the $\mathcal{L}(\cdot)$ using Jensen's inequality:
{\normalsize
\begin{equation*}
\begin{aligned}
\mathcal{L}(\xi_{1:l^{d}}, \gamma_{1:K}; \eta_{1:L}, \pi_{1:L}, \theta_{1:L}, \psi_{1:K}) &= E[\log p(T_{1:l^{d}} | \eta_{1:L})] + E[\log p(\varepsilon^{d}| T^{d} \times \pi)]  \\
&+\sum_{i=1}^N E[\log p(z_i | (\varepsilon^{d})^\mathrm{T} \times T^{d} \times \theta)] +\sum_{i=1}^N E[\log p(w_i | z_i, \psi_{1:K})] + H(q),
\end{aligned}
\end{equation*}
}
where $\xi$ is a $l^{d}$-dimensional Dirichlet parameter vector and $\gamma$ is $1 \times K $ vector, both of which are variational parameters of variational distribution. $H(q)$ indicates the entropy of the variational distribution:
{\normalsize
\begin{equation*}
\begin{aligned}
H(q) = - E[\log q(\varepsilon^{d})] - E[\log q(z)].
\end{aligned}
\end{equation*}
}
Here the exception is taken with respect to a variational distribution $q(\varepsilon^{d}, z_{1:N})$, and we choose the following fully factorized distribution: 
{\normalsize
\begin{equation*}
\begin{aligned}
&q(\varepsilon^{d}, z_{1:N} | \xi_{1:L}, \gamma_{1:K} ) = q( \varepsilon^{d} | \xi ) \prod_{i=1}^N q(z_i | \gamma_i),
\end{aligned}
\end{equation*}
}
where the dimension of parameter $\xi$ is changed with different documents.

In the $\mathcal{L}(\cdot)$,
{\normalsize
\begin{equation*}
\begin{aligned}
&E[\log p(z_i | (\varepsilon^{d})^\mathrm{T}\times T^{d}\times \theta)]=\sum_{k=1}^K \gamma_{ik} E[\log((\varepsilon^{d})^\mathrm{T}\times T^{d}\times \theta)_k].
\end{aligned}
\end{equation*}
}

To preserve the lower bound on the log probability, we upper bound the log normalizer in $E[\log((\varepsilon^{d})^\mathrm{T}\times T^{d}\times \theta)_k]$ using Jensen's inequality again:
{\normalsize
\begin{equation*}
\begin{aligned}
&E[\log((\varepsilon^{d})^\mathrm{T} \times T^{d} \times \theta)_k] 
= E[\log \sum_{i=1}^{l^{d}}\varepsilon_i^{d} \theta_k^{(i)}] \geq E[\sum_{i=1}^{l^{d}}\varepsilon_i^{d} \log \theta_k^{(i)}]
= \sum_{i=1}^{l^{d}} \log \theta_k^{(i)} E[\varepsilon_i^{d}],
\end{aligned}
\end{equation*}
}
where the expression of $\theta^{(i)}$ , $i \in \{ 1, \cdots, l^{d} \}$, means the $i$-th tag's topic assignment vector, corresponding to the $i$-th row of $\Theta^{d}$.
Note that the expectation of Dirichlet random variable is 
$
E[\varepsilon_i^{d}] = \frac{\xi_i}{ \sum_{j=1}^{l^{d}} \xi_j}.
$

Thus, for the document $d$,
{\normalsize
\begin{equation*}
\begin{aligned}
&\sum_{i=1}^N E[\log p(z_i | (\varepsilon^{d})^\mathrm{T} \times T^{d} \times \theta)] = \sum_{i=1}^N \sum_{k=1}^K \gamma_{ik} \cdot \sum_{j=1}^{l^{d}} \log \theta_k^{(j)} \frac{\xi_j}{ \sum_{j^{'}=1}^{l^{d}} \xi_{j^{'}} }.
\end{aligned}
\end{equation*}
}

Finally, we expand $\mathcal{L}(\cdot)$ in terms of the model parameters ($\eta, \pi, \theta, \psi$) and the variational parameters ($\xi,\gamma$) as follows:
{\normalsize
\begin{equation*}
\begin{aligned}
\mathcal{L}(\xi, \gamma; \eta, \pi, \theta, \psi) &= \sum_{l=1}^L (t_l^d \log \eta_l^d + (1-t_l^d) \log (1-\eta_l^d)) \\
&+ \log \Gamma (\sum_{i=1}^{l^{d}}{(T^{d} \times \pi)}_i) - \sum_{i=1}^{l^{d}}\log \Gamma \left({(T^{d} \times \pi)}_i \right) + \sum_{i=1}^{l^{d}} ({(T^{d} \times \pi)}_i - 1) \left(\Psi(\xi_i) - \Psi(\sum_{j=1}^{l^{d}} \xi_j) \right) \\
&+ \sum_{i=1}^N \sum_{k=1}^K \gamma_{ik} \sum_{j=1}^{l^{d}} \log \theta_k^{(j)} \frac{\xi_j}{ \sum_{j^{'}=1}^{l^{d}} \xi_{j^{'}} }  + \sum_{i=1}^N \sum_{k=1}^K \sum_{j=1}^V \gamma_{ik} ({w^{d}})_i^{j} \log \psi_{kj} \\
& - \log \Gamma(\sum_{i=1}^{l^{d}}{\xi_i}) + \sum_{i=1}^{l^{d}} \log \Gamma(\xi_i) - \sum_{i=1}^{l^{d}} (\xi_i - 1) \left(\Psi(\xi_i) - \Psi(\sum_{j^{'}=1}^{l^{d}} \xi_{j^{'}})\right) - \sum_{i=1}^N \sum_{k=1}^K \gamma_{ik}^d \log \gamma_{ik}^d.
\end{aligned}
\end{equation*}
}
Then, we maximize the lower bound $\mathcal{L}(\xi, \gamma; \eta, \pi, \theta, \psi)$ with respect to the variational parameters $\xi$ and $\gamma$, using a variational expectation-maximization(EM) procedure as follows.
\subsection{Variational E-step}
\subsubsection{$\xi$}
We first maximize $\mathcal{L}(\cdot)$ with respect to $\xi_i$ for the document $d$. Maximize the terms which contain $\xi$:
{\normalsize
\begin{equation*}
\begin{aligned}
\mathcal{L}_{[\xi]} &=\sum_{i=1}^{l^{d}}( \sum_{l^{'}=1}^L \pi_{l^{'}} T_{i{l^{'}}}^{d} - 1) \left(\Psi(\xi_i) - \Psi(\sum_{j^{'}=1}^{l^{d}} \xi_{j^{'}})\right) +\sum_{i=1}^N \sum_{k=1}^K \gamma_{ik} \cdot \sum_{j=1}^{l^{d}} \log \theta_k^{(j)} \frac{\xi_j}{ \sum_{j^{'}=1}^{l^{d}} \xi_{j^{'}} } - \log \Gamma(\sum_{i=1}^{l^{d}}{\xi_i}) \\
& + \sum_{i=1}^{l^{d}} \log \Gamma(\xi_i) - \sum_{i=1}^{l^{d}} (\xi_i - 1) \left( \Psi(\xi_i) - \Psi(\sum_{j^{'}=1}^{l^{d}} \xi_{j^{'}})\right),
\end{aligned}
\end{equation*}
}
where $\Psi(\cdot)$ denotes the digamma function, the first derivative of the log of the Gamma function, and ${(T^{d} \times \pi)}_i$ = $\sum_{i=1}^{l^{d}} \sum_{l=1}^L \pi_l T_{il}^{d}$. The derivative of $\mathcal{L}_{[\xi]}$ with respect to $\xi_i$ is
{\small
\begin{equation*}
\begin{aligned}
\mathcal{L}^{'}(\xi_i) &=\Psi^{'}(\xi_i)(\sum_{l=1}^L \pi_l T_{il}^{d} - \xi_i) - \Psi^{'}(\sum_{j=1}^{l^{d}} \xi_j)\cdot \sum_{i=1}^{l^{d}}(\sum_{l=1}^L \pi_l T_{il}^{d} - \xi_i) + \sum_{i^{'}=1}^N \sum_{k=1}^K \gamma_{i^{'}k}^{d}\cdot \left (  \frac{\log \theta_k^{(i)} (\sum_{j=1}^{l^{d}} \xi_j) - \sum_{j=1}^{l^{d}} \log \theta_k^{(j)} \xi_j}{{\sum_{j'=1}^{l^{d}} \xi_{j'} }^2} \right).
\end{aligned}
\end{equation*}
}
Here we use gradient descent method to find the $\xi$ to make the maximization of $\mathcal{L}_{[\xi]}$.
\subsubsection{$\gamma$}
Next, we maximize $\mathcal{L}(\cdot)$ with respect to $\gamma_{ik}$. Adding the Lagrange multipliers to the terms which contain $\gamma_{ik}$, we get the following equation:
{\normalsize
\begin{equation*}
\begin{aligned}
\mathcal{L}_{[\gamma]} &= \sum_{i=1}^N \sum_{k=1}^K \gamma_{ik} \sum_{j=1}^{l^{d}} \log \theta_k^{(j)} \frac{\xi_j}{ \sum_{j^{'}=1}^{l^{d}} \xi_{j^{'}} } + \sum_{i=1}^N \sum_{k=1}^K \sum_{j=1}^V \gamma_{ik} ({w^{d}})_i^{j} \log \psi_{kj} - \sum_{i=1}^N \sum_{k=1}^K \gamma_{ik}^d \log \gamma_{ik}^d + \sum_{i=1}^N \lambda_i(\sum_{k=1}^K \gamma_{ik} -1).
\end{aligned}
\end{equation*}
}

By taking the derivative with respect to $\gamma_{ik}$, and setting the derivative to zero yields, we obtain the update equation of $\gamma_{ik}$:
{\normalsize
\begin{equation*}
\begin{aligned}
&\gamma_{ik} \propto \psi_{k,v^{w_i}} \exp\{ \sum_{j=1}^{l^{d}} \log \theta_k^{(j)} \frac{\xi_j}{ \sum_{j'=1}^{l^{d}} \xi_{j'} } \},
\end{aligned}
\end{equation*}
}
where $v^{w_i}$ denotes the index of $w_i$ in the dictionary.
\subsection{M-step estimation}
The M-step needs to update four parameters: $\eta$, the tagging prior probability, $\pi$, the Dirichlet prior of the tags' weights, $\theta$, the topic distribution over all tags in the corpus, and $\psi$, the probability of a word under a topic.
\subsubsection{$\eta$}
For a given corpus, the $\eta_i$ is estimated by adding up the number of $i^{th}$ label which appears in all documents. It does not depend any parameter in the proposed model, except itself. 
By maximizing the terms which contain $\eta$, we have
{\normalsize
\begin{equation*}
\begin{aligned}
\eta_l = \frac{\sum_d^D t_l^d}{D},
\end{aligned}
\end{equation*}
}
where $D$ is the size of corpus. Because each document's tags-set is observed, the Bernoulli prior $\eta$ is unused, which is included for model completeness.
\subsubsection{$\pi$}
For the document $d$, the terms that involve the Dirichlet prior $\pi$:
{\normalsize
\begin{equation*}
\begin{aligned}
\mathcal{L}_{[\pi]} =
\log \Gamma(\sum_{i=1}^{l^{d}}{(T^{d} \times \pi)}_i) - \sum_{i=1}^{l^{d}}\log \Gamma \left({(T^{d} \times \pi)}_i \right) \nonumber + \sum_{i=1}^{l^{d}} \left({(T^{d} \times \pi)}_i - 1 \right) \left(\Psi(\xi_i) - \Psi(\sum_{j=1}^{l^{d}} \xi_j) \right).
\end{aligned}
\end{equation*}
}
We use gradient descent method by taking derivative of $\mathcal{L}_{[\pi]}$ with respect to $\pi_l$ on the whole corpus to find the estimation of $\pi$. Taking derivatives with respect to $\pi_l$ on the corpus, we obtain:
{\normalsize
\begin{equation*}
\begin{aligned}
\mathcal{L}^{'}_{[\pi_l]} = \sum_{d=1}^D \Psi( \sum_{i=1}^{l^{d}} \sum_{l^{'}=1}^L \pi_{l^{'}} \cdot T_{i{l^{'}}}^{d}) \cdot \sum_{i=1}^{l^{d}}T_{il}^{d} - \sum_{d=1}^D \sum_{i=1}^{l^{d}} \Psi( \sum_{l^{'}=1}^L \pi_{l^{'}} \cdot T_{i{l^{'}}}^{d}) \cdot T_{il}^{d} + \sum_{d=1}^D \sum_{i=1}^{l^{d}} \left(\Psi(\xi_i) -\Psi(\sum_{j=1}^{l^{d}})\right) \cdot T_{il}^{d}.
\end{aligned}
\end{equation*}
}
\subsubsection{$\theta$}
The only term that involves $\theta$ is:
{\normalsize
\begin{equation*}
\begin{aligned}
\mathcal{L}_{[\theta]} = \sum_{d=1}^D \sum_{i=1}^N \sum_{k=1}^K \gamma_{ik} \sum_{j=1}^{l^{d}} \log \theta_k^{(j)} \frac{\xi_j}{ \sum_{j^{'}=1}^{l^{d}} \xi_{j^{'}} },
\end{aligned}
\end{equation*}
}
where $\xi_j$, $j \in \{ 1, \cdots, l^{d} \}$ in the document $d$ needs to be extended to ${t}^{d}_l \cdot \xi_l^{d}$, $ l \in \{ 1, \cdots, L \}$ for convenient to simplify $\mathcal{L}_{[\theta]}$. With the Lagrangian of the $\mathcal{L}_{[\theta]}$, which incorporate the constraint that the K-components of $\theta_l$ sum to one, adding $\sum_{l=1}^L \lambda_l (\sum_{k=1}^K \theta_{lk} - 1)$ to $\mathcal{L}_{[\theta]}$, taking the derivative with respect to $\theta_{lk}$, and setting the derivative to zero yields, we obtain the estimation of $\theta$ over the whole corpus,
{\normalsize
\begin{equation*}
\begin{aligned}
&\theta_{lk} \propto \sum_{d=1}^D \sum_{i=1}^N \gamma_{ik}^{d} \frac{\xi_l^{d} {t}^{d}_l}{\sum_{l=1}^L ( \xi_l^{d} {t}^{d}_l )}.
\end{aligned}
\end{equation*}
}
\subsubsection{$\psi$}
To maximize with respect to $\psi$, we isolate corresponding terms and add Lagrange multipliers:
{\normalsize
\begin{equation*}
\begin{aligned}
&\mathcal{L}_{[\psi]} =\sum_{d=1}^D \sum_{i=1}^N \sum_{k=1}^K \sum_{j=1}^V \gamma_{ik} ({w^{d}})_i^{j} \log \psi_{kj}+ \sum_{k=1}^K \lambda_k(\sum_{j=1}^v \psi_{kj} -1).
\end{aligned}
\end{equation*}
}
Take the derivative with respect to $\psi_{kj}$, and set it to zero, we get:
{\normalsize
\begin{equation*}
\begin{aligned}
&\psi_{kj} \propto \sum_{d=1}^D \sum_{i=1}^N \gamma_{ik}^{d} ({w^{d}})_i^{j}.\label{eq:psi}
\end{aligned}
\end{equation*}
}
\newpage
\section{Tag-Weighted Dirichlet Allocation}
In TWDA, we treat $\pi$, $\mu$, $\eta$, $\theta$ and $\psi$ as unknown constants to be estimated, and use a variational expectation-maximization (EM) procedure to carry out approximate maximum likelihood estimation as TWTM.
Given the document $d$, we can easily get the posterior distribution of the latent variables in the TWDA model, as:
{\normalsize
\begin{equation*}
\begin{aligned}
p(\varepsilon^{d}, \lambda^{d}, \mathbf{z} | \mathbf{w}^{d}, T^{d}, \theta, \eta,\psi,\pi, \mu)
= \frac{p(\varepsilon^{d}, \lambda^{d}, \mathbf{z}, \mathbf{w}^{d}, T^{d} | \theta, \eta,\psi,\pi, \mu)}{p(\mathbf{w}^{d}, T^{d} | \theta, \eta,\psi,\pi, \mu)}.\label{eq:pd}
\end{aligned}
\end{equation*}
}

As with TWTM, it is not efficiently computable. We maximize the evidence lower bound(ELBO) $\mathcal{L}(\cdot)$using Jensen's inequality, and for a document $d$ we have the form:
{\normalsize
\begin{equation*}
\begin{aligned}
\mathcal{L}(\xi_{1:l^{d}}, \gamma_{1:K}, \rho_{1:K}; \eta_{1:L}, \pi_{1:L}, \mu_{1:K}, \theta_{1:L}, \psi_{1:K}) &= E[\log p(T_{1:l^{d}} | \eta_{1:L})] + E[\log p(\varepsilon^{d}| T^{d} \times \pi)] + E[\log p(\lambda^{d} | \mu)] \\
& + \sum_{i=1}^N E[\log p(z_i | (\varepsilon^{d})^\mathrm{T} \times T^{d} \times (\frac{ \theta}{\lambda}))] + \sum_{i=1}^N E[\log p(w_i | z_i, \psi_{1:K})] \\
& + H(q),
\end{aligned}
\end{equation*}
}
where $\xi$ is a $l^{d}$-dimensional Dirichlet parameter vector, $\rho$ is a $1 \times K$ vector and $\gamma$ is $1 \times K $ vector, all of which are variational parameters of variational distribution. Unlike the TWTM, $l^{d}$in TWDA is one more than the number of the observed tags in the document $d$. $H(q)$ indicates the entropy of the variational distribution:
{\normalsize
\begin{equation*}
\begin{aligned}
H(q) = - E[\log q(\varepsilon^{d})] - E[\log q(\lambda)] - E[\log q(z)].
\end{aligned}
\end{equation*}
}
Here the exception is taken with respect to a variational distribution $q(\varepsilon^{d}, q(\lambda^{d}), z_{1:N})$, and we choose the following fully factorized distribution: 
{\normalsize
\begin{equation*}
\begin{aligned}
&q(\varepsilon^{d}, \lambda^{d}, z_{1:N} | \xi_{1:L}, \rho_{1:K}, \gamma_{1:K} )=q( \varepsilon^{d} | \xi ) q(\lambda^{d} | \rho ) \prod_{i=1}^N q(z_i | \gamma_i).
\end{aligned}
\end{equation*}
}

The term of the expected log probability of the topic assignment:
{\normalsize
\begin{equation*}
\begin{aligned}
&E[\log p(z_i | (\varepsilon^{d})^\mathrm{T} \times T^{d} \times (\frac{ \theta}{\lambda}))] = \sum_{k=1}^K \gamma_{ik} E[\log((\varepsilon^{d})^\mathrm{T} \times T^{d} \times (\frac{ \theta}{\lambda}))_k],
\end{aligned}
\end{equation*}
}
which could be difficult to compute, because of tag-weighted topic assignment which is used in TWDA. Thus we use Jensen's inequality:
{\normalsize
\begin{equation*}
\begin{aligned}
E[\log((\varepsilon^{d})^\mathrm{T} \times T^{d} \times (\frac{ \theta}{\lambda}))_k] &= E[\log( \sum_{i=1}^{l^{d}-1}\varepsilon_i^{d} \theta_k^{(i)} + \varepsilon_{l^d}^d \lambda_k)] \\
& \geq E[\sum_{i=1}^{l^{d}-1}\varepsilon_i^{d} \log \theta_k^{(i)} + \varepsilon_{l^d}^d \cdot\log \lambda_k] \\
& = \sum_{i=1}^{l^{d}-1} \log \theta_k^{(i)} E[\varepsilon_i^{d}] + E[\varepsilon_{l^d}^d \cdot \log \lambda_k],
\end{aligned}
\end{equation*}
}
where the expression of $\theta^{(i)}$, $i \in \{ 1, \cdots, l^{d}-1 \}$, means the $i$-th tag's topic assignment vector, corresponding to the $i$-th row of $\Theta^{d}$.

because the variational distribution is fully factorized, so we can get:
{\normalsize
\begin{equation*}
\begin{aligned}
&E[\log((\varepsilon^{d})^\mathrm{T} \times T^{d} \times (\frac{ \theta}{\lambda}))_k] = \sum_{i=1}^{l^{d}-1} \log \theta_k^{(i)} E[\varepsilon_i^{d}] + E[\varepsilon_{l^d}^d] \cdot E[\log \lambda_k],
\end{aligned}
\end{equation*}
}
where
{\normalsize
\begin{equation*}
\begin{aligned}
&E[\varepsilon_{l^d}^d] = {\xi_{l^d}}/{ \sum_{j=1}^{l^{d}} \xi_j}, \\
&E[\log \lambda_k] = \Psi(\rho_k) - \Psi(\sum_{j^{'}=1}^{K} \rho_{j^{'}}).
\end{aligned}
\end{equation*}
}

With $E[\varepsilon_i^{d}] = \frac{\xi_i}{ \sum_{j=1}^{l^{d}} \xi_j}$, Thus, for the document $d$,
{\normalsize
\begin{equation*}
\begin{aligned}
& \sum_{i=1}^N E[\log p(z_i | (\varepsilon^{d})^\mathrm{T} \times T^{d} \times (\frac{ \theta}{\lambda}))]=\sum_{i=1}^N \sum_{k=1}^K \gamma_{ik} \cdot [\sum_{j=1}^{l^{d}-1} \log \theta_k^{(j)} \frac{\xi_j}{\sum_{j^{'}=1}^{l^{d}} \xi_{j^{'}}} + ( \Psi(\rho_k) - \Psi(\sum_{j^{'}=1}^{K} \rho_{j^{'}}) )\frac{\xi_{l^d}}{ \sum_{j=1}^{l^{d}} \xi_j}].
\end{aligned}
\end{equation*}
}

Then we expand the $\mathcal{L}(\cdot)$ of TWDA as follows:
{\normalsize
\begin{equation*}
\begin{aligned}
\mathcal{L}(\xi, \gamma, \rho; \eta, \pi, \mu, \theta, \psi) &= \sum_{l=1}^L(t_l^d \log \eta_l^d + (1-t_l^d) \log (1-\eta_l^d)) \\
& + \log \Gamma (\sum_{i=1}^{l^{d}}{(T^{d} \times \pi)}_i ) - \sum_{i=1}^{l^{d}}\log \Gamma ({(T^{d} \times \pi)}_i) + \sum_{i=1}^{l^{d}} ({(T^{d} \times \pi)}_i - 1)\left(\Psi(\xi_i) - \Psi(\sum_{j=1}^{l^{d}} \xi_j)\right) \\
& + \log \Gamma(\sum_{j=1}^K \mu_j) - \sum_{i=1}^K \log \Gamma(\mu_i) + \sum_{i=1}^K(\mu_i -1)\left(\Psi(\rho_i^d) - \Psi(\sum_{j=1}^K \rho_j^d)\right) \\
& + \sum_{i=1}^N \sum_{k=1}^K \gamma_{ik} \cdot \sum_{j=1}^{l^{d}} C_k^{(j)} \frac{\xi_j}{ \sum_{j^{'}=1}^{l^{d}} \xi_{j^{'}}} + \sum_{i=1}^N \sum_{k=1}^K \sum_{j=1}^V \gamma_{ik} ({w^{d}})_i^{j} \log \psi_{kj} \\
& - \log \Gamma(\sum_{i=1}^{l^{d}}{\xi_i}) + \sum_{i=1}^{l^{d}} \log \Gamma(\xi_i) - \sum_{i=1}^{l^{d}} (\xi_i - 1)\left(\Psi(\xi_i) - \Psi(\sum_{j^{'}=1}^{l^{d}} \xi_{j^{'}})\right) \\
& - \sum_{i=1}^N \sum_{k=1}^K \gamma_{ik} \cdot \log \gamma_{ik} \\
& - \log \Gamma(\sum_{j=1}^K \rho_j) + \sum_{i=1}^K \log \Gamma(\rho_i) - \sum_{i=1}^K(\rho_i -1)\left(\Psi(\rho_i) - \Psi(\sum_{j=1}^K \rho_j)\right).
\end{aligned}
\end{equation*}
}
where 
{\normalsize
\begin{equation*}
\begin{aligned}
&C_k^{(j)}=
\begin{cases}
\log \theta_k^{(j)} & j \in \{ 1, \cdots, l^{d}-1 \}\\
\Psi(\rho_k) - \Psi(\sum_{j^{'}=1}^{K} \rho_{j^{'}}) & j=l^d
\end{cases},
\end{aligned}
\end{equation*}
}
and
{\normalsize
\begin{equation*}
\begin{aligned}
&{(T^{d} \times \pi)}_i = \sum_{l=1}^{L+1} \pi_l T_{il}^{d}.
\end{aligned}
\end{equation*}
}

\subsection{Variational E-step}
For a single document $d$, the variational parameters include $\xi^d$, $\rho^d$ and $\gamma_{ik}$. First, we maximize $\mathcal{L}(\cdot)$ with respect to the variational parameters to obtain an estimate of the posterior. 
\subsubsection{Optimization with respect to $\xi$}
We first maximize $\mathcal{L}(\cdot)$ with respect to $\xi_i$ for the document $d$. Maximize the terms which contain $\xi$:
{\normalsize
\begin{equation*}
\begin{aligned}
\mathcal{L}_{[\xi]} &= 
\sum_{i=1}^{l^{d}}( \sum_{l^{'}=1}^{L+1} \pi_{l^{'}} T_{i{l^{'}}}^{d} - 1)\left(\Psi(\xi_i) - \Psi(\sum_{j^{'}=1}^{l^{d}} \xi_{j^{'}})\right) +\sum_{i=1}^N \sum_{k=1}^K \gamma_{ik} \cdot \sum_{j=1}^{l^{d}} C_k^{(j)} \frac{\xi_j}{ \sum_{j^{'}=1}^{l^{d}} \xi_{j^{'}}} - \log \Gamma(\sum_{i=1}^{l^{d}}{\xi_i}) \\ 
& + \sum_{i=1}^{l^{d}} \log \Gamma(\xi_i) - \sum_{i=1}^{l^{d}} (\xi_i - 1) \left(\Psi(\xi_i) - \Psi(\sum_{j^{'}=1}^{l^{d}} \xi_{j^{'}})\right),
\end{aligned}
\end{equation*}
}
The derivative of $\mathcal{L}_{[\xi]}$ with respect to $\xi_i$ is
{\normalsize
\begin{equation*}
\begin{aligned}
\mathcal{L}^{'}(\xi_i) &=\Psi^{'}(\xi_i)(\sum_{l=1}^{L+1} \pi_l T_{il}^{d} - \xi_i) - \Psi^{'}(\sum_{j=1}^{l^{d}} \xi_j)\sum_{i=1}^{l^{d}}(\sum_{l=1}^{L+1} \pi_l T_{il}^{d} - \xi_i) + \sum_{i^{'}=1}^N \sum_{k=1}^K \gamma_{i^{'}k}^{d} \cdot \left(\frac{ C_k^{(i)} (\sum_{j=1}^{l^{d}} \xi_j) - \sum_{j=1}^{l^{d}} C_k^{(j)} \xi_j}{(\sum_{j'=1}^{l^{d}} \xi_{j'} )^2}\right).
\end{aligned}
\end{equation*}
}
Here we use gradient descent method to find the $\xi$ to make the maximization of $\mathcal{L}_{[\xi]}$.

\subsubsection{Optimization with respect to $\rho$}
Next, we maximize $\mathcal{L}(\cdot)$ with respect to $\rho$. The terms that involve the variational Dirichlet $\rho$ are:
{\normalsize
\begin{equation*}
\begin{aligned}
\mathcal{L}_{[\rho]} &= \sum_{i=1}^K (\mu_i - 1)\left(\Psi(\rho_i) - \Psi(\sum_{j=1}^K \rho_j)\right) - \log \Gamma(\sum_{j=1}^K \rho_j) + \sum_{i=1}^K \log \Gamma(\rho_i) - \sum_{i=1}^K(\rho_i -1)\left(\Psi(\rho_i) - \Psi(\sum_{j=1}^K \rho_j)\right) \\
&+ \sum_{k=1}^K \sum_{i=1}^N \gamma_{ik} \cdot \frac{\xi_{l^d}}{\sum_{j=1}^{l^d} \xi_j} \cdot \left(\Psi(\rho_k) - \Psi(\sum_{j=1}^K \rho_j)\right).
\end{aligned}
\end{equation*}
}
This simplifies to:
{\normalsize
\begin{equation*}
\begin{aligned}
\mathcal{L}_{[\rho]} &= \sum_{i=1}^K\left(\Psi(\rho_i) - \Psi(\sum_{j=1}^K \rho_j)\right) \cdot \left(\mu_i - \rho_i + \sum_{n=1}^N \gamma_{ni} \cdot \frac{\xi_{l^d}}{\sum_{j=1}^{l^d} \xi_j}\right) - \log \Gamma(\sum_{j=1}^K \rho_j) + \sum_{i=1}^K \log \Gamma(\rho_i).
\end{aligned}
\end{equation*}
}
Taking the derivative with respect to $\rho_i$ and setting it to zero, we obtain a maximum at:
{\normalsize
\begin{equation*}
\begin{aligned}
\rho_i = \mu_i + \sum_{n=1}^N \gamma_{ni} \cdot \frac{\xi_{l^d}}{\sum_{j=1}^{l^d} \xi_j}.
\end{aligned}
\end{equation*}
}

\subsubsection{Optimization with respect to $\gamma$}
The terms that contain $\gamma$ are:
{\normalsize
\begin{equation*}
\begin{aligned}
\mathcal{L}_{[\gamma]} &= \sum_{i=1}^N \sum_{k=1}^K \gamma_{ik} \sum_{i=1}^{l^d} C_k^{(i)} \cdot \frac{\xi_i}{\sum_{j=1}^{l^d} \xi_j} + \sum_{i=1}^N \sum_{k=1}^K  \sum_{j=1}^V \gamma_{ik} w_{ij} \log \psi_{k,v^{w_i}} - \sum_{i=1}^N \sum_{k=1}^K \gamma_{ik} \cdot \log \gamma_{ik}
\end{aligned}
\end{equation*}
}
Adding the Lagrange multipliers to the terms which contain $\gamma_{ik}$, taking the derivative with respect to $\gamma_{ik}$, and setting the derivative to zero yields, we obtain the update equation of $\gamma_{ik}$:
{\normalsize
\begin{equation*}
\begin{aligned}
&\gamma_{ik} \propto \psi_{k,v^{w_i}} \exp\{ \sum_{j=1}^{l^{d}} C_k^{(j)} \frac{\xi_j}{ \sum_{j'=1}^{l^{d}} \xi_{j'} } \},
\end{aligned}
\end{equation*}
}
where $v^{w_i}$ denotes the index of $w_i$ in the dictionary.

In E-step, we update the $\xi$, $\rho$ and $\gamma$ for each document with the initialized model parameters.

\subsection{M-step estimation}
The M-step needs to update five parameters: $\eta$, the tagging prior probability, $\pi$, the Dirichlet prior of the tags' weights, $\theta$, the topic distribution over all tags in the corpus, $\psi$, the probability of a word under a topic, and $\mu$, a Dirichlet prior of model. It is worthy to note that we update $\eta$ with the same method as in TWTM.
\subsubsection{Optimization with respect to $\pi$}
For the document $d$, the terms that involve the Dirichlet prior $\pi$:
{\normalsize
\begin{equation*}
\begin{aligned}
\mathcal{L}_{[\pi]} &= 
\log \Gamma (\sum_{i=1}^{l^{d}}{(T^{d} \times \pi)}_i ) - \sum_{i=1}^{l^{d}}\log \Gamma ({(T^{d} \times \pi)}_i) + \sum_{i=1}^{l^{d}} ({(T^{d} \times \pi)}_i - 1)\left(\Psi(\xi_i) - \Psi(\sum_{j=1}^{l^{d}} \xi_j)\right),
\end{aligned}
\end{equation*}
}
where ${(T^{d} \times \pi)}_i$ = $\sum_{l=1}^{L+1} \pi_l T_{il}^{d}$.
We use gradient descent method by taking derivative of $\mathcal{L}_{[\pi]}$ with respect to $\pi_l$ on the corpus to find the estimation of $\pi$.
Taking derivatives with respect to $\pi_l$ on the whole corpus, we obtain:
{\normalsize
\begin{equation*}
\begin{aligned}
\mathcal{L}^{'}_{[\pi_l]}&=\sum_{d=1}^D \Psi( \sum_{i=1}^{l^{d}} \sum_{l^{'}=1}^{L+1} \pi_{l^{'}} \cdot T_{i{l^{'}}}^{d}) \cdot \sum_{i=1}^{l^{d}}T_{il}^{d} - \sum_{d=1}^D \sum_{i=1}^{l^{d}} \Psi( \sum_{l^{'}=1}^{L+1} \pi_{l^{'}} \cdot T_{i{l^{'}}}^{d}) \cdot T_{il}^{d} + \sum_{d=1}^D \sum_{i=1}^{l^{d}}\left(\Psi(\xi_i) -\Psi(\sum_{j=1}^{l^{d}} \xi_j)\right) \cdot T_{il}^{d}.
\end{aligned}
\end{equation*}
}

\subsubsection{Optimization with respect to $\theta$}
The only term that involves $\theta$ is:
{\normalsize
\begin{equation*}
\begin{aligned}
\mathcal{L}_{[\theta]} = \sum_{d=1}^D \sum_{i=1}^N \sum_{k=1}^K \gamma_{ik} \sum_{j=1}^{l^{d}} \log \theta_k^{(j)} \frac{\xi_j}{ \sum_{j^{'}=1}^{l^{d}} \xi_{j^{'}} },
\end{aligned}
\end{equation*}
}
where $\xi_j$, $j \in \{ 1, \cdots, l^{d} \}$ in the document $d$ needs to be extended to ${t}^{d}_l \cdot \xi_l^{d}$, $ l \in \{ 1, \cdots, L+1 \}$ for convenient to simplify $\mathcal{L}_{[\theta]}$.
With the Lagrangian of the $\mathcal{L}_{[\theta]}$, which incorporate the constraint that the K-components of $\theta_l$ sum to one, we obtain the estimation of $\theta$ over the whole corpus,
{\normalsize
\begin{equation*}
\begin{aligned}
&\theta_{lk} \propto \sum_{d=1}^D \sum_{i=1}^N \gamma_{ik}^{d} \frac{\xi_l^{d} {t}^{d}_l}{\sum_{l=1}^{L+1} ( \xi_l^{d} {t}^{d}_l )}.
\end{aligned}
\end{equation*}
}

\subsubsection{Optimization with respect to $\psi$}
To maximize with respect to $\psi$, we isolate corresponding terms and add Lagrange multipliers:
{\normalsize
\begin{equation*}
\begin{aligned}
\mathcal{L}_{[\psi]} &=\sum_{d=1}^D \sum_{i=1}^N \sum_{k=1}^K \sum_{j=1}^V \gamma_{ik} ({w^{d}})_i^{j} \log \psi_{kj} + \sum_{k=1}^K \lambda_k(\sum_{j=1}^v \psi_{kj} -1).
\end{aligned}
\end{equation*}
}
Take the derivative with respect to $\psi_{kj}$ over the whole corpus, and set it to zero, we get:
{\normalsize
\begin{equation*}
\begin{aligned}
&\psi_{kj} \propto \sum_{d=1}^D \sum_{i=1}^N \gamma_{ik}^{d} ({w^{d}})_i^{j}.
\end{aligned}
\end{equation*}
}

\subsubsection{Optimization with respect to $\mu$}
For the Dirichlet parameters $\mu$, the involved terms are:
{\normalsize
\begin{equation*}
\begin{aligned}
\mathcal{L}_{[\mu]} &= \sum_{d=1}^D \left(\log \Gamma(\sum_{j=1}^K \mu_j) - \sum_{i=1}^K \log \Gamma(\mu_i) + \sum_{i=1}^K(\mu_i -1)(\Psi(\rho_i^d) - \Psi(\sum_{j=1}^K \rho_j^d)) \right).
\end{aligned}
\end{equation*}
}
Taking the derivative with respect to $\mu_i$ gives:
{\normalsize
\begin{equation*}
\begin{aligned}
\mathcal{L}^{'}_{[\mu_i]} = D\left(\Psi(\sum_{j=1}^K \mu_i) - \Psi(\mu_i)\right) + \sum_{d=1}^D \left(\Psi(\rho_i^d) - \Psi(\sum_{j=1}^K \rho_j^d)\right)
\end{aligned}
\end{equation*}
}
We can invoke the linear-time Newton-Raphson algorithm to estimate $\mu$ as same as in LDA.
\newpage
\section{Cluster Algorithm in Solution \Rmnum{3}}
\begin{figure}[h]
\begin{center}
\psscalebox{0.8 0.8} 
{
\begin{pspicture}(0,-3.2480128)(19.2,3.2480128)
\definecolor{colour0}{rgb}{0.2,0.2,0.2}
\definecolor{colour1}{rgb}{0.4,0.4,0.4}
\psframe[linecolor=colour0, linewidth=0.016, dimen=middle](9.0,2.7819872)(0.6,-3.2180128)
\psline[linecolor=colour0, linewidth=0.016](1.8,2.7819872)(1.8,-3.2180128)
\psline[linecolor=colour0, linewidth=0.016](3.0,2.7819872)(3.0,-3.2180128)
\psline[linecolor=colour0, linewidth=0.016](4.2,2.7819872)(4.2,-3.2180128)
\psline[linecolor=colour0, linewidth=0.016](5.4,2.7819872)(5.4,-3.2180128)
\psline[linecolor=colour0, linewidth=0.016](6.6,2.7819872)(6.6,-3.2180128)
\psline[linecolor=colour0, linewidth=0.016](7.8,2.7819872)(7.8,-3.2180128)
\psline[linecolor=colour0, linewidth=0.016](0.6,1.5819873)(9.0,1.5819873)
\psline[linecolor=colour0, linewidth=0.016](0.6,0.3819873)(9.0,0.3819873)
\psline[linecolor=colour0, linewidth=0.016](0.6,-0.8180127)(9.0,-0.8180127)
\psline[linecolor=colour0, linewidth=0.016](0.6,-2.0180128)(9.0,-2.0180128)
\rput[bl](1.1,2.9819872){$t_1$}
\rput[bl](2.3,2.9819872){$t_2$}
\rput[bl](3.5,2.9819872){$t_3$}
\rput[bl](4.7,2.9819872){$t_4$}
\rput[bl](5.9,2.9819872){$t_5$}
\rput[bl](7.1,2.9819872){$t_6$}
\rput[bl](8.3,2.9819872){$t_7$}
\rput[bl](0.0,2.0819874){$d_1$}
\rput[bl](0.0,0.88198733){$d_2$}
\rput[bl](0.0,-0.31801268){$d_3$}
\rput[bl](0.0,-1.5180126){$d_4$}
\rput[bl](0.0,-2.7180128){$d_5$}
\rput[bl](1.1,2.0819874){1}
\rput[bl](2.3,2.0819874){1}
\rput[bl](3.5,2.0819874){1}
\rput[bl](3.5,0.88198733){1}
\rput[bl](4.7,0.88198733){1}
\rput[bl](4.7,-2.7180128){1}
\rput[bl](5.9,-0.31801268){1}
\rput[bl](7.1,-0.31801268){1}
\rput[bl](7.1,-1.5180126){1}
\rput[bl](8.3,-1.5180126){1}
\psline[linecolor=red, linewidth=0.06, linestyle=dashed, dash=0.17638889cm 0.10583334cm](0.6,2.7819872)(5.4,2.7819872)(5.4,0.3819873)(0.6,0.3819873)(0.6,2.7819872)
\psline[linecolor=red, linewidth=0.06, linestyle=dashed, dash=0.17638889cm 0.10583334cm](0.6,-2.0180128)(5.4,-2.0180128)(5.4,-3.2180128)(0.6,-3.2180128)(0.6,-2.0180128)
\psline[linecolor=blue, linewidth=0.06, linestyle=dotted, dotsep=0.10583334cm](5.4,0.3819873)(9.0,0.3819873)(9.0,-2.0180128)(5.4,-2.0180128)(5.4,0.3819873)
\psframe[linecolor=colour1, linewidth=0.016, dimen=outer](19.2,0.5819873)(10.5,-0.1180127)
\rput[bl](10.0,0.08198731){$\pi$}
\rput[bl](10.7,-0.5180127){1}
\rput[bl](18.8,-0.5180127){L}
\psline[linecolor=colour1, linewidth=0.016](11.1,0.5819873)(11.1,-0.1180127)
\psline[linecolor=colour1, linewidth=0.016](11.7,0.5819873)(11.7,-0.1180127)
\psline[linecolor=colour1, linewidth=0.016](12.3,0.5819873)(12.3,-0.1180127)
\psline[linecolor=colour1, linewidth=0.016](12.9,0.5819873)(12.9,-0.1180127)
\psline[linecolor=colour1, linewidth=0.016](13.5,0.5819873)(13.5,-0.1180127)
\psline[linecolor=colour1, linewidth=0.016](14.1,0.5819873)(14.1,-0.1180127)
\psline[linecolor=colour1, linewidth=0.016](14.7,0.5819873)(14.7,-0.1180127)
\psline[linecolor=colour1, linewidth=0.016](15.3,0.5819873)(15.3,-0.1180127)
\psline[linecolor=colour1, linewidth=0.016](15.9,0.5819873)(15.9,-0.1180127)
\psline[linecolor=colour1, linewidth=0.016](18.6,0.5819873)(18.6,-0.1180127)
\psline[linecolor=colour1, linewidth=0.016](18.0,0.5819873)(18.0,-0.1180127)
\psline[linecolor=colour1, linewidth=0.016](17.4,0.5819873)(17.4,-0.1180127)
\psframe[linecolor=red, linewidth=0.04, linestyle=dashed, dash=0.17638889cm 0.10583334cm, dimen=outer, framearc=0.15](12.9,0.5819873)(10.5,-0.1180127)
\psframe[linecolor=blue, linewidth=0.04, linestyle=dotted, dotsep=0.10583334cm, dimen=outer, framearc=0.15](15.9,0.5819873)(12.9,-0.1180127)
\psframe[linecolor=green, linewidth=0.04, linestyle=dashed, dash=0.17638889cm 0.10583334cm, dimen=outer, framearc=0.15](19.2,0.5819873)(17.4,-0.1180127)
\rput[bl](11.1,0.9819873){Cluster 1}
\rput[bl](13.6,0.9819873){Cluster 2}
\rput[bl](17.6,0.9819873){Cluster c}
\psdots[linecolor=colour0, dotsize=0.09527344](16.64,0.1819873)
\psdots[linecolor=colour0, dotsize=0.09527344](16.84,0.1819873)
\psdots[linecolor=colour0, dotsize=0.09527344](16.44,0.1819873)
\end{pspicture}
}
\end{center}
\caption{Left: An example of the clustering result. Each row represents a document $d$ in a corpora $D$, and Each column represents a tag $t$. $D_{ij} = 1$ means that $t_j$ is given in $d_i$. The documents in the red circle belong to one cluster, and the documents in the blue circle belong to another cluster. Right: The illustration to update $\pi$ by combine the different parts.}
\label{fig:cluster}
\end{figure}
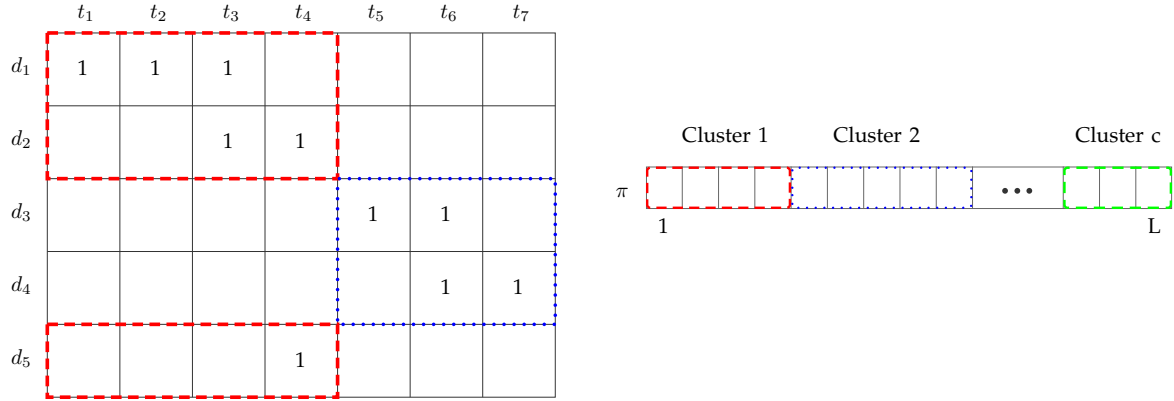

As shown in Eq.~4, $\pi_l, l \in (1, \cdots, L)$ is only associated with the documents who contain the $l^{th}$ tag. Thus, before running TWTM, we can cluster the documents into several clusters with the condition that the documents which contain the same tags should be in the same cluster. It means that the documents are divided into the mutually independent space by the tags. We show a simple example as shown in Figure~\ref{fig:cluster}, left panel.

After document clustering, the tags contained in one cluster are not appeared to any other clusters. In this case, we could assign each cluster to different computed nodes. When update the $\pi$, we just simply combine the $\pi^c$ where $c \in (1, \cdots, C)$ and $C$ is the number of document clusters, just as shown as in Figure~\ref{fig:cluster}, right panel.
We show the cluster process of Solution \Rmnum{3} in Algorithm~\ref{table:cluster_Algorithm}.

\begin{algorithm}\small
\caption{The cluster process of Solution \Rmnum{3}}
\label{table:cluster_Algorithm}
\begin{algorithmic}[1]
\STATE{\textbf{Input:} a semi-structured corpora $D = \{(\mathbf{w}^{1}, \mathbf{t}^{1}), \ldots, (\mathbf{w}^{M}, \mathbf{t}^{M}) \}$ and the tag set $T$ of the corpora.}
\STATE{\textbf{Output:} a cluster set $C$ that contains all the clusters, and each cluster $c$ in $C$ contains a set of documents.}
\STATE{create a cluster set $C=\{\}$.}
\STATE{create a document cluster $c=\{\}$.}
\STATE{create $pre\_added\_docs = \{\}$ to store the documents which are ready to add into cluster $c$.}
\STATE{create a tag set $scanned\_tags = \{\}$ to store the tags which have been scanned.}
\STATE{add $c$ into $C$.}

\FOR{ each tag $t$ in $T$}
\IF{$t$ is not in $scanned\_tags$}
\STATE{add $t$ into $scanned\_tags$;}
\STATE{create a new cluster $c$, and add $c$ into $C$;}
\ELSE
\STATE{continue;}
\ENDIF
\STATE{add the documents which own $t$ into $pre\_added\_docs$;}
\REPEAT
\FOR{ each $d$ in the $pre\_added\_docs$}
\IF{$d$ is not in $c$}
\STATE{add $d$ into $c$;}
\ENDIF
\FOR{each tag $t^d$ in $d$}
\STATE{add $t^d$ into $scanned\_tags$;}
\STATE{add the documents which have $t^d$ and not in $c$ into $pre\_added\_docs$;}
\ENDFOR
\STATE{remove $d$ from $pre\_added\_docs$;}
\ENDFOR
\UNTIL{$pre\_added\_docs$ is empty.}
\ENDFOR
\RETURN{$C$}
\end{algorithmic}
\end{algorithm}

\end{document}